# Robust Perspective Correction for Real-World Crack Evolution Tracking in Image-Based Structural Health Monitoring

Xinxin Sun* (ORCID: 0000-0002-9963-2109) and Peter Chang

Department of Civil and Environmental Engineering, University of Maryland, College Park, MD 20742, USA

*Corresponding Author

Email: xinxin68@umd.edu

**Abstract**

Accurate image alignment is essential for monitoring crack evolution in structural health monitoring (SHM), particularly under real-world conditions involving perspective distortion, occlusion, and low contrast. However, traditional feature detectors—such as SIFT and SURF, which rely on Gaussian-based scale spaces—tend to suppress high-frequency edges, making them unsuitable for thin crack localization. Lightweight binary alternatives like ORB and BRISK, while computationally efficient, often suffer from poor keypoint repeatability on textured or shadowed surfaces.

This study presents a physics-informed alignment framework that adapts the open KAZE architecture to SHM-specific challenges. By utilizing nonlinear anisotropic diffusion to construct a crack-preserving scale space, and integrating RANSAC-based homography estimation, the framework enables accurate geometric correction without the need for training, parameter tuning, or prior calibration.

The method is validated on time-lapse images of masonry and concrete acquired via handheld smartphone under varied field conditions, including shadow interference, cropping, oblique viewing angles, and surface clutter. Compared to classical detectors, the proposed framework reduces crack area and spine length errors by up to 70% and 90%, respectively, while maintaining sub-5% alignment error in key metrics under typical field conditions.

Unsupervised, interpretable, and computationally lightweight, this approach supports scalable deployment via UAVs and mobile platforms. By tailoring nonlinear scale-space modeling to SHM image alignment, this work offers a robust and physically grounded alternative to conventional techniques for tracking real-world crack evolution.

## 1. Introduction

Structural Health Monitoring (SHM) plays a critical role in maintaining the safety, serviceability, and longevity of civil infrastructure. Surface cracks, a primary focus of non-destructive evaluation (NDE), are among the earliest indicators of material degradation. If undetected, they can compromise load-bearing capacity, accelerate fatigue or corrosion, and lead to structural failure, economic loss, and public safety risks [1], [2], [3]. For example, several recent bridge collapses—including the 2024 Francis Scott Key Bridge incident—have been linked to undetected local failures, emphasizing the importance of robust monitoring [4].

Vision-based inspection has emerged as a non-contact, scalable, and cost-effective alternative to manual surveys, supporting both classical image processing and learning-based paradigms [5], [6], [7]. However, tracking crack evolution across time-lapse images remains difficult due to geometric inconsistencies in field-acquired data. UAV and handheld imaging often introduce variations in viewpoint and distance, resulting in distortion, rotation, and misalignment—factors that constitute key obstacles in SHM imaging workflows [8], [9]. These distortions skew crack area, length, and orientation measurements, compromising condition assessments and temporal analysis.

To reveal the impact of geometric distortion, this study conducts controlled imaging experiments to quantify its effect on crack metrics. Results show that even minor viewpoint changes can cause up to 5% error in crack area and over 36% deviation in spine length (i.e., the longitudinal extent of the crack), as shown in Table 1.1.

Table 1.1: Crack metric variations caused by viewpoint distortion under consistent scale normalization

| Image | 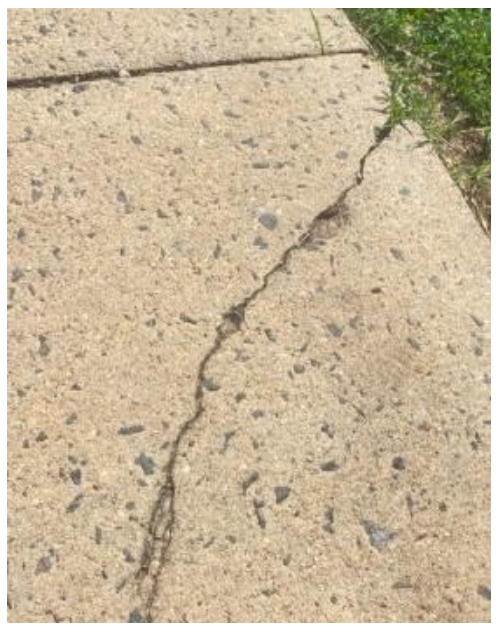 | 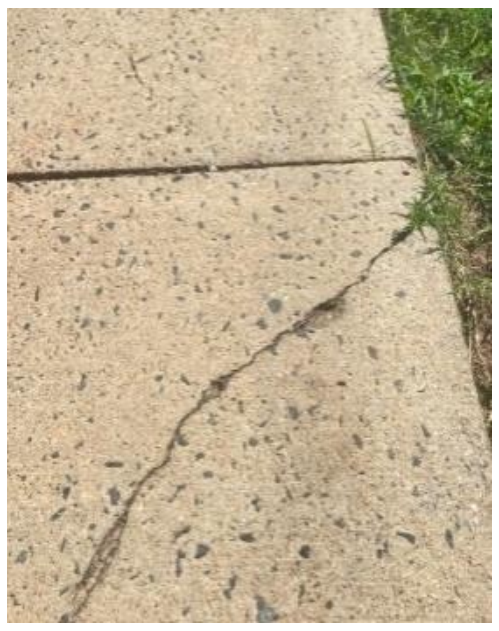 | 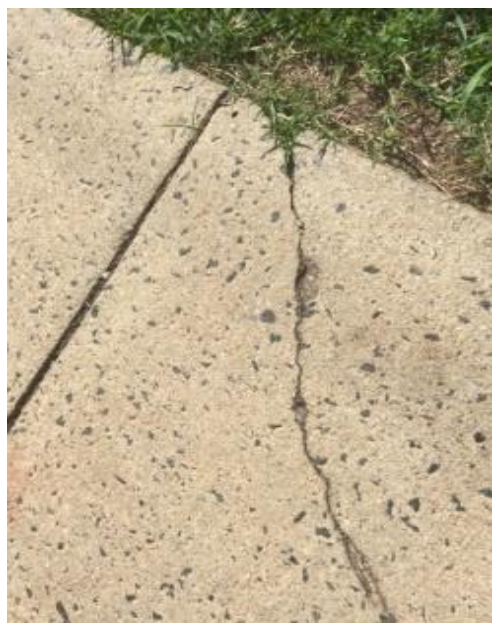 |
|---|---|---|---|
| Crack area (pixel) | 4968 | 5231 | 5236 |
| Spine length (pixel) | 1339 | 1558 | 1149 |
| Crack average width | 4.0 | 3.9 | 5.0 |

*Note: Images were captured under natural lighting to simulate field conditions.*

To address such misalignment, keypoint-based matching methods—such as SIFT (Scale-Invariant Feature Transform), SURF (Speeded-Up Robust Features), ORB (Oriented FAST and Rotated BRIEF), and BRISK (Binary Robust Invariant Scalable Keypoints)—are widely adopted in SHM and computer vision tasks [10], [11]. While SIFT and SURF explicitly construct Difference-of-Gaussian (DoG) or Laplacian-of-Gaussian (LoG) pyramids, ORB and BRISK achieve scale invariance using Gaussian-based image pyramids without relying on DoG. However, Gaussian filtering suppresses the high-frequency edge information required for crack localization, reducing keypoint stability under conditions such as shadow, texture, or noise [12], [13].

Moreover, many existing approaches assume ideal imaging conditions or focus exclusively on single-frame detection, limiting their applicability in uncontrolled environments and long-term SHM scenarios [14].

To overcome these limitations, this study proposes a physics-informed image alignment framework based on the open KAZE architecture. KAZE constructs a nonlinear scale space using anisotropic diffusion, which selectively preserves structural discontinuities (e.g., cracks) while reducing lighting variations and surface noise. Feature points extracted from this enhanced representation are matched and filtered through a RANSAC-based homography estimation pipeline, enabling robust geometric correction across time-lapse imagery.

Rather than introducing a new algorithm, this work repurposes the KAZE framework—originally designed for general-purpose vision tasks—for SHM-specific crack alignment. This adaptation leverages KAZE's nonlinear scale space to address domain-specific challenges, offering a novel integration of physics-informed feature extraction and robust geometric correction tailored to structural integrity monitoring.

The key contributions of this study are:

- Proposing a nonlinear, crack-preserving scale space using anisotropic diffusion, which significantly enhances localization under shadows, texture, and low-contrast conditions.
- Implementing robust homography estimation via a RANSAC-based geometric correction pipeline, improving time-lapse crack alignment in uncontrolled imaging environments.
- Developing a deployable, training-free crack monitoring framework optimized for real-time processing on UAVs and mobile platforms, supporting long-term SHM in resource-constrained field environments.

Experimental results show that the proposed method significantly outperforms classical detectors in alignment accuracy and robustness across diverse real-world conditions. The remainder of this paper is structured as follows: Section 2 details the methodology, and Section 3 presents the experimental evaluation.

.

## 2. Methodology

This study addresses a core challenge in vision-based SHM: ensuring consistent spatial alignment of crack imagery over time despite geometric distortions introduced by field acquisition. In long-term monitoring scenarios, images are often captured from variable viewpoints due to factors such as UAV motion, operator positioning, or environmental constraints. These variations in perspective, scale, and rotation distort the apparent geometry of cracks and hinder direct comparisons across time-lapse sequences. Meanwhile, cracks themselves may evolve between inspections due to aging, loading, or environmental exposure, further altering their appearance and complicating tracking. Accurate crack evolution tracking is critical for predicting structural degradation and informing maintenance decisions.

To support reliable crack evolution monitoring, a robust alignment framework is needed—one that compensates for acquisition-induced distortions while preserving the fine structural features critical for quantitative crack assessment. Feature-based image alignment pipelines offer a promising solution. These methods typically consist of two key stages: keypoint detection and geometric transformation estimation. Once repeatable features are extracted from a pair of images, a transformation model (e.g., homography) is computed to align them.

Many prior works adopt classical feature detectors such as SIFT, SURF, ORB, and BRISK due to their efficiency and success in general-purpose image matching [15]. These methods construct multiscale representations using Gaussian filtering and rely on gradient magnitude or intensity comparisons to select salient keypoints. Although these methods perform well in textured environments, their reliability diminishes in SHM scenarios where fine cracks are visually degraded by shadow interference, surface noise, or contrast loss—highlighting the need for edge-preserving alternatives in challenging field conditions.

Experimental observations in this study confirm that under common field conditions—such as weak lighting, shadow interference, and surface noise—these detectors often fail to identify enough stable, distinctive features for reliable alignment. The root cause lies in their reliance on Gaussian-based filtering, which tends to smooth high-frequency edge content and blur fine crack boundaries. These limitations motivate the need for alternative detection approaches tailored to the unique geometric and visual characteristics of cracks.

The following sections revisit four classical feature detection methods—SIFT, SURF, ORB, and BRISK—as introduced in Section 1—and evaluate their theoretical foundations and performance in SHM contexts.

### 2.1 Traditional Feature Detection Methods

Feature detection and matching techniques form the foundation of many image alignment pipelines in vision-based SHM. These methods are commonly employed to correct geometric distortions caused by variations in perspective, scale, and rotation across time-lapse imagery. This section reviews four widely used algorithms—SIFT, SURF, ORB, and BRISK—and summarizes their underlying mathematical principles and limitations in the context of SHM.

#### 2.1.1 SIFT

SIFT constructs a scale-space representation by convolving the input image with Gaussian kernels of increasing standard deviation:

$$L(x, y, \sigma) = G(x, y, \sigma) \cdot I(x, y) \quad (1)$$

Where $L(x, y, \sigma)$ is the scale-space image, $G(x, y, \sigma)$ is the Gaussian kernel, and $I(x, y)$ is the original image [16]. Keypoints are detected as extrema in the Difference of Gaussians (DoG), computed as:

$$DoG(x, y, \sigma) = L(x, y, k\sigma) - L(x, y, \sigma) \quad (2)$$

where $k$ is a constant multiplicative factor between successive scales.

**2.1.2 SURF**

SURF approximates Gaussian convolution with box filters to accelerate computation. Keypoints are identified using the determinant of the Hessian matrix:

$$H(x, y, \sigma) = \begin{bmatrix} L_{xx}(x, y, \sigma) & L_{xy}(x, y, \sigma) \\ L_{xy}(x, y, \sigma) & L_{yy}(x, y, \sigma) \end{bmatrix} \quad (3)$$

$$Det(H) = L_{xx}L_{yy} - (0.9 \cdot L_{xy})^2 \quad (4)$$

where $L_{xx}$, $L_{xy}$, and $L_{yy}$ are the second-order Gaussian derivatives [17].

**2.1.3 ORB**

ORB combines the FAST keypoint detector with the BRIEF descriptor. FAST identifies corner points by evaluating intensity differences around a circular pattern:

$$|I(p) - I(p_i)| > T \quad (5)$$

where $I(p)$ and $I(p_i)$ are pixel intensities at the center and neighboring points, respectively, and is a predefined threshold [18].

**2.1.4 BRISK**

BRISK constructs a binary descriptor by comparing intensity pairs sampled around each keypoint:

$$Decriptor = \sum_{i,j} sgn(I(p_i) - I(p_j)) \quad (6)$$

Where $p_i$ and $p_j$ are sample points on a concentric circular pattern [19].

These methods are computationally efficient and perform well in textured environments. However, their reliability declines in SHM imagery where cracks are affected by dynamic shadows, low contrast, or blur—conditions that degrade keypoint stability and descriptor distinctiveness, motivating deeper analysis of their limitations. All four methods involve Gaussian-based multi-scale representations, which degrade stability under low contrast or noisy conditions.

**2.1.5 Limitations of Gaussian-Based Approaches**

A key limitation of SIFT and SURF stems from their reliance on Gaussian filtering, which inherently smooths the image and suppresses fine structural discontinuities. This process is described by:

$$L(x, y, \sigma_{n+1}) = \frac{1}{4} \sum_{m,n} G[m, n] \cdot L(2x + m, 2y + n, \sigma_n) \tag{7}$$

Where $G[m, n]$ is the Gaussian kernel and $L(\cdot)$ is the scale-space image at the previous level. This operation reduces local contrast and blurs crack edges, especially those with small widths.

In low-contrast or noisy images, such suppression leads to fewer and less distinctive keypoints, impairing match quality. As a result, Gaussian-based detectors often yield incomplete or inaccurate alignment—particularly problematic in SHM where sub-pixel variations in crack geometry are critical. These challenges underscore the need for alternative techniques that preserve edge fidelity and remain robust under uncontrolled field conditions.

*Note: While ORB and BRISK do not explicitly apply DoG or LoG filters, both rely on Gaussian pyramid structures to construct their multiscale representations. As a result, they share similar limitations in high-frequency feature suppression, particularly for thin cracks and low-contrast edges.*

## 2.2 Impact of Gaussian Blurring and Weak Gradients in Crack Detection

### 2.2.1 Gaussian Scale-Space and Crack Detail Loss

Gaussian-based methods such as SIFT employ a geometric progression of the scale parameter σ to generate multi-scale representations:

$$\sigma_{i+1} = \sigma_i \cdot \sqrt{2} \tag{8}$$

While this facilitates scale-invariant keypoint detection, it also introduces progressive blurring of image details, which is particularly problematic for thin or low-contrast crack regions.

Figure 2.1 illustrates the degradation of crack edge clarity at increasing scale levels using DoG images. As the Gaussian scale increases, subtle crack structures become increasingly diffused, reducing keypoint saliency and degrading the reliability of downstream feature matching.

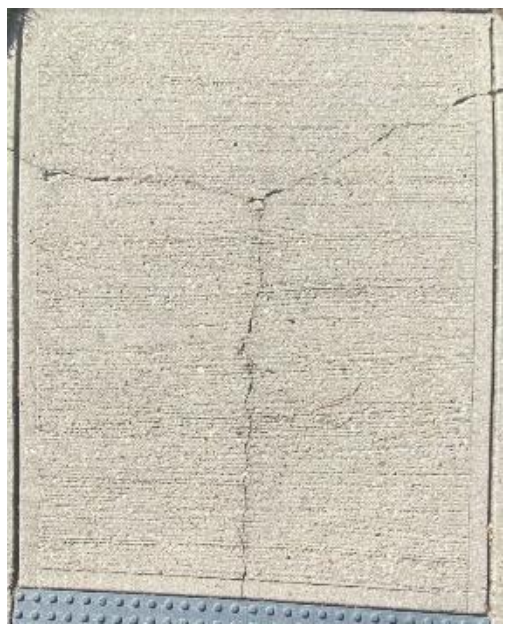

(a) the original image

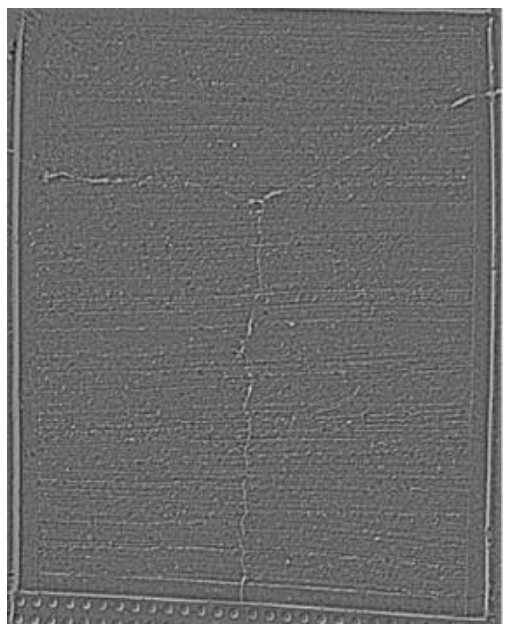

(b) the DoG image at $\Delta\sigma = (1.6 \cdot \sqrt{2}) - 1.6$

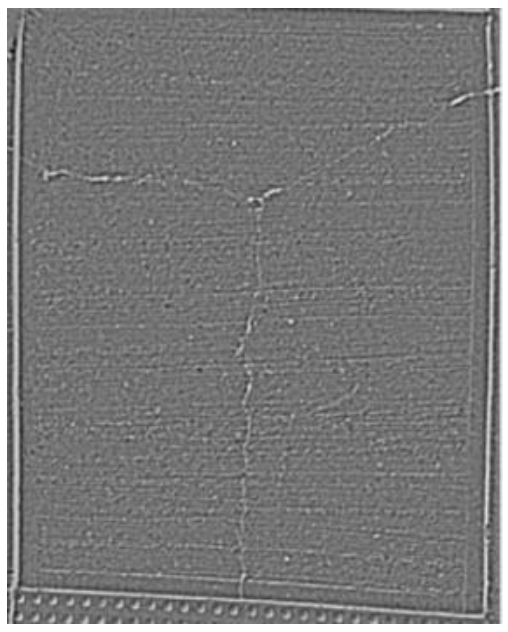

(c) the DoG image at $\Delta\sigma = (1.6 \cdot (\sqrt{2})^2) - (1.6 \cdot \sqrt{2})$

Fig. 2.1: DoG Images at Various Scales

This blurring effect significantly impacts SHM applications where accurate crack edge delineation is essential for long-term damage assessment. High-frequency crack features reflect fine widths and sharp transitions—essential for precise structural assessment.

### 2.2.2 Influence of Weak Gradients in Low-Contrast Conditions

Gradient-based feature detectors rely on pronounced intensity transitions to identify keypoints. The gradient magnitude at each pixel is defined as:

$$|G(x,y)| = \sqrt{(\frac{\partial I}{\partial x})^2 + (\frac{\partial I}{\partial y})^2} \quad (9)$$

In low-contrast crack images—common in field-acquired SHM datasets—the grayscale difference between cracks and background is often minimal, producing weak gradient responses. As shown in Fig. 2.2, this leads to reduced keypoint distinctiveness and lower feature matching accuracy.

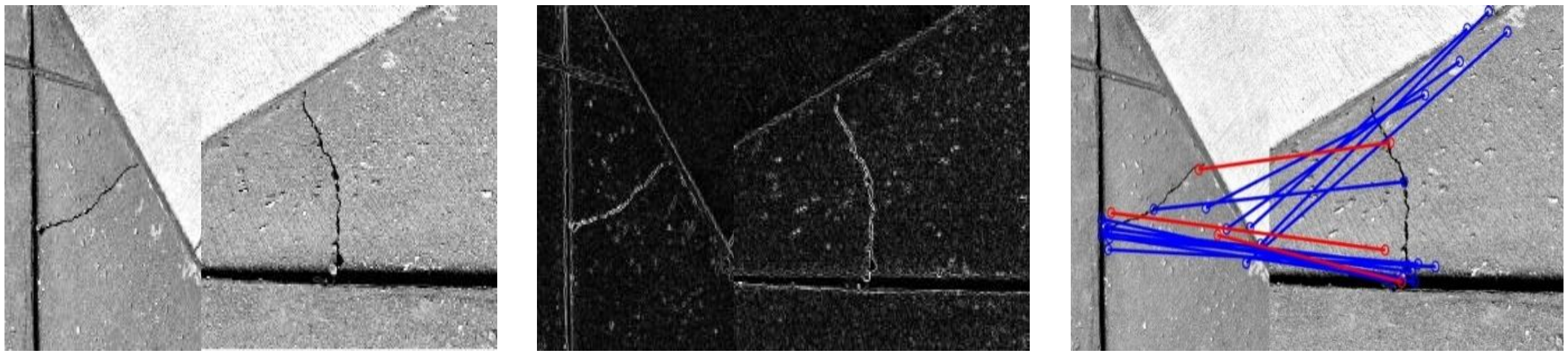

(a) High-contrast crack images (b) Gradient magnitude of high-contrast images (c) Feature matching results for high-contrast images

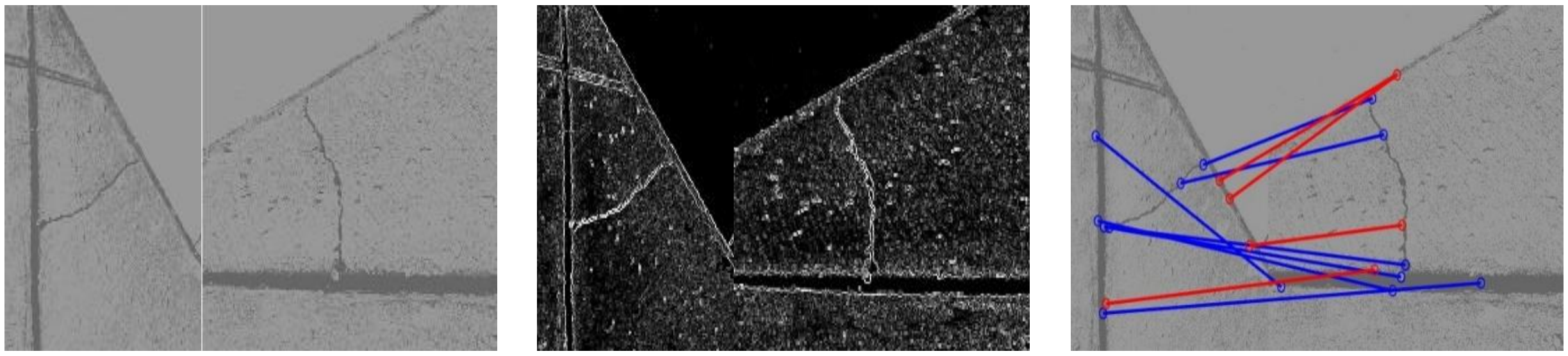

(d) Low-contrast crack images (e) Gradient magnitude of low-contrast images (f) Feature matching results for low-contrast images

Fig. 2.2: Effect of Contrast on Gradient Magnitude

In the high-contrast case (a–c), crack edges are well-preserved, yielding strong gradients and a high number of inliers (correct matches shown in blue). In contrast, the low-contrast case (d–f) exhibits blurry gradients, increased background interference, and a rise in mismatches (incorrect matches shown in red), resulting in degraded alignment performance.

Together, the effects of Gaussian blurring and weak gradient responses reveal critical limitations of conventional feature detectors when applied to crack-rich, low-texture environments typical of SHM tasks. This motivates the need for alternative approaches, such as nonlinear diffusion-based scale spaces, to enhance feature robustness and geometry-consistent alignment.

**2.3 Advanced Feature Detection using KAZE (Nonlinear Scale-Space)**

To address the limitations of traditional Gaussian-based feature detectors discussed in Section 2.1, this study adopts KAZE—a nonlinear scale-space approach designed to preserve fine crack structures under visually challenging conditions. Unlike conventional methods that blur critical features through isotropic Gaussian smoothing, KAZE constructs its scale space through anisotropic diffusion, enabling enhanced edge preservation and robustness to noise.

The nonlinear scale space is constructed via the partial differential equation:

$$\frac{\partial L}{\partial t} = div(c(x, y, t) \cdot \nabla L) \tag{10}$$

where $L$ is the evolving image at scale $t$, $\nabla L$ is the image gradient, and $c(x, y, t)$ is the conductivity function that governs the diffusion process. The conductivity is defined as:

$$c(x, y, t) = \frac{1}{1+(\frac{|\nabla L|}{\kappa})^2} \tag{11}$$

where $|\nabla L|$ denotes the gradient magnitude and $\kappa$ is a contrast-sensitive parameter that controls edge preservation. High gradient regions (e.g., crack edges) experience reduced diffusion, thus maintaining their sharpness, while homogeneous areas are smoothed to suppress noise.

Through this nonlinear framework, KAZE dynamically adapts to local image structure, retaining fine crack contours that are often lost under Gaussian smoothing. This makes it particularly effective for detecting features in low-texture, low-contrast, and noise-contaminated crack imagery—conditions commonly encountered in field-based SHM [11], [20], [21], [22], [23]. These characteristics make KAZE an ideal candidate for robust keypoint detection in time-lapse crack monitoring applications, motivating its comparative evaluation in the following section. Unlike Gaussian-based methods, KAZE preserves high-frequency crack structures through anisotropic diffusion, enabling geometry-consistent alignment in low-contrast, noisy field imagery.

**2.4 Comparative Evaluation of Feature Detectors Under Challenging Conditions**

To evaluate the performance of KAZE against traditional feature detection algorithms (SIFT, SURF, ORB, and BRISK), a series of controlled experiments were conducted under diverse image conditions. The evaluation used field-acquired crack images captured by an iPhone 11 at chest height (~1.3–1.4 m) under natural outdoor lighting, without artificial enhancement. Each detector was tested on these images with systematically varied texture, contrast, noise, blur, and perspective distortion. Performance was assessed both qualitatively (via visual matching outcomes) and quantitatively (via the number of inliers identified through homography estimation).

Table 2.1: Visual Feature Matching Results under High, Medium, and Low Texture Conditions

| | High Texture | Medium Texture | Low Texture |
|---|---|---|---|
| Images | 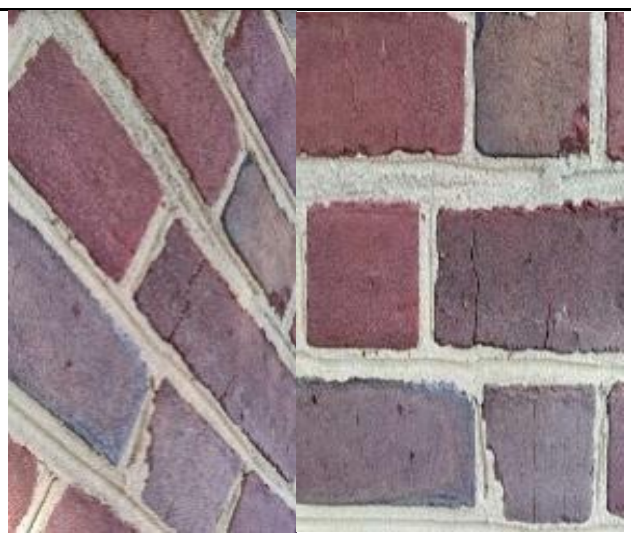 | 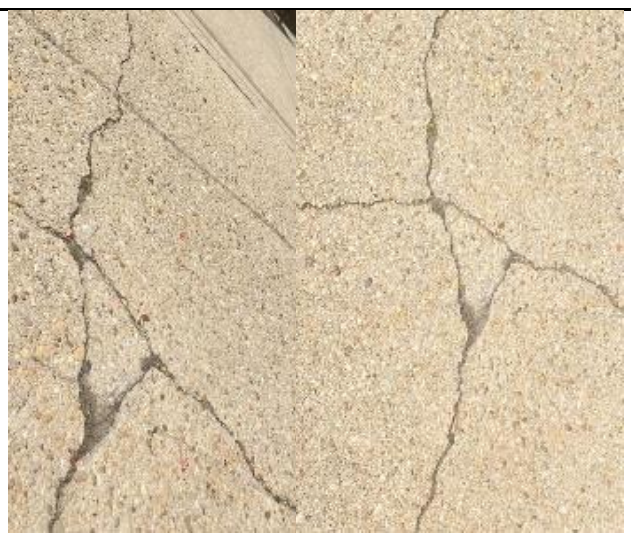 | 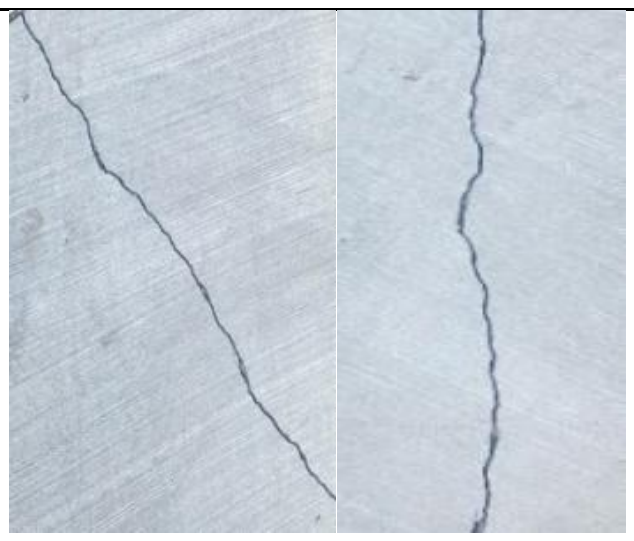 |
| SIFT | 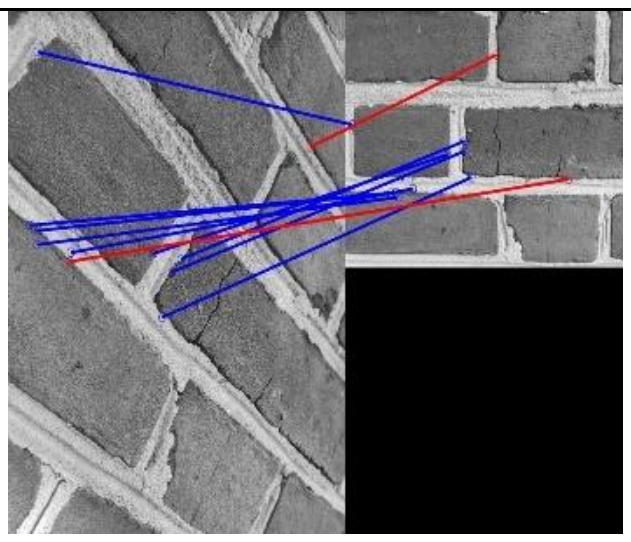 | 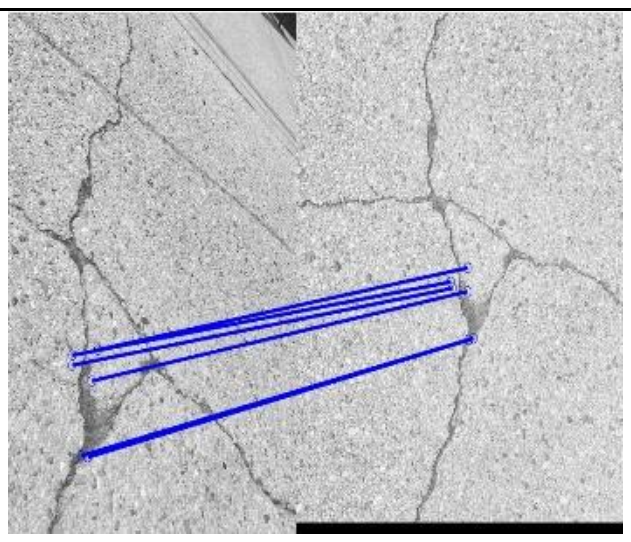 | 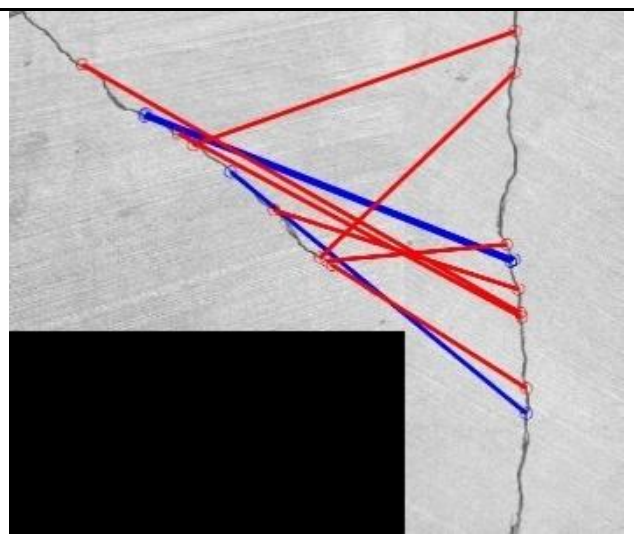 |
| SURF | 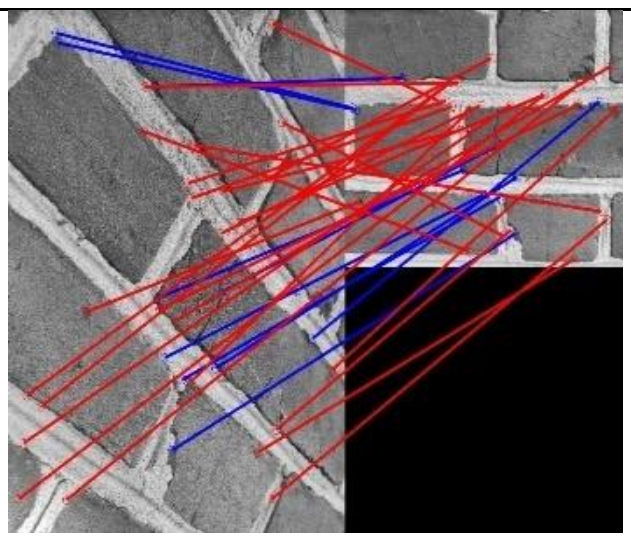 | 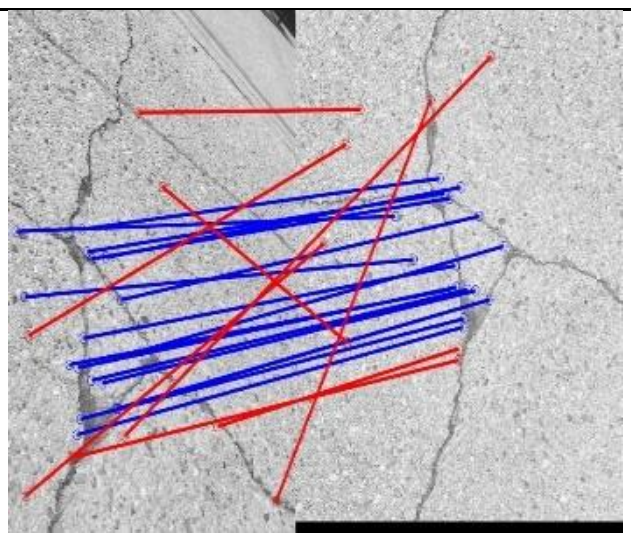 | 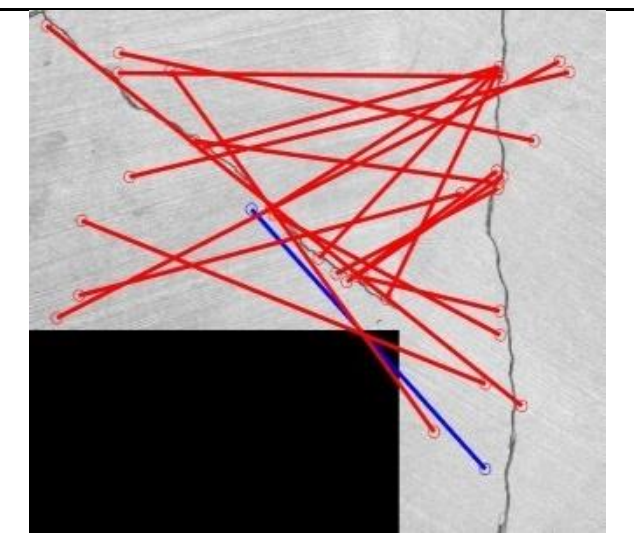 |
| ORB | 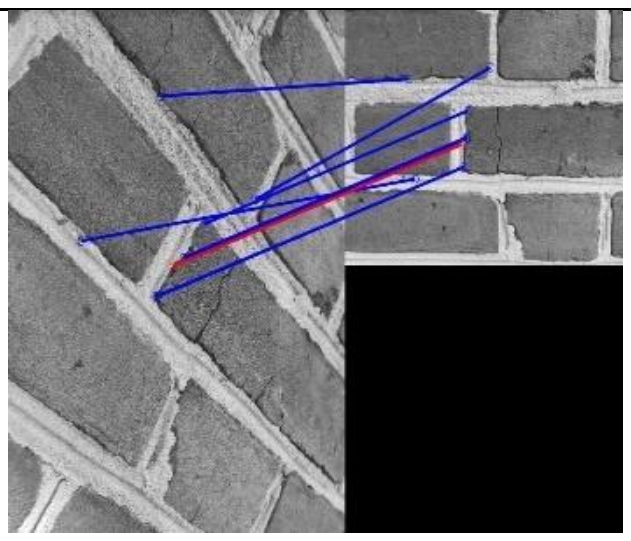 | 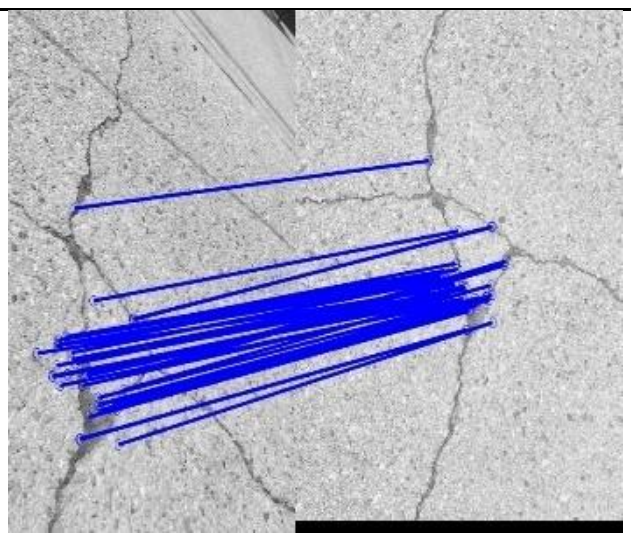 | 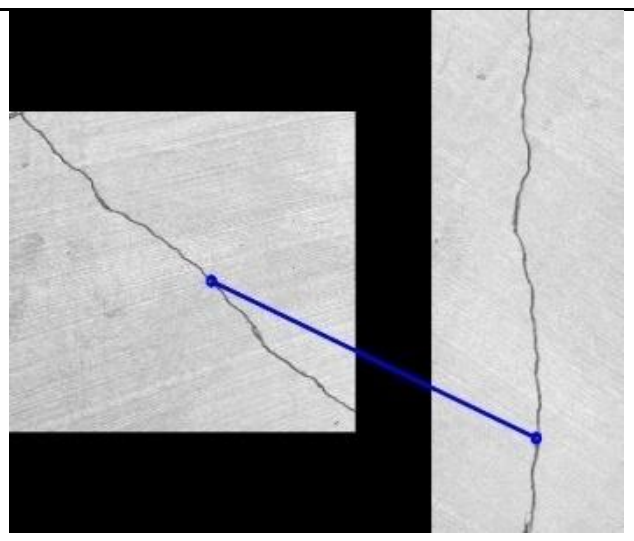 |
| BRISK | 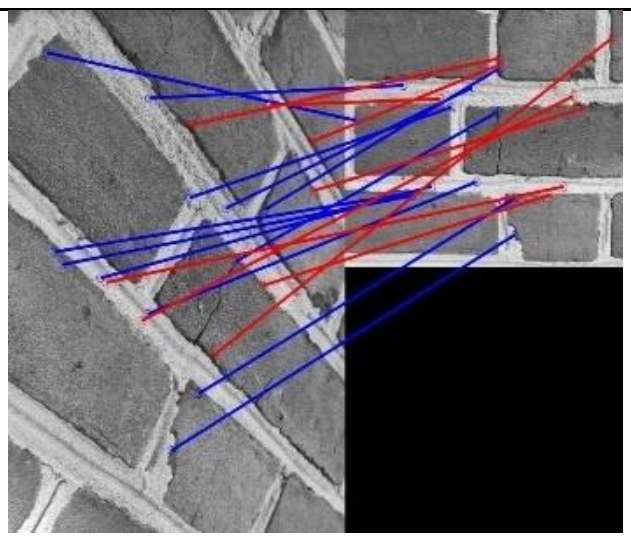 | 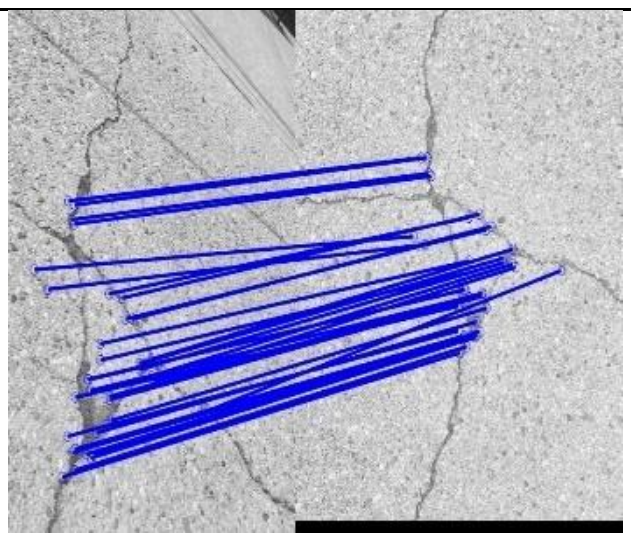 | 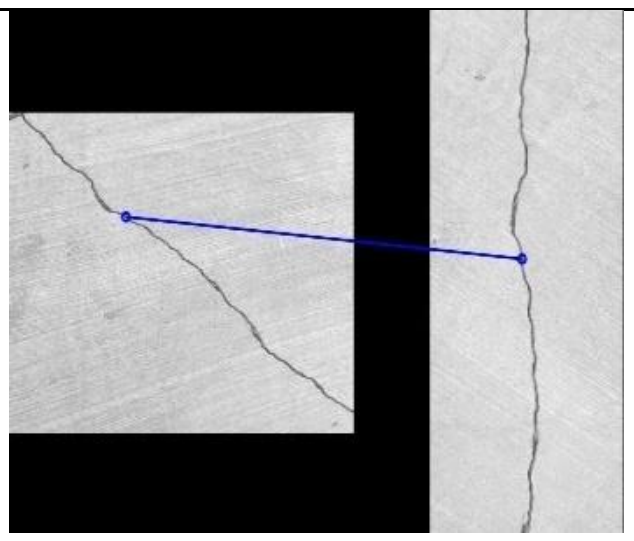 |
| KAZE | 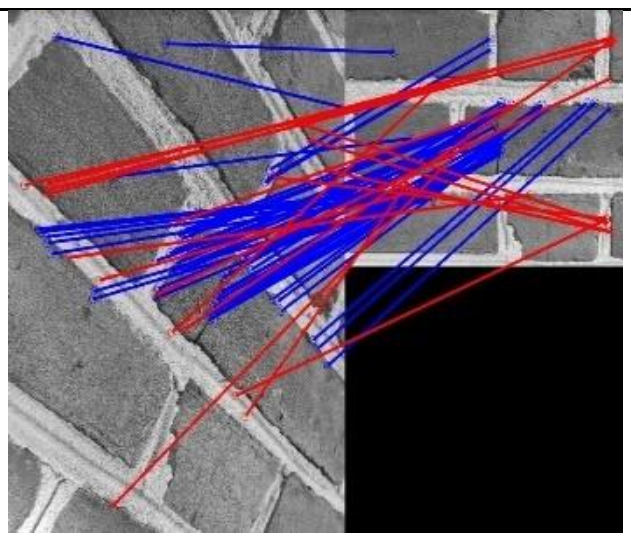 | 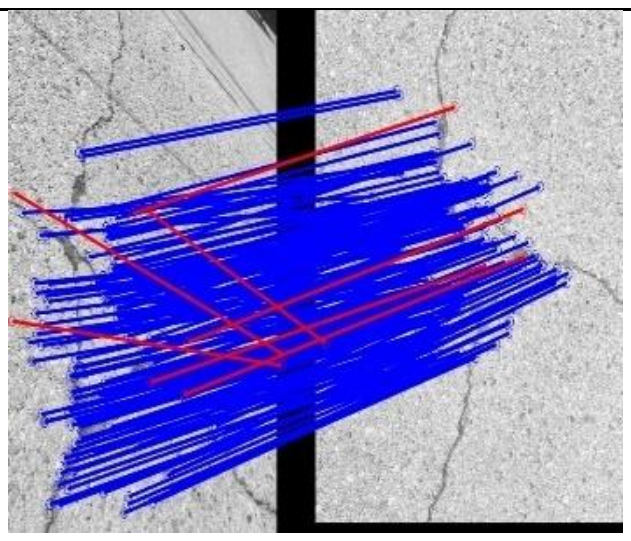 | 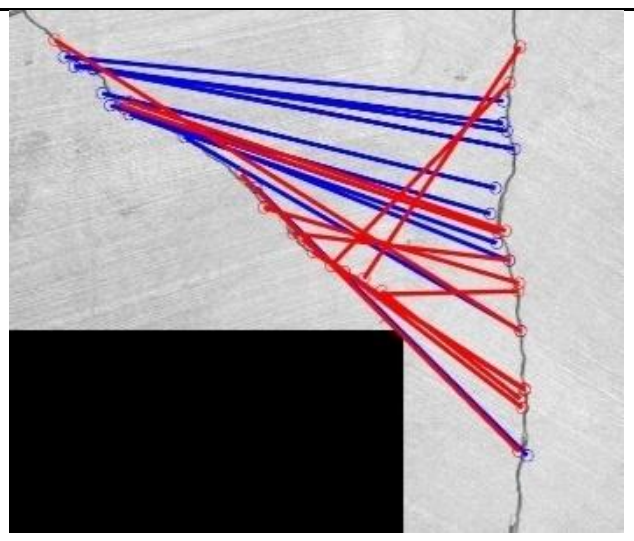 |

Table 2.2: Inlier Count Comparison Under Texture Variations

| Texture \ Inliers | SIFT | SURF | ORB | BRISK | KAZE |
|---|---|---|---|---|---|
| High | 9 | 10 | 5 | 5 | 53 |
| Medium | 6 | 17 | 43 | 35 | 180 |
| Low | 3 | 1 | 1 | 1 | 14 |

Table 2.3: Visual Feature Matching Results Under High, Medium, and Low Contrast Conditions

| | High Contrast | Medium Contrast | Low Contrast |
|---|---|---|---|
| Image | 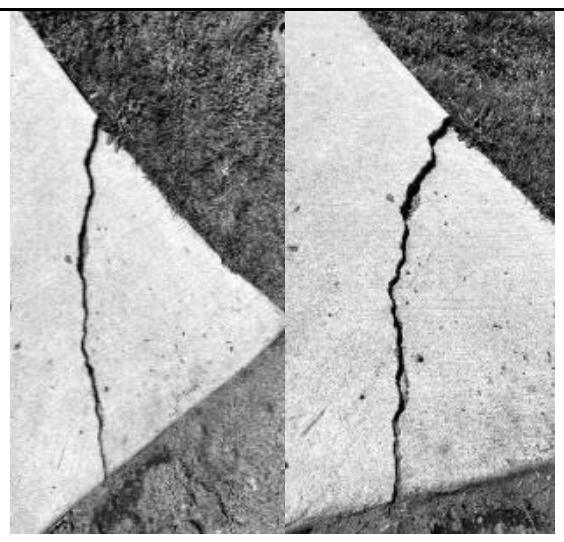 | 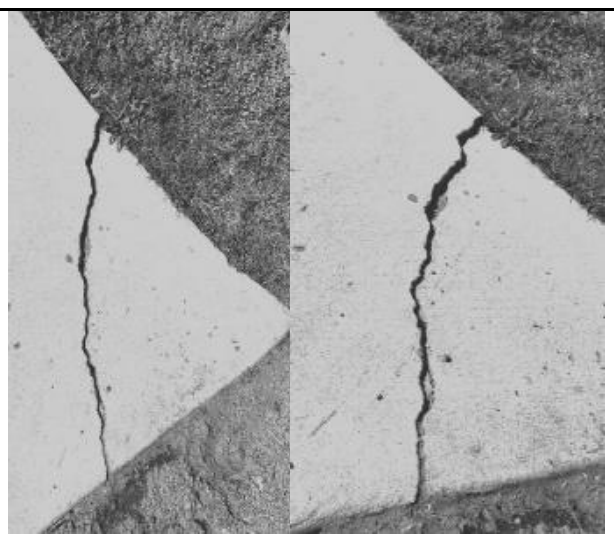 | 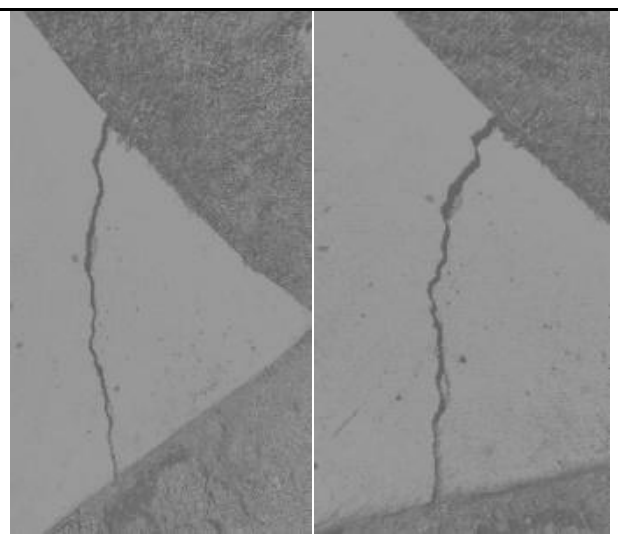 |
| SIFT | 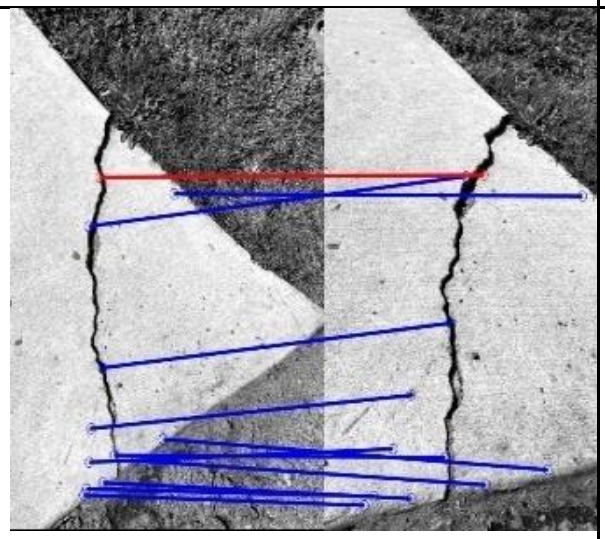 | 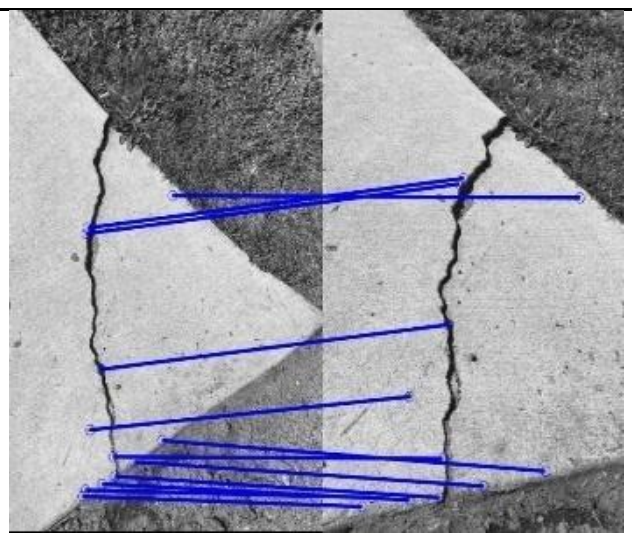 | 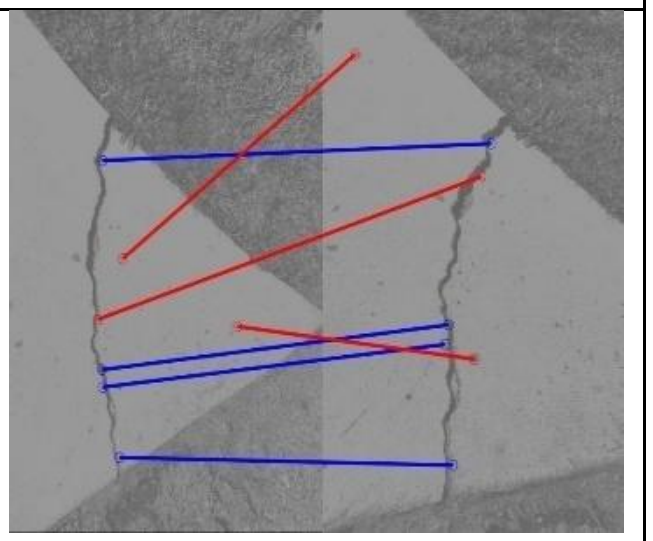 |
| SURF | 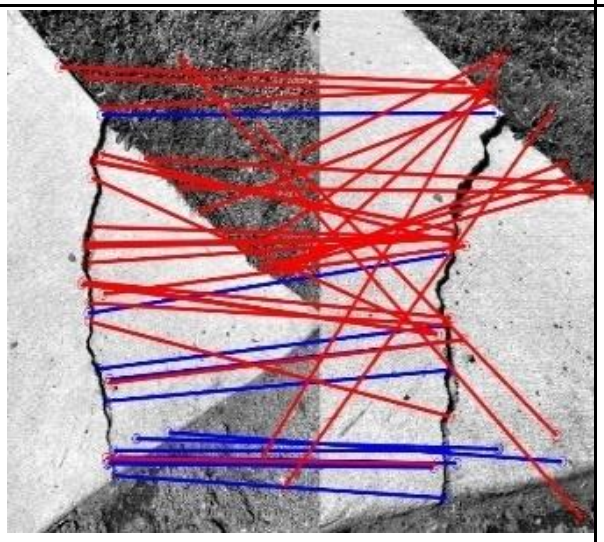 | 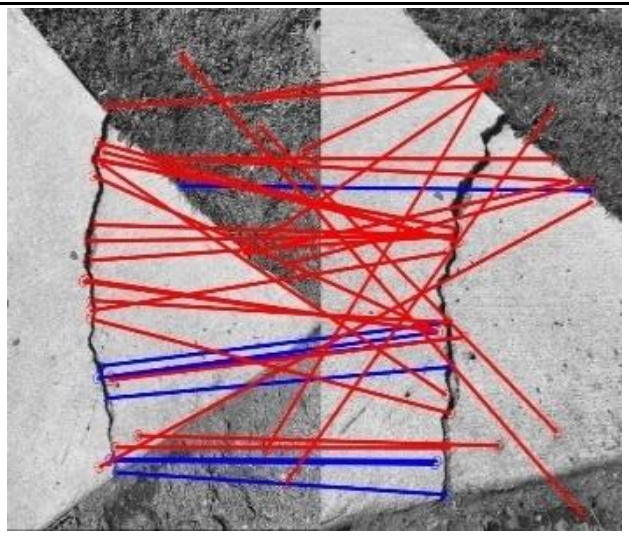 | 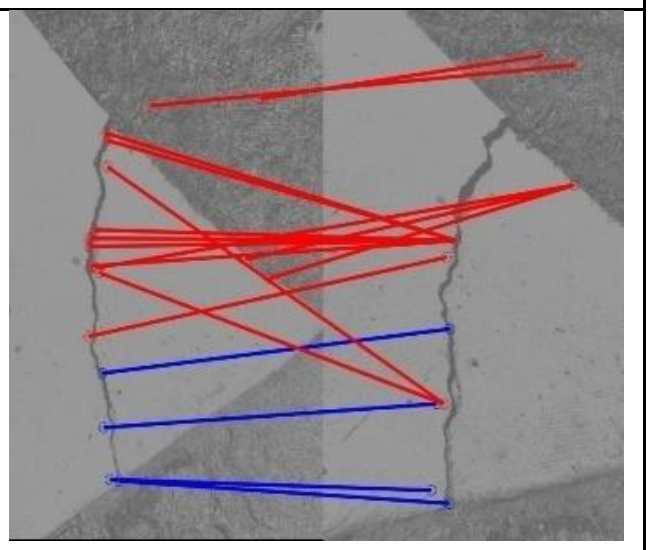 |
| ORB | 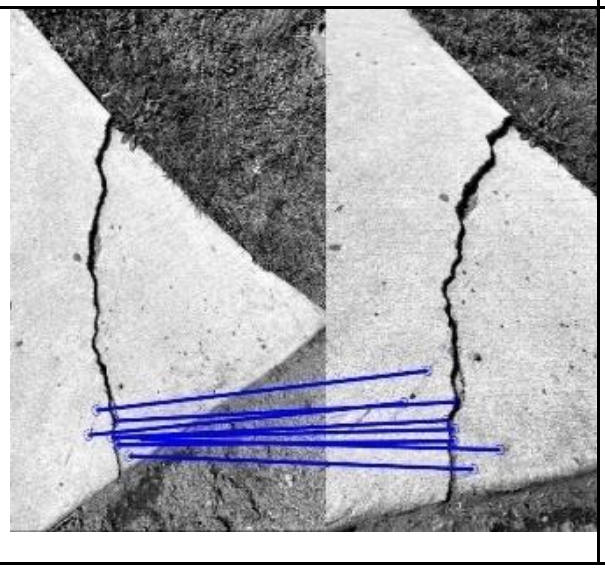 | 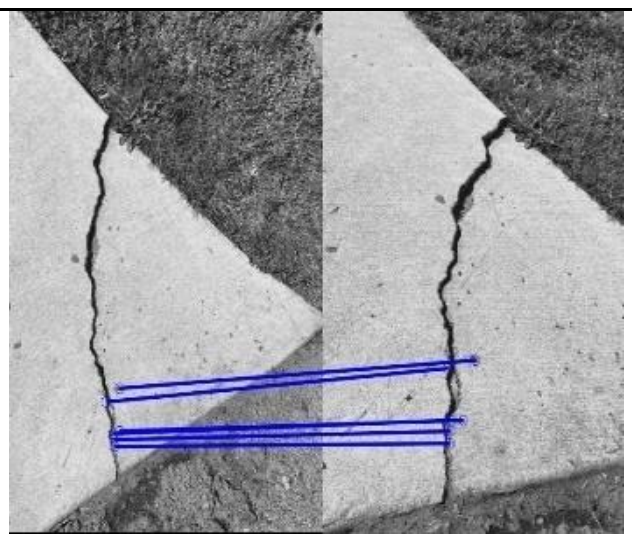 | 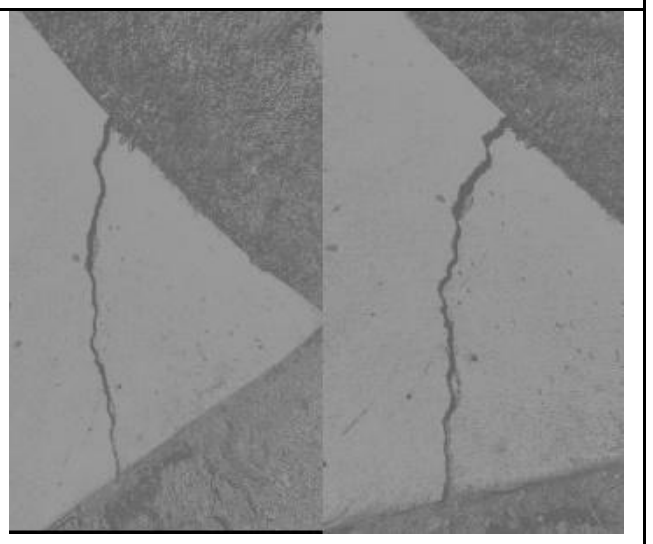 |
| BRISK | 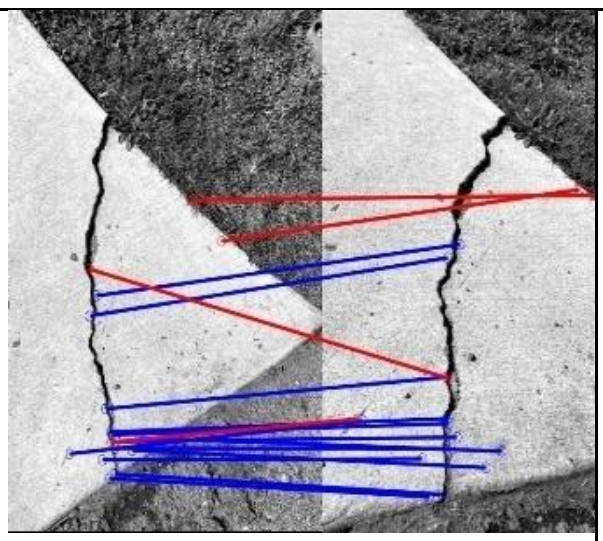 | 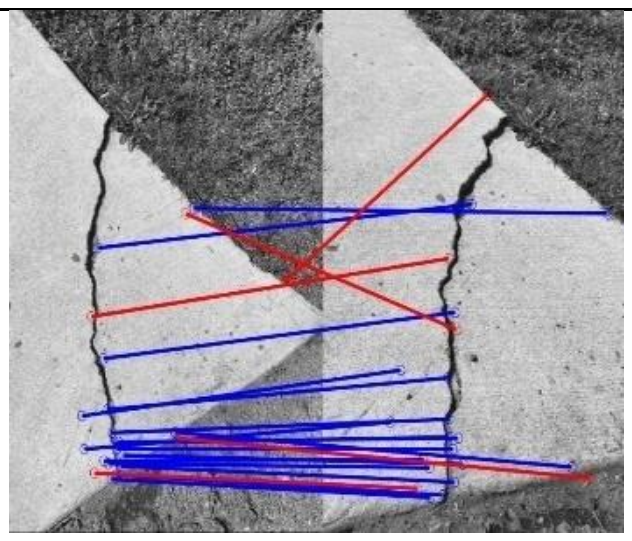 | 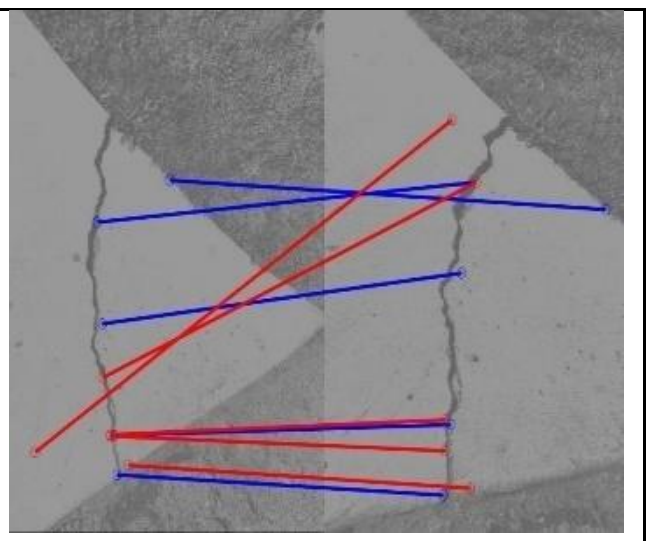 |
| KAZE | 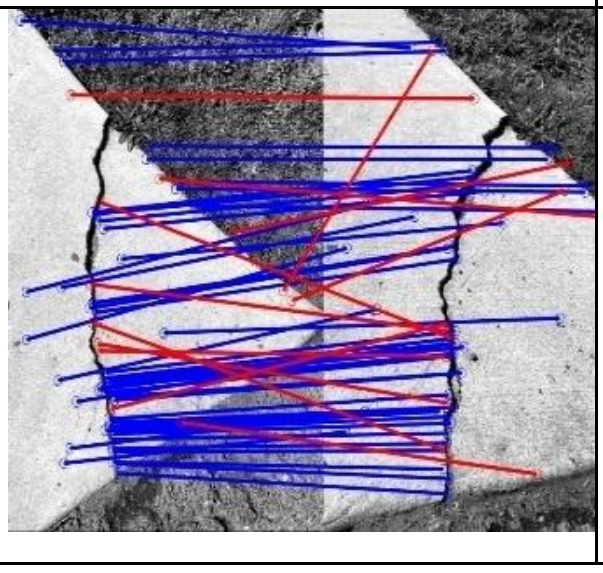 | 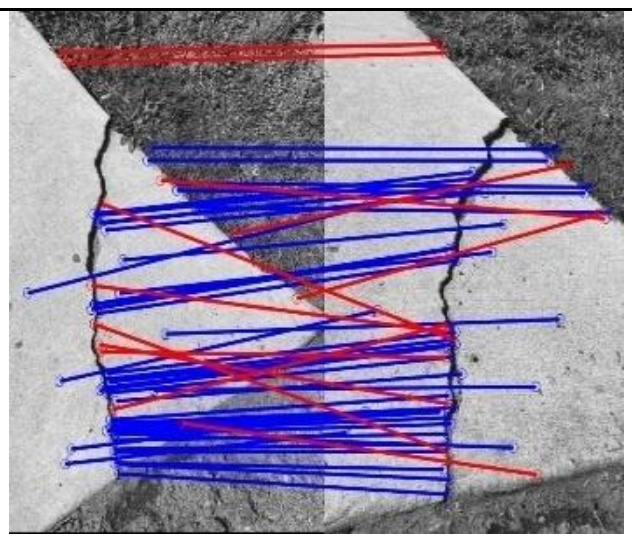 | 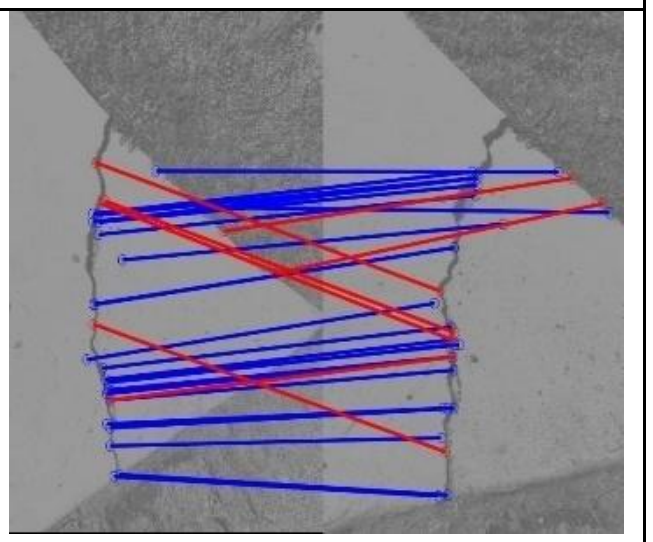 |

Table 2.4: Inlier Count Comparison Under Contrast Variations

| Contrast \ inliers | SIFT | SURF | ORB | BRISK | KAZE |
|---|---|---|---|---|---|
| High | 12 | 12 | 8 | 13 | 48 |
| Medium | 13 | 8 | 6 | 17 | 43 |
| Low | 4 | 4 | 0 | 5 | 24 |

Table 2.5: Visual Feature Matching Results Under Low, Medium, and High Noise Conditions

| | Low Noise | Medium Noise | High Noise |
|---|---|---|---|
| Image | 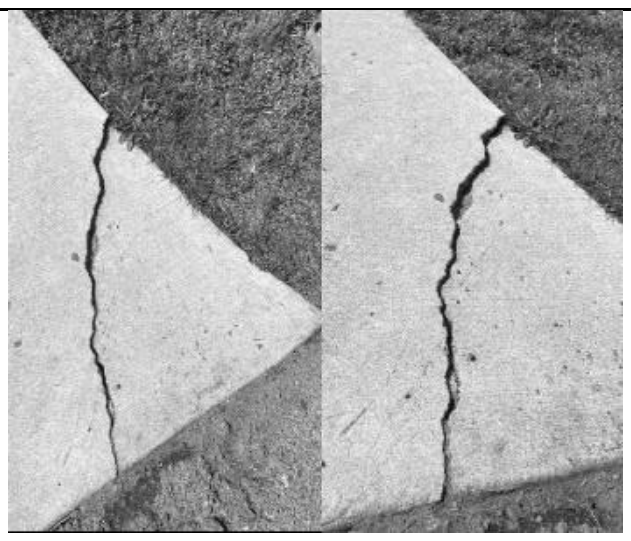 | 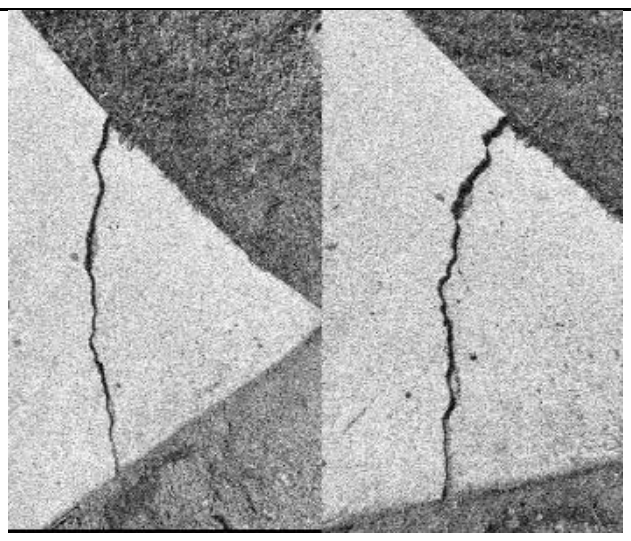 | 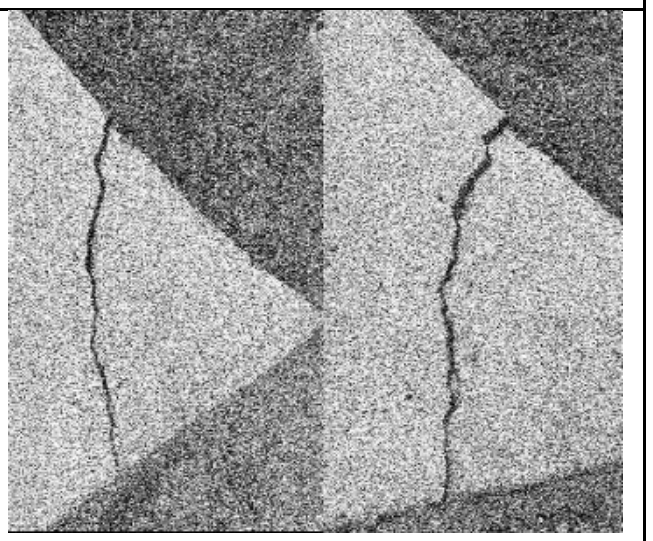 |
| SIFT | 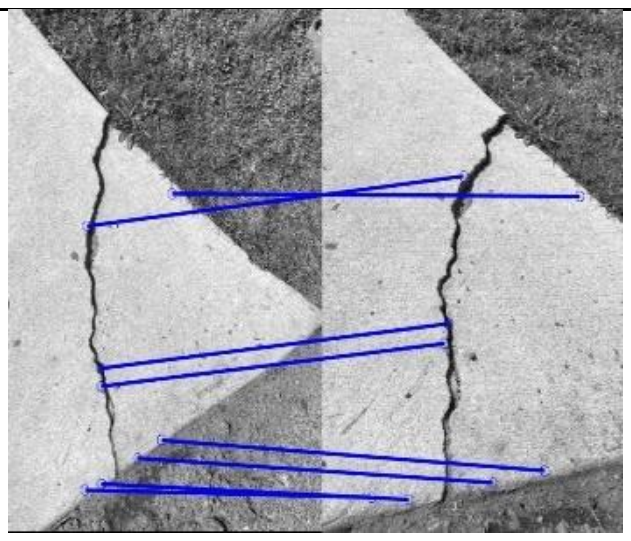 | 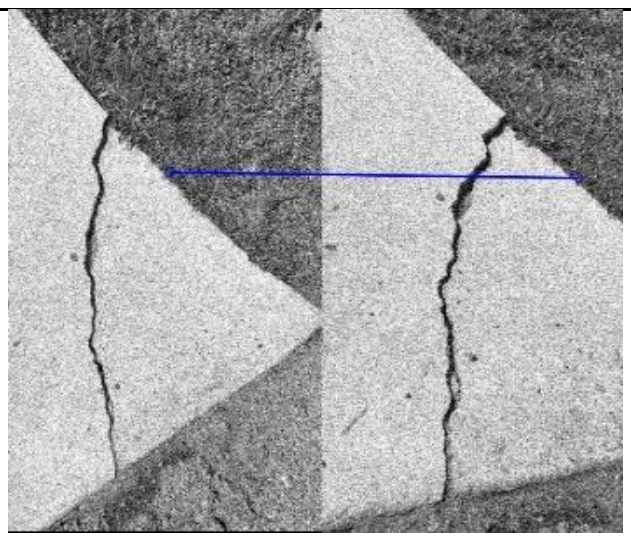 | 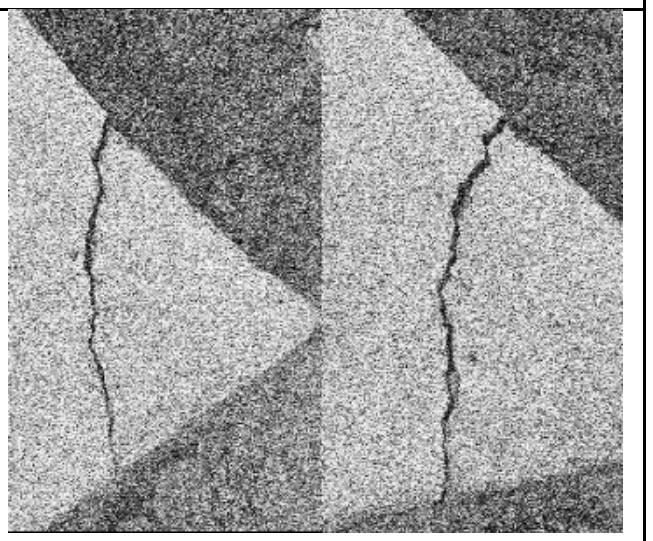 |
| SURF | 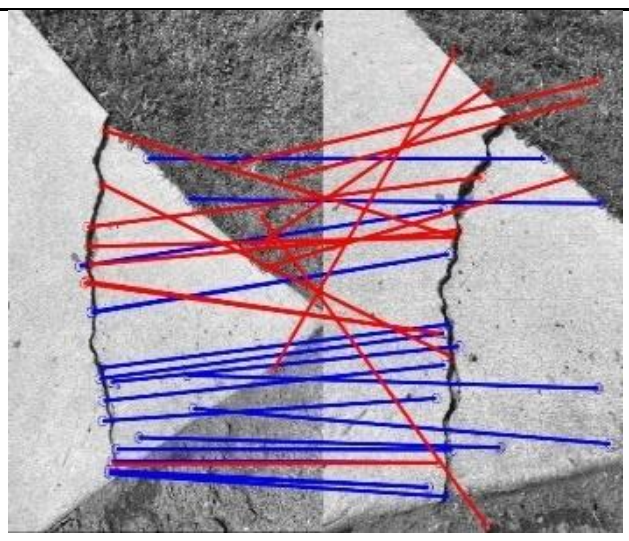 | 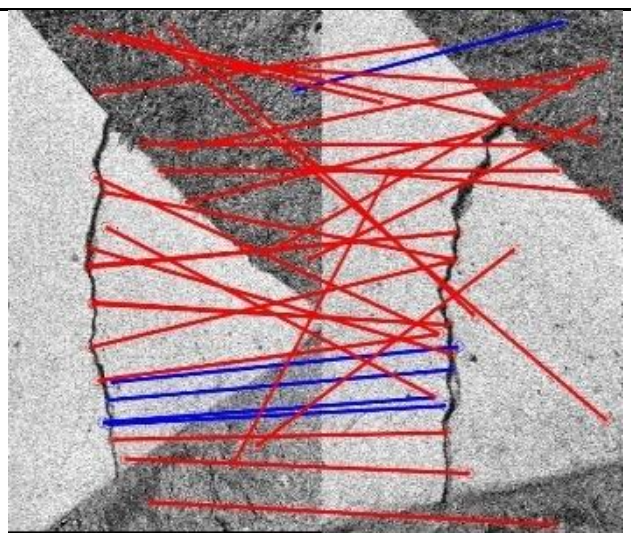 | 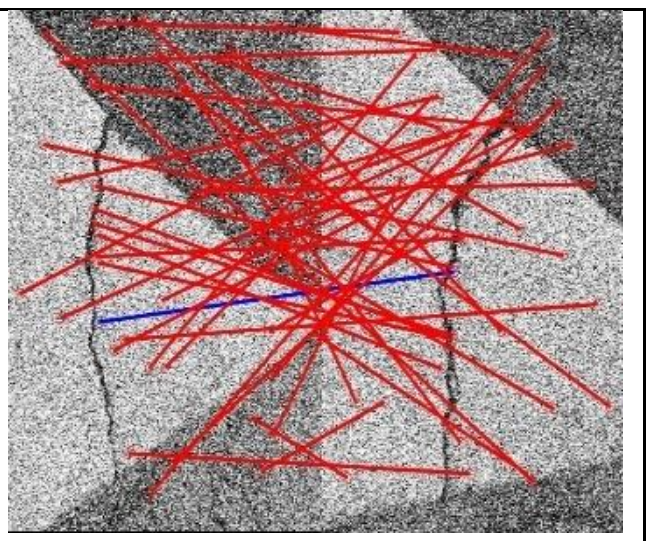 |
| ORB | 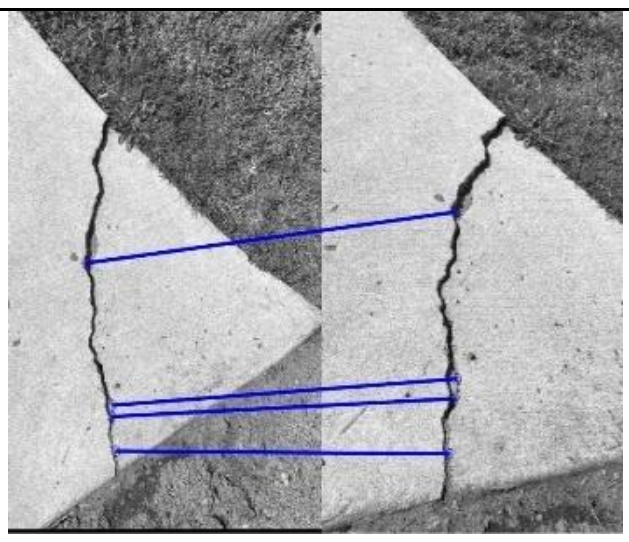 | 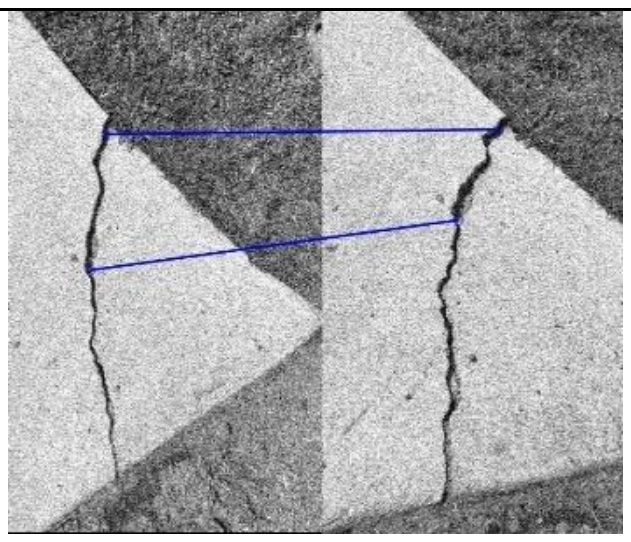 | 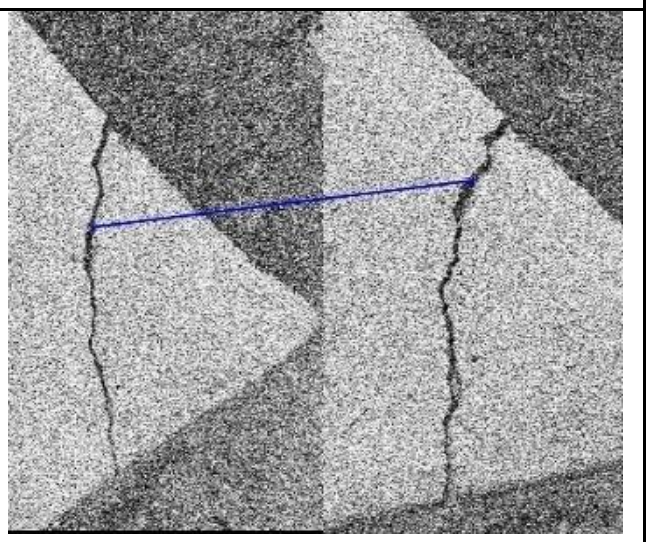 |
| BRISK | 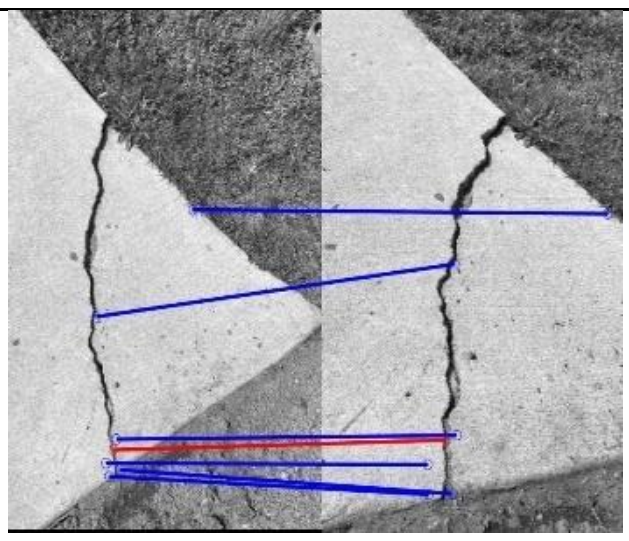 | 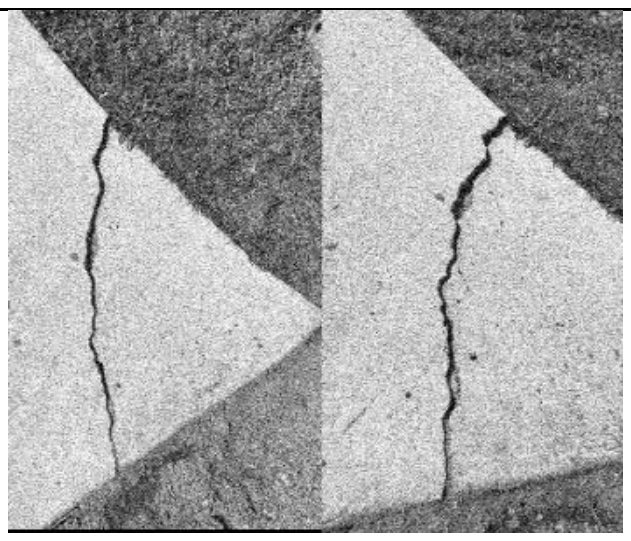 | 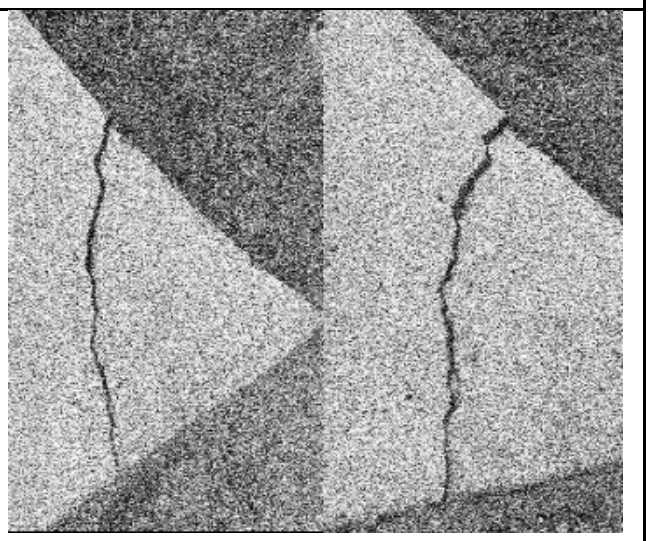 |
| KAZE | 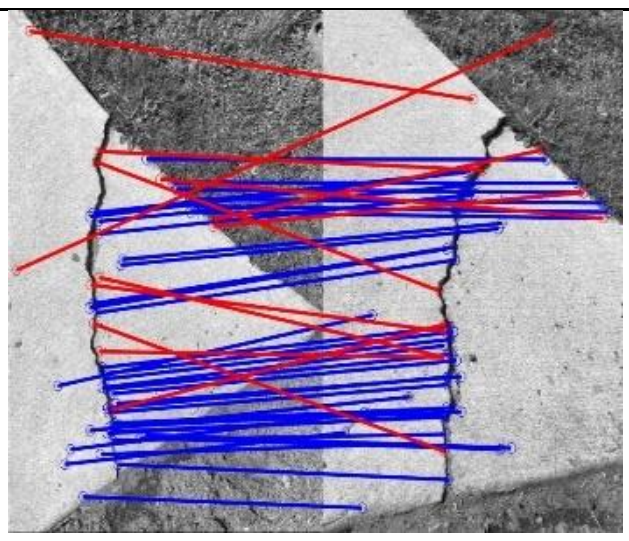 | 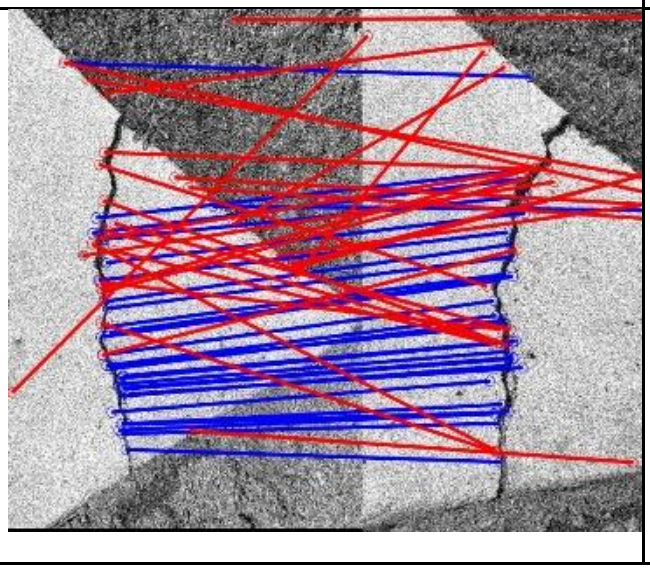 | 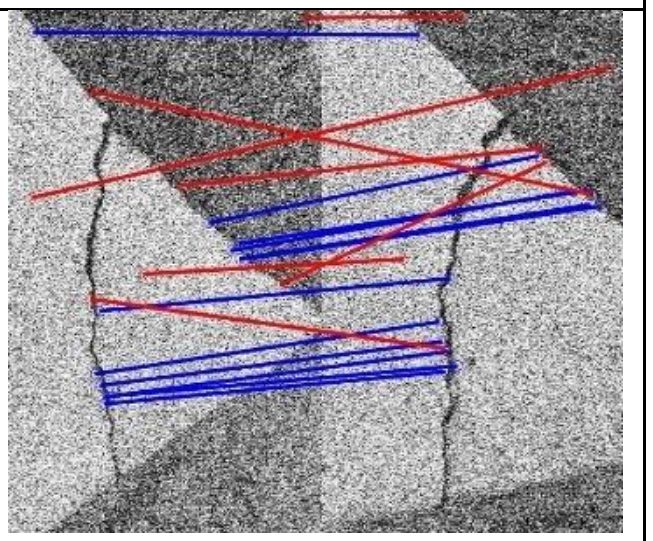 |

Table 2.6: Inlier Count Comparison Under Noise Variations

| Noise \ inliers | SIFT | SURF | ORB | BRISK | KAZE |
|---|---|---|---|---|---|
| Low | 8 | 18 | 4 | 6 | 41 |
| Medium | 1 | 5 | 2 | 0 | 31 |
| High | 0 | 1 | 1 | 0 | 12 |

Table 2.7: Visual Feature Matching Results Under Low, Medium, and High Blur Conditions

| | Low Blur | Medium Blur | High Blur |
|---|---|---|---|
| Image | 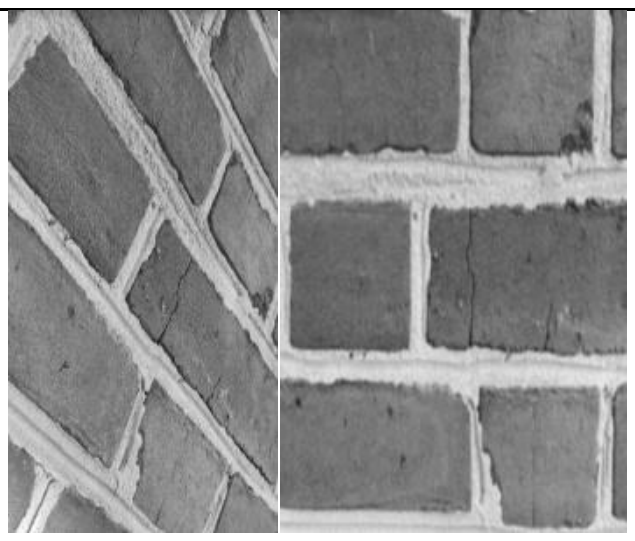 | 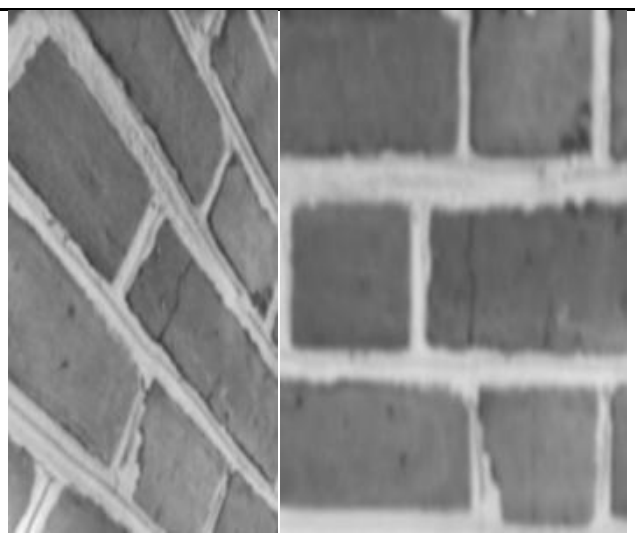 | 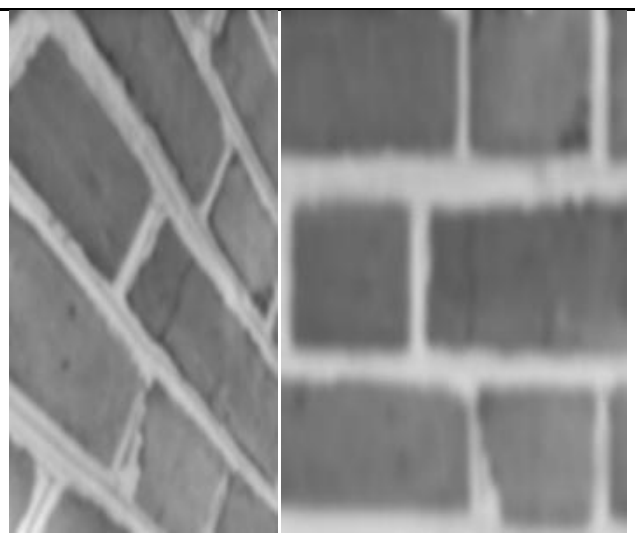 |
| SIFT | 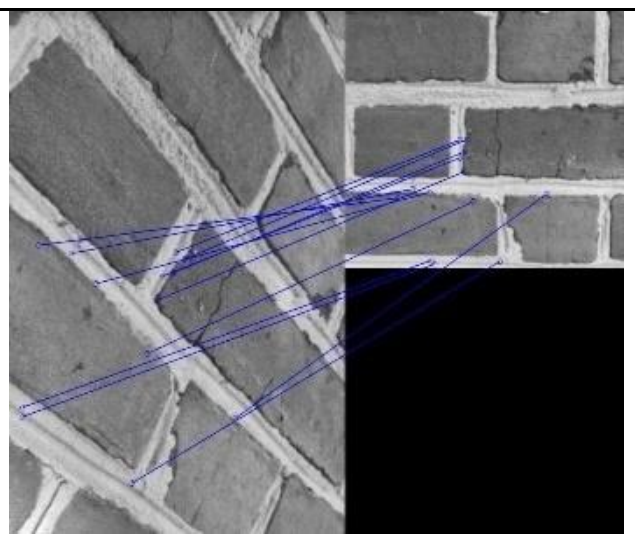 | 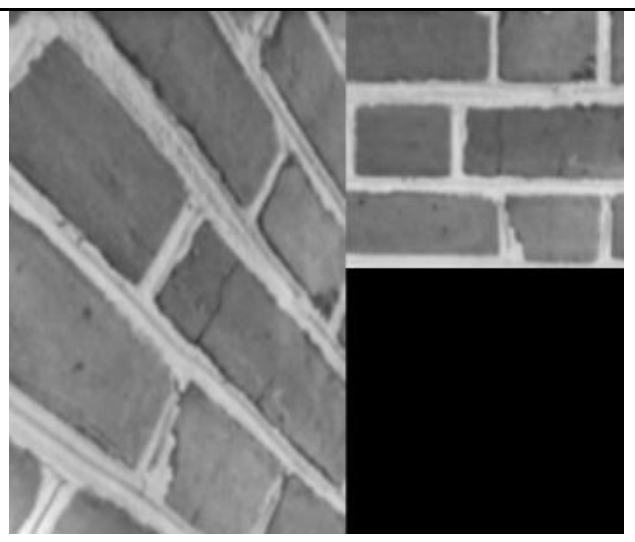 | 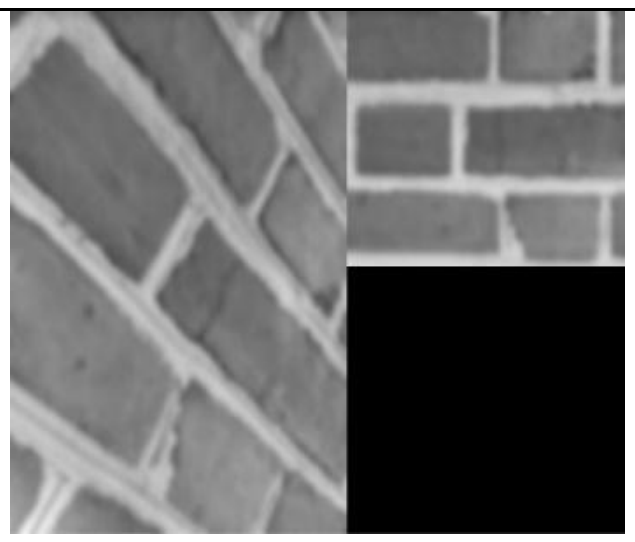 |
| SURF | 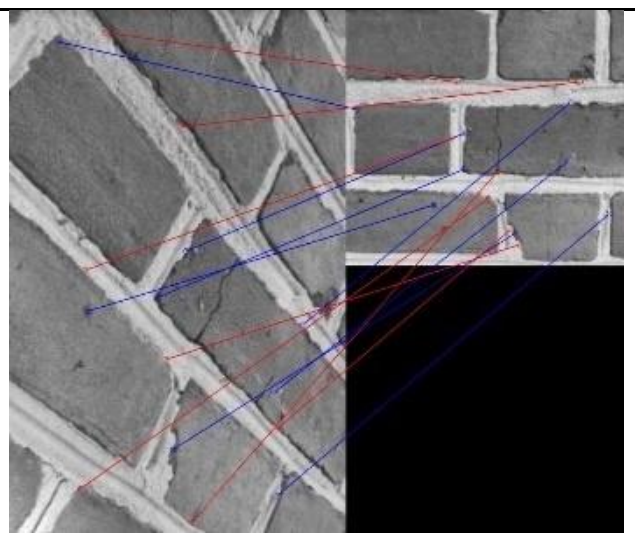 | 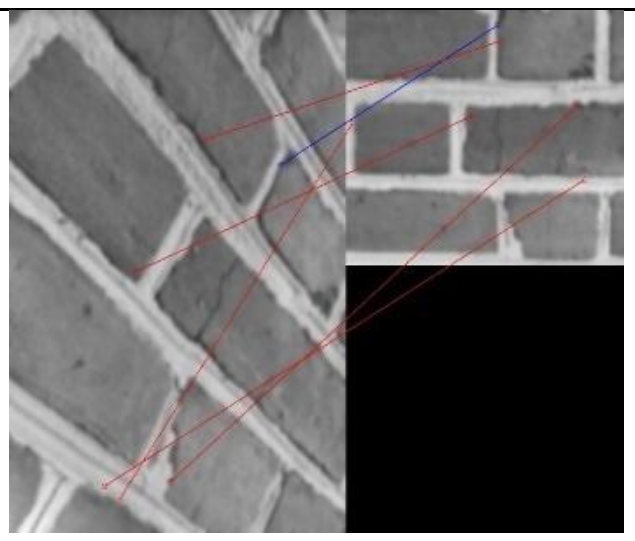 | 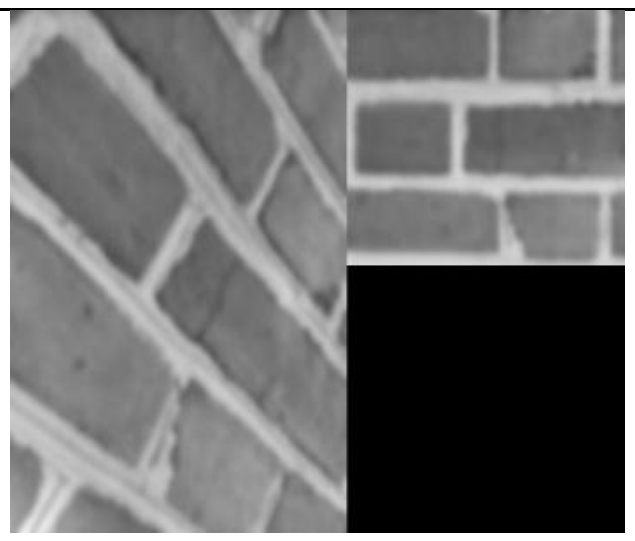 |
| ORB | 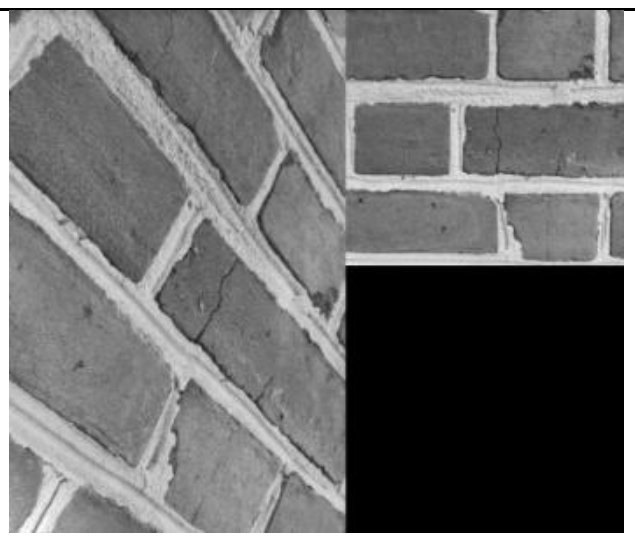 | 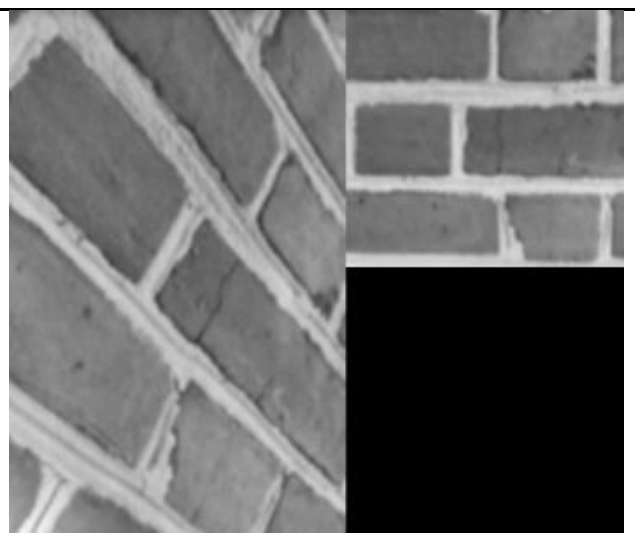 | 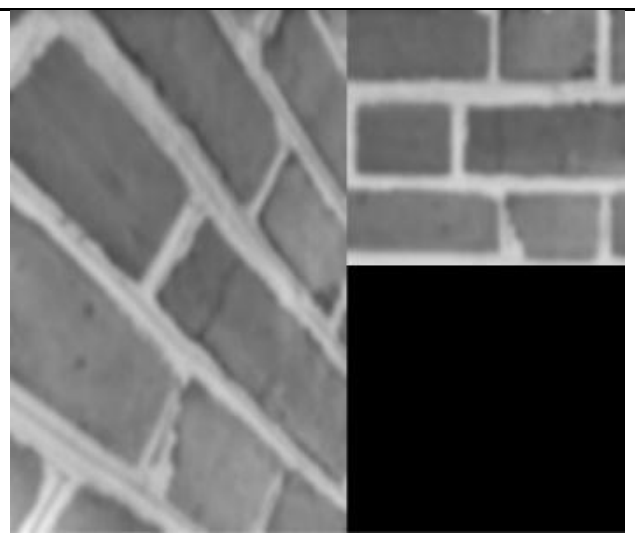 |
| BRISK | 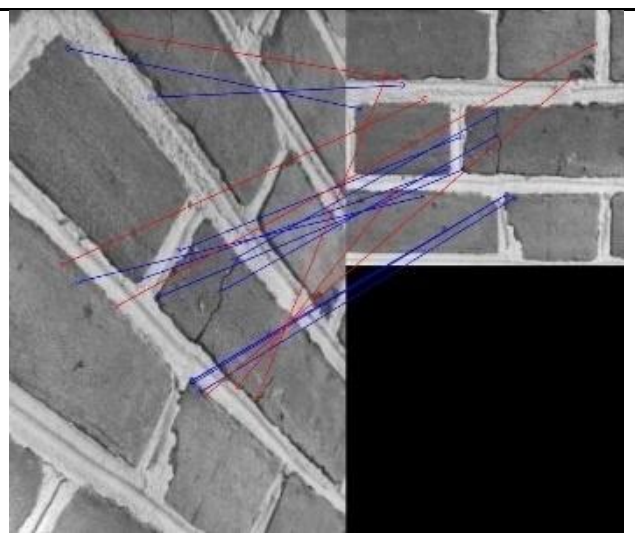 | 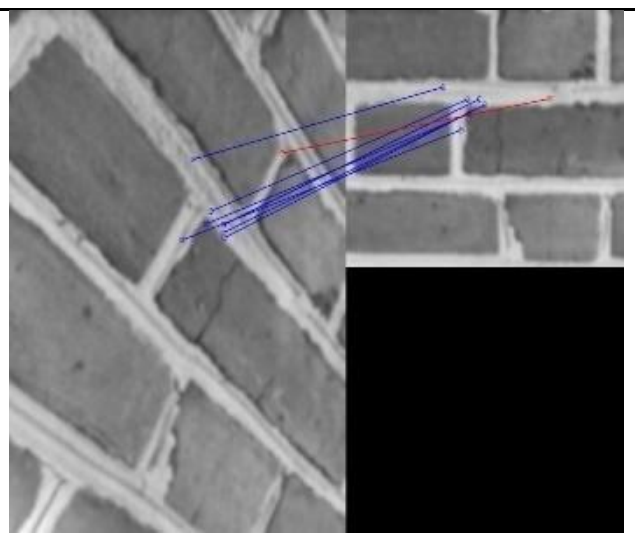 | 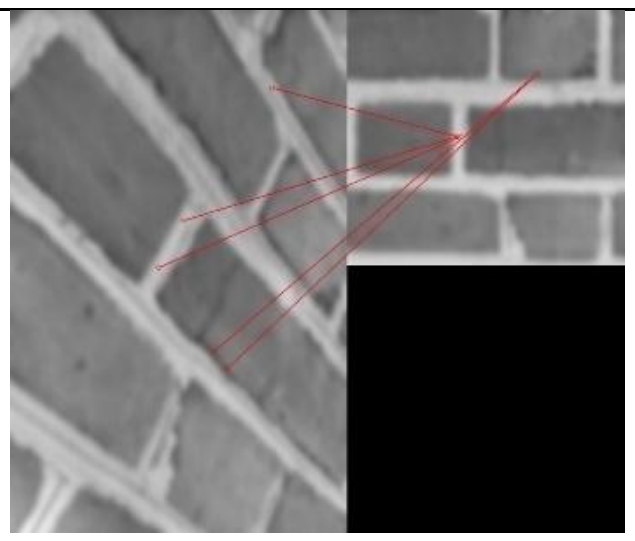 |
| KAZE | 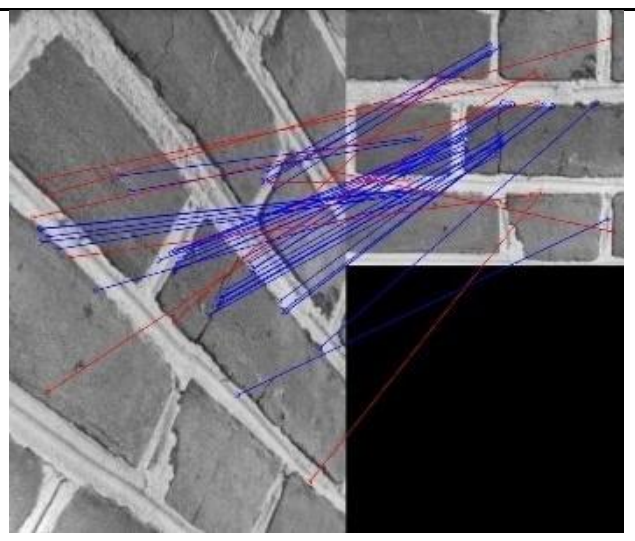 | 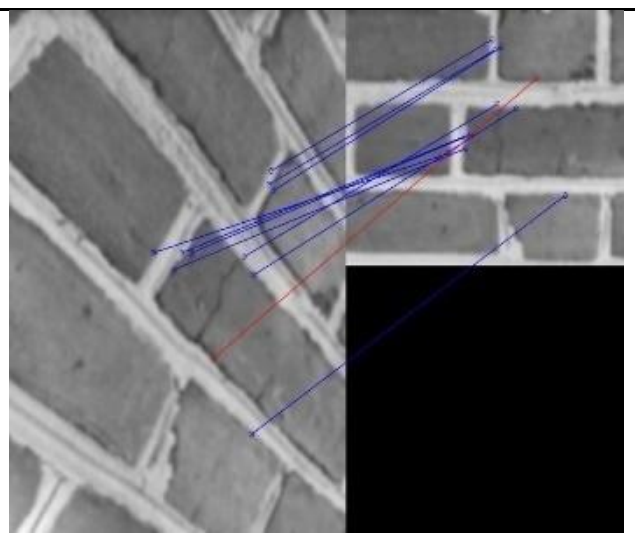 | 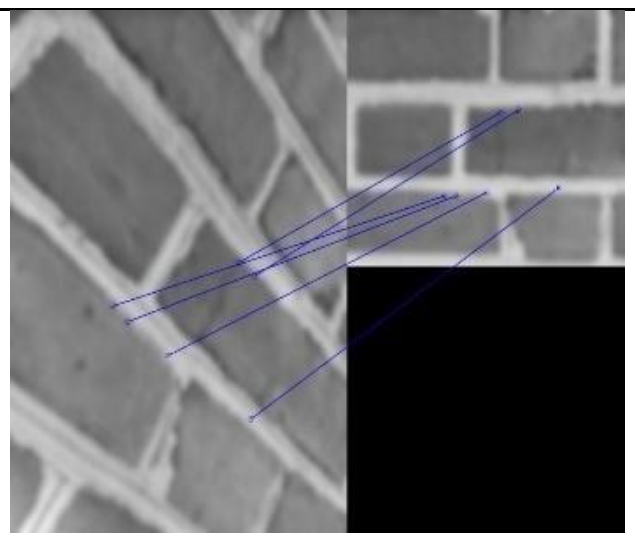 |

Table 2.8: Inlier Count Comparison Under Blur Variations

| Blurring \ inliers | SIFT | SURF | ORB | BRISK | KAZE |
|---|---|---|---|---|---|
| Low | 13 | 8 | 0 | 12 | 36 |
| Medium | 0 | 1 | 0 | 8 | 10 |
| High | 0 | 0 | 0 | 0 | 6 |

Table 2.9: Visual Feature Matching Results Under Mild, Medium, and Severe Perspective Distortion

| Distortion | Mild | Medium | Severe |
|---|---|---|---|
| Image | 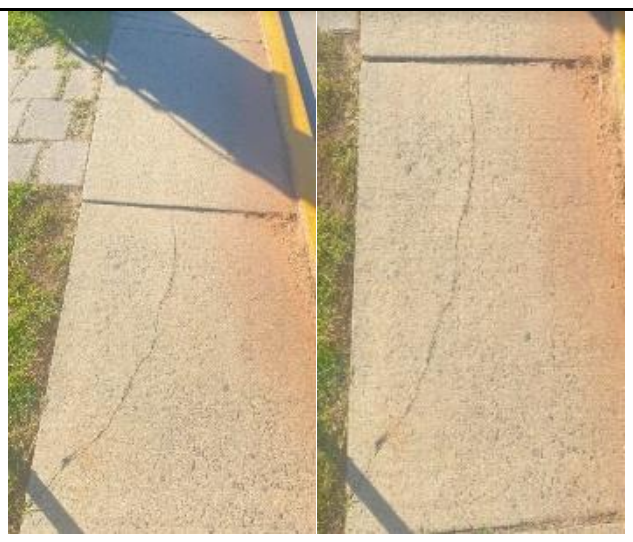 | 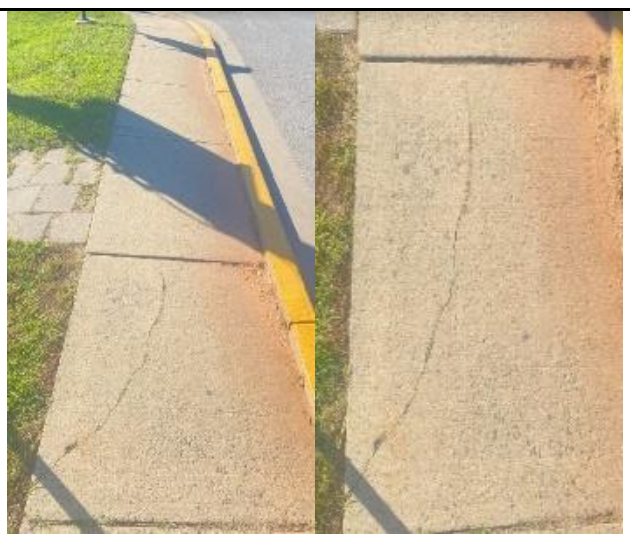 | 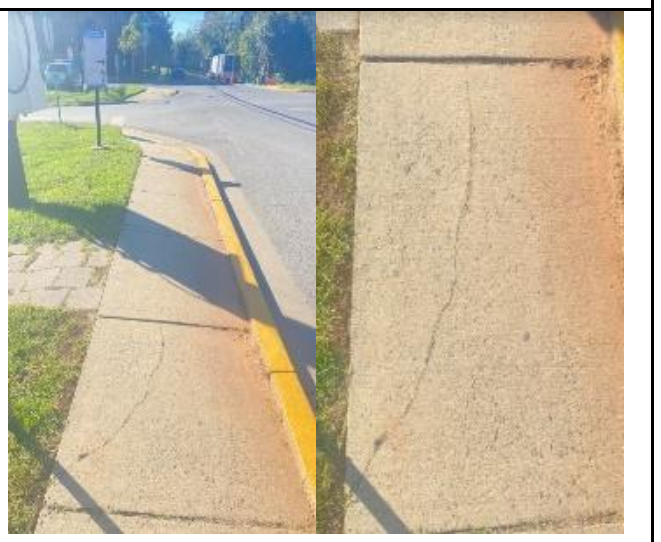 |
| SIFT | 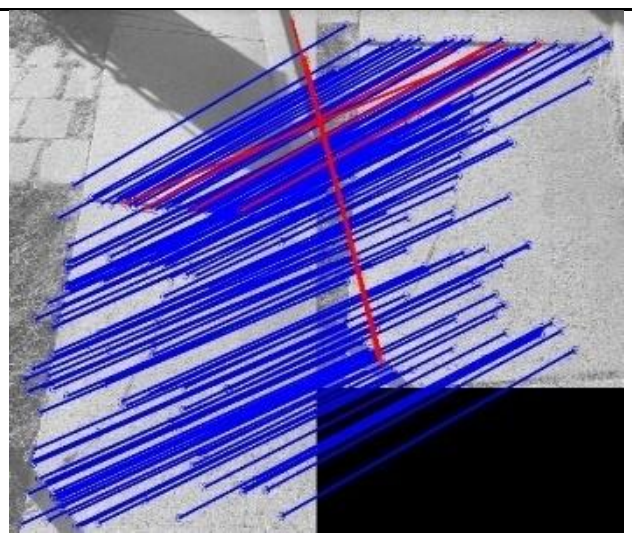 | 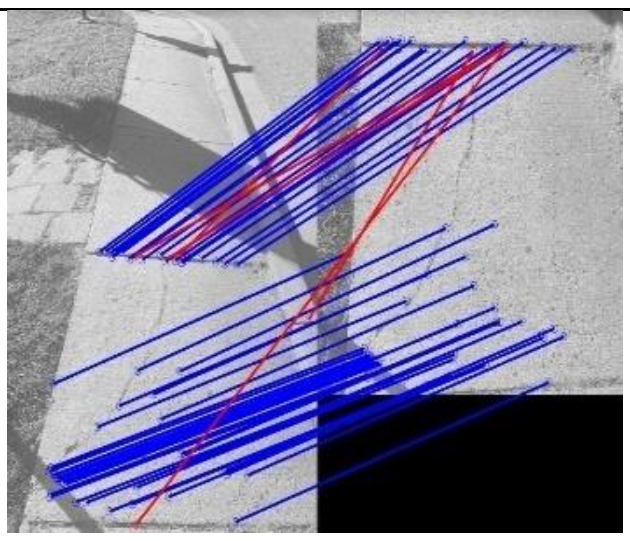 | 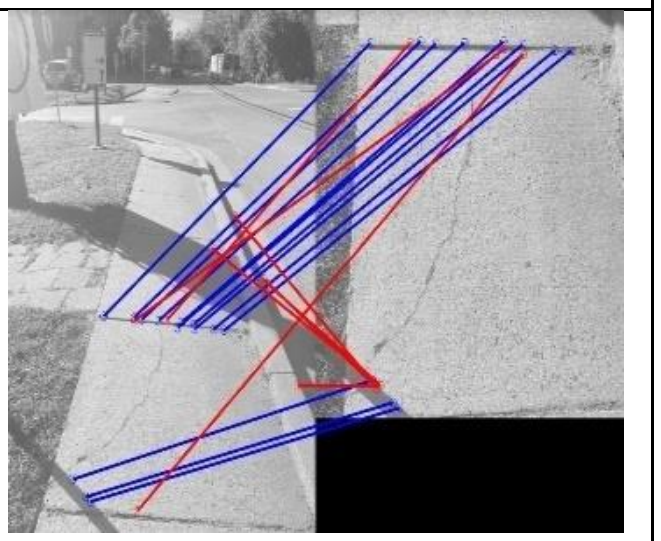 |
| SURF | 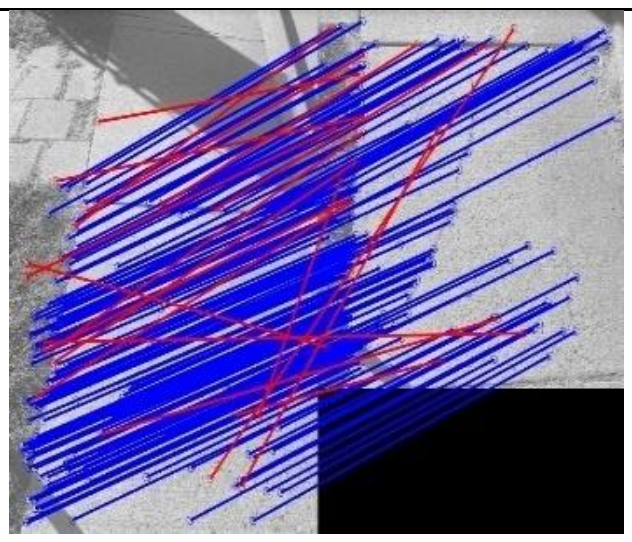 | 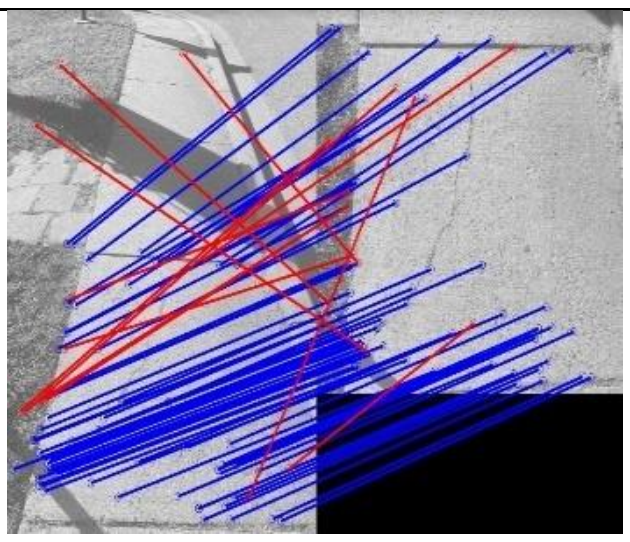 | 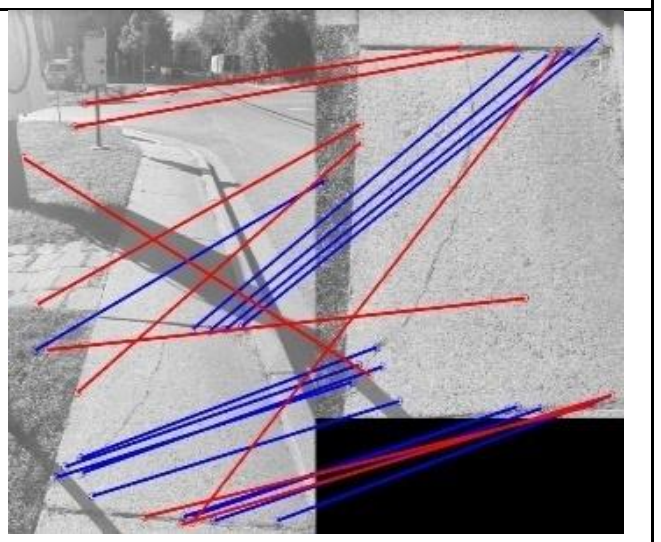 |
| ORB | 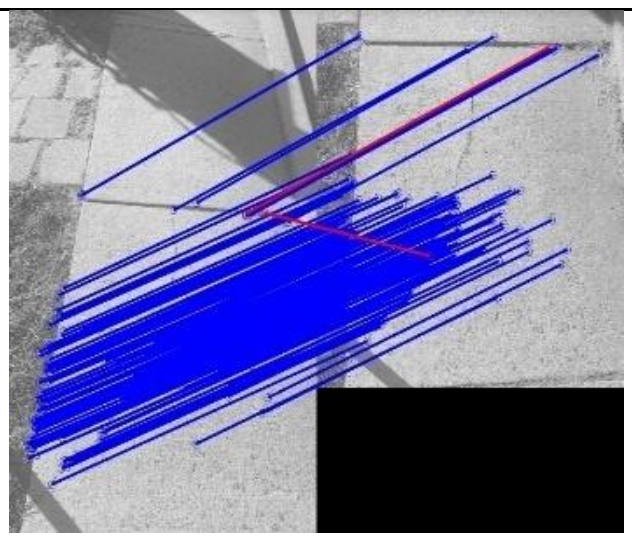 | 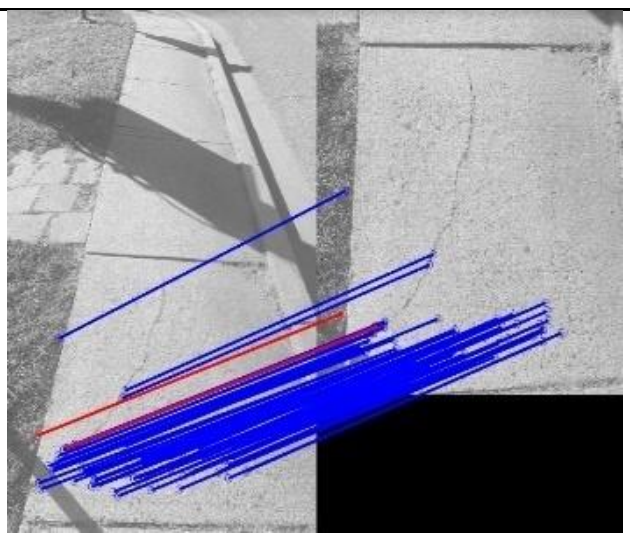 | 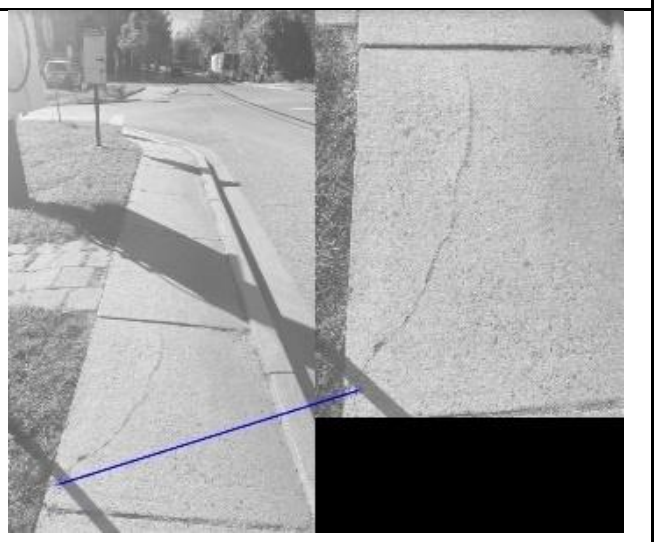 |
| BRISK | 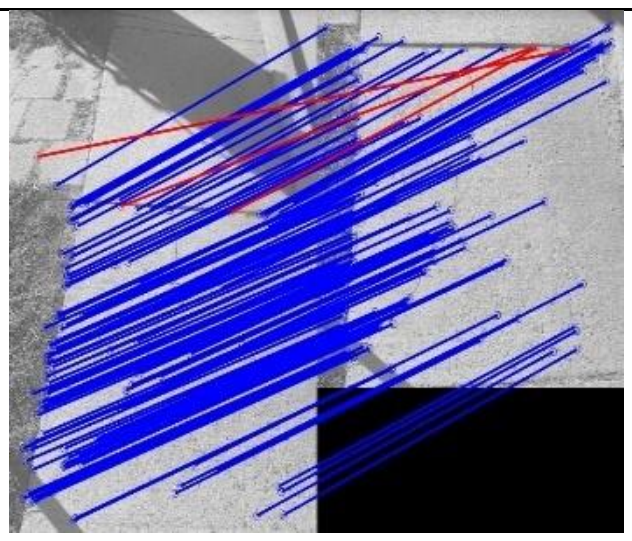 | 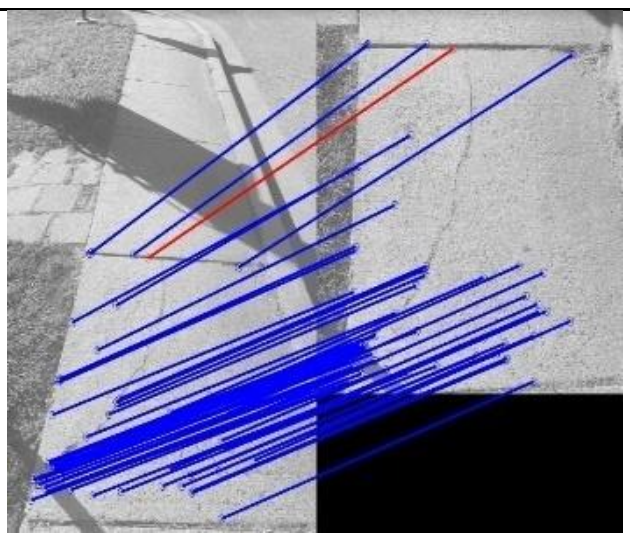 | 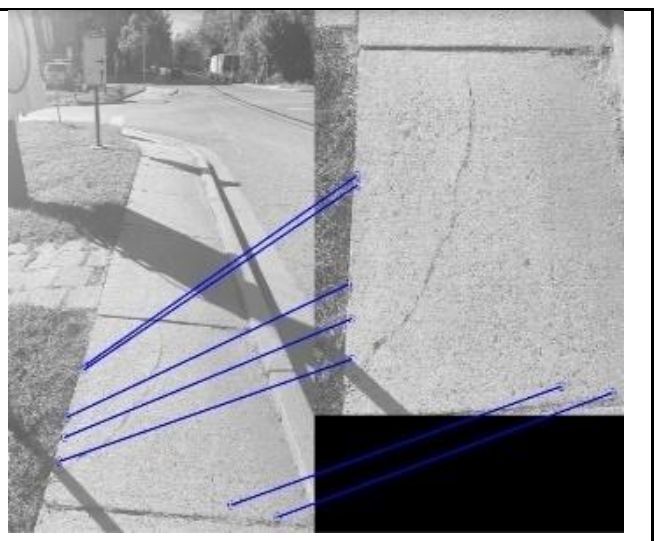 |
| KAZE | 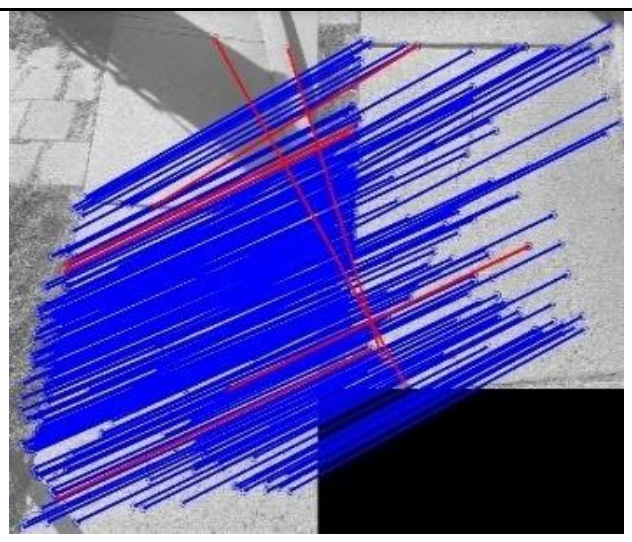 | 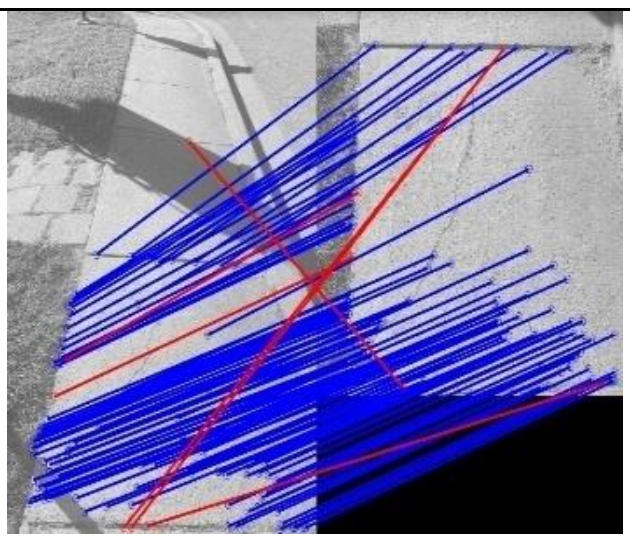 | 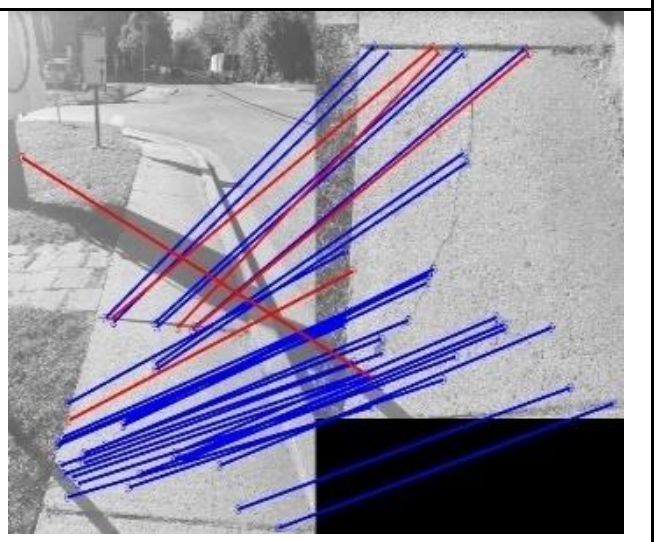 |

Table 2.10: Inlier Count Comparison Under Perspective Distortion Levels

| Distortion \ Inliers | SIFT | SURF | ORB | BRISK | KAZE |
|---|---|---|---|---|---|
| Severe | 14 | 14 | 1 | 7 | 32 |
| Medium | 57 | 64 | 62 | 53 | 125 |
| Mild | 146 | 152 | 154 | 149 | 263 |

Across all test scenarios, KAZE consistently outperformed SIFT, SURF, ORB, and BRISK in both the number of inliers and the visual quality of keypoint matching. Notably, under low-texture and low-contrast conditions—where Gaussian-based detectors typically fail—KAZE preserved more crack-related features and demonstrated superior geometry-consistent alignment. These improvements directly support accurate alignment across time-lapse sequences, enabling reliable crack evolution tracking under real-world SHM conditions. These results demonstrate KAZE's suitability for SHM under uncontrolled field conditions. To leverage these results for practical alignment, the following section describes the homography estimation process using a robust RANSAC-based pipeline

## 2.5 Outlier Elimination and Homography Estimation

Despite the robustness of KAZE in detecting distinctive keypoints, mismatches and outliers can still occur—especially in complex field environments with shadows, occlusions, or dynamic textures. These erroneous correspondences degrade homography estimation accuracy and may lead to misalignments that compromise the reliability of crack tracking. Such misalignments, if uncorrected, can result in false crack growth interpretations or drift in longitudinal assessments.

To address this, we integrate the RANSAC algorithm [24] to filter out incorrect matches and retain only inliers—correspondences that conform to a consistent geometric model [25], [26], [27], [28]. By iteratively estimating and refining homography transformations, RANSAC enhances the robustness and reliability of alignment in SHM image sequences.

### 2.5.1 Mathematical Formulation of Homography Estimation

A homography matrix $H$ models a projective transformation between two image planes. Given a point $\mathrm{X}_1 = [x_1 \quad y_1 \quad 1]^T$ in the source image, the corresponding point in the target image is:

$$\mathrm{X}_2 = H \cdot \mathrm{X}_1 \tag{12}$$

Where $H$ is a 3×3 matrix:

$$H = \begin{bmatrix} h_{11} & h_{12} & h_{13} \\ h_{21} & h_{22} & h_{23} \\ h_{31} & h_{32} & h_{33} \end{bmatrix} \tag{13}$$

Since projective transformations are defined up to scale, $h_{33}$ is typically normalized to 1, yielding 8 degrees of freedom.

### 2.5.2 Linear System Formulation and SVD

To compute $H$, a linear system is constructed based on $n$ pairs of corresponding points. Let $h$ be the 9-element vector reshaped from $H$, The system becomes:

$$A \cdot h = 0 \tag{14}$$

where $A \in R^{2n \times 9}$ is constructed from the point correspondences. This homogeneous system is solved via Singular Value Decomposition (SVD) [29]:

$$A = U \Sigma V^T \tag{15}$$

The optimal solution for $h$ is the last column of $V$, corresponding to the smallest singular value and minimizing algebraic error in a least-squares sense.

### 2.5.3 Detailed Steps of RANSAC-Based Outlier Elimination

RANSAC identifies and excludes outliers via the following steps:

- Step 1: Random Sampling

Randomly select a subset of matched keypoints. In this study, 10 correspondences per iteration are used. Sampling more than the minimum (4 for homography) improves robustness.

- Step 2: Homography Estimation

Estimate a candidate homography using the Direct Linear Transform (DLT) method described in 2.5.2.

- Step 3: Inlier Classification via Reprojection Error

For each correspondence, calculate the reprojection error:

$$error_i = \sqrt{(x_2 - {x_2}^{Projected})^2 + (y_2 - {y_2}^{Projected})^2} \quad (16)$$

A point is considered an inlier if its error falls below a threshold $\sigma$, which is dynamically updated after each iteration:

$$\sigma = \sqrt{\frac{1}{N}\sum_{i=1}^{N} {error_i}^2} \quad (17)$$

This adaptive thresholding improves tolerance to variable feature localization precision.

- Step 4: Iteration Update Based on Outlier Ratio

The number of required iterations $N_{iter}$ is adjusted dynamically based on the estimated outlier ratio $e$ and desired confidence level $p$ (e.g., 99%):

$$N_{iter} = \frac{\log(1-p)}{\log(1-(1-e)^k)} \quad (18)$$

where $k$ is the number of points used per iteration (here, $k$ =10). This dynamic strategy ensures computational efficiency while maintaining statistical robustness—particularly useful in UAV-based SHM workflows with real-time constraints.

- Step 5: Final Model Selection and Refinement

Select the homography yielding the highest inlier count. In cases of ties, models with lower total reprojection error are preferred. A final homography is then re-estimated using all inliers, followed by least-squares refinement to improve stability and minimize projection residuals.

**2.5.4 Algorithm Flowchart**

Figure 2.3 illustrates the complete RANSAC process—from sampling to inlier filtering and model refinement—providing a clear operational overview.

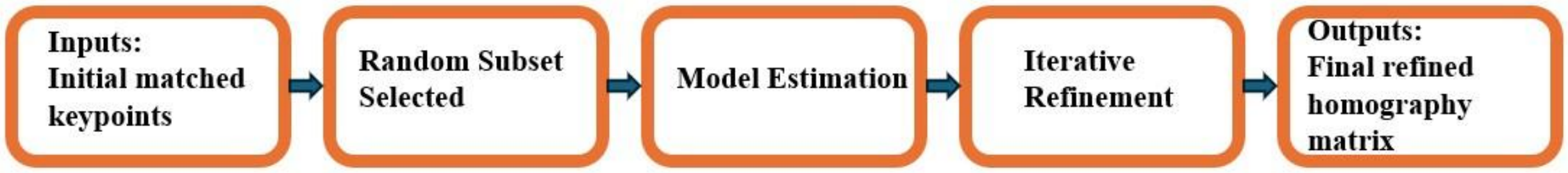

Fig. 2.3: Flowchart of the RANSAC Algorithm

This makes RANSAC especially suitable for field-based SHM, where viewpoint variation and image noise are common.

**2.6 Implementation Parameters**

All experiments in this study were conducted using MATLAB. KAZE features were extracted with the default detectKAZEFeatures function, which uses nonlinear scale-space construction via additive operator splitting (AOS).

This formulation, based on solving the anisotropic diffusion PDE, preserves edge sharpness across scales and enhances descriptor stability—particularly on low-texture concrete surfaces common in field-acquired SHM data. No manual tuning of parameters was performed to ensure reproducibility and ease of deployment.

For homography estimation, we implemented a custom RANSAC framework based on the Direct Linear Transform (DLT) and Singular Value Decomposition (SVD). The framework incorporates the following configuration:

- Sampling strategy: 10 matched keypoints per iteration. This fixed-size sampling was used for the KAZE-RANSAC pipeline to balance computational cost and robustness. Other detectors used variable sampling based on match availability.

  This value was empirically determined through extensive experiments. Across varied scenes, 10 keypoints consistently provided sufficient geometric constraint for stable homography estimation, while preserving computational efficiency. Higher values offered diminishing returns, and lower values led to unstable alignment under occlusion or low texture.

- Inlier classification: A correspondence was considered an inlier if its reprojection error was less than $\sqrt{5.99} \cdot \sigma$, where $\sigma$ was iteratively refined from the root-mean-square error of current inliers.
- Initial outlier ratio: $e = 0.5$
- Confidence level: $p = 0.99$
- Iteration count: $N$ was dynamically updated at runtime based on the estimated inlier ratio using Equation (18).
- Sigma refinement: $\sigma$ was updated after each iteration using geometric error between inlier pairs and the current homography.

This adaptive RANSAC implementation provides fine-grained control over thresholding and convergence, offering improved stability and interpretability compared to black-box estimators. It proved effective under varying degrees of noise, distortion, and keypoint sparsity—especially in UAV-based SHM scenarios where matching reliability and computational efficiency are critical.

**2.7 Algorithm Workflow**

The integration of KAZE feature detection and RANSAC-based homography estimation is systematically illustrated in Fig. 2.4. The flowchart provides a step-by-step representation of the methodology, detailing key stages including feature extraction, outlier elimination, and homography computation. This visual framework offers a concise overview of the algorithm’s structure, highlighting its sequential operations and iterative refinement process for robust geometric transformations.

Such a modular workflow not only enhances interpretability, but also facilitates reproducible deployment in SHM systems under uncontrolled field conditions.

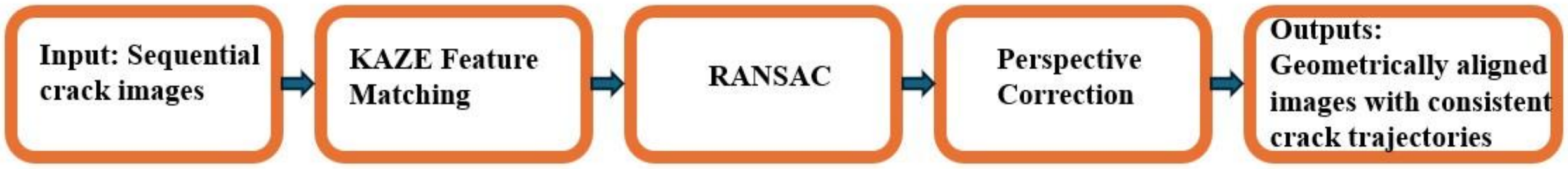

Fig. 2.4: Workflow of the Proposed Algorithm

## 3. Experiments and analysis

This section presents four scenario-driven experiments designed to evaluate the robustness and alignment accuracy of the proposed KAZE-RANSAC framework under real-world Structural Health Monitoring (SHM) conditions. The experimental validation is based on a field-acquired dataset of approximately 100 images captured across diverse concrete and masonry surfaces. For each experiment, a representative reference-target image pair is selected to simulate realistic SHM challenges—such as perspective distortion, occlusion, textured backgrounds, and dynamic lighting.

All images were acquired using the rear wide camera of an iPhone 11 (12 MP, 1/2.55″ sensor, f/1.8 aperture, 26 mm equivalent focal length), held at chest height (~1.3–1.4 m) under natural outdoor lighting without artificial enhancement. This device was chosen for its representative camera parameters among typical handheld and UAV-based SHM inspection tools, ensuring practical relevance and reproducibility.

Five feature detection algorithms—SIFT, SURF, ORB, BRISK, and KAZE—were evaluated in conjunction with RANSAC-based homography estimation. For SIFT, SURF, ORB, and BRISK, the number of keypoints sampled per iteration was adjusted based on detection output. In contrast, KAZE-RANSAC consistently employed a fixed set of 10 matched keypoints per iteration across all experiments. This uniform strategy isolates the influence of keypoint quality and ensures a fair comparison across varying field conditions.

Table 3.1: Summary of Experimental Scenarios and Associated Visual Challenges

| Experiment | Distortion Type | Interference Type | Challenge Leve |
|---|---|---|---|
| Ideal perspective | Moderate geometry | Low visual interference | Baseline |
| Cropped crack images | Severe geometry | Occlusion, truncation | Very high |
| Textured brick background | Severe geometry | High-frequency texture | High |
| Moving shadow | Moderate geometry | Dynamic lighting | High |

*Note: Challenge levels reflect a combination of geometric distortion severity and visual interference type, rather than algorithmic performance alone. The "Cropped Crack Images" scenario presented the most severe challenge, combining extreme geometric distortion with partial occlusion. Only the KAZE-RANSAC pipeline successfully recovered alignment in this setting, justifying the "Very High" difficulty rating.*

### 3.1: Ideal Perspective Correction with Clear Four-Point Correspondences

This baseline experiment evaluates alignment accuracy under moderate geometric distortion. A surface crack on a flat concrete slab was imaged from two viewpoints: vertical (top-down) and oblique. Four distinct ground anchors provided reliable reference points. The crack, representative of fine-width shrinkage damage, appeared clearly in both images without occlusion or texture interference.

All five detectors—SIFT, SURF, ORB, BRISK, and KAZE—were paired with RANSAC for homography estimation.

Perspective distortion in the oblique image caused the crack area to decrease by 45%, spine length by 41%, and average width by 9% (Table 3.2), underscoring the need for precise correction in SHM workflows.

Table 3.2: Crack Dimensions Before Perspective Correction (All units in pixels)

| Image Description | Image 1 (baseline) | Image 2 |
|---|---|---|
| Image | 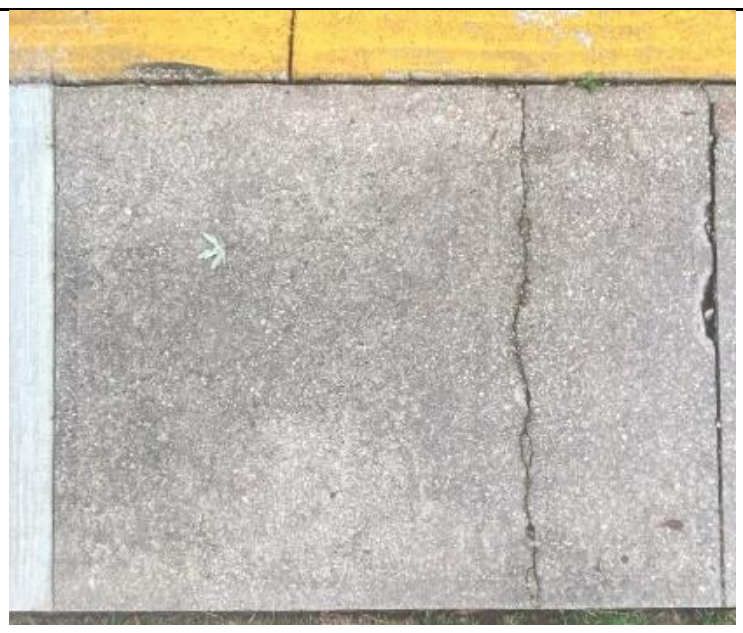 | 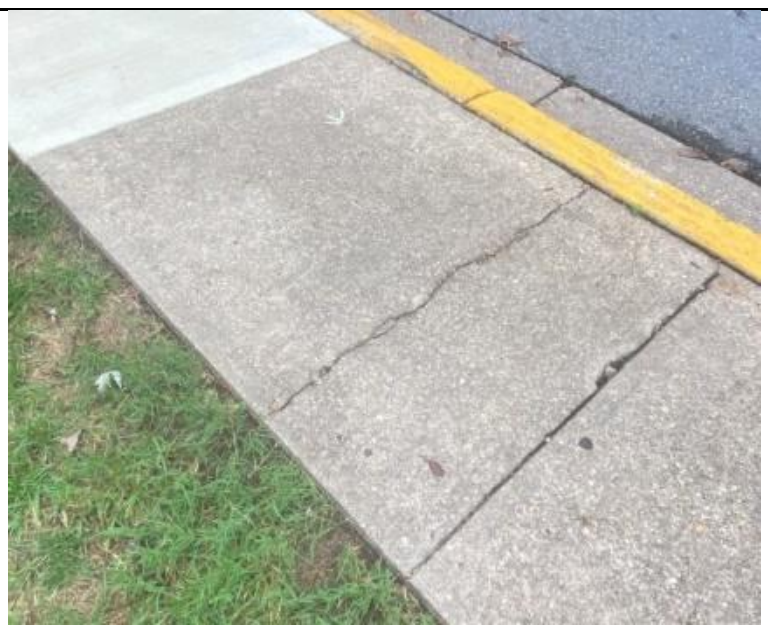 |
| Crack area | 2422 | 1335 |
| Spine length | 768 | 451 |
| Crack average width | 3.2 | 2.9 |

Following correction, Table 3.3 summarizes keypoint matching results, alignment quality, and spatial inlier distribution for each method.

Table 3.3: Keypoint Matching and Calibration Results

(a) Matching Before/After RANSAC

| | Matches Before RANSAC | Matches After RANSAC |
|---|---|---|
| SIFT | 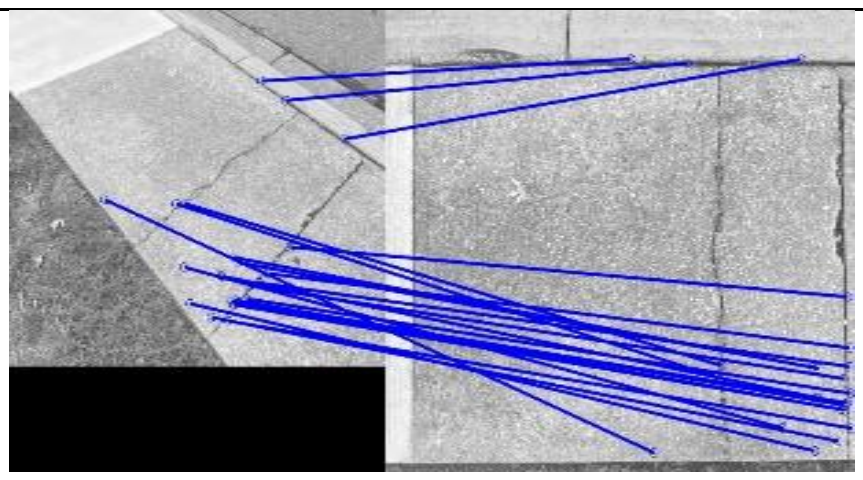 | 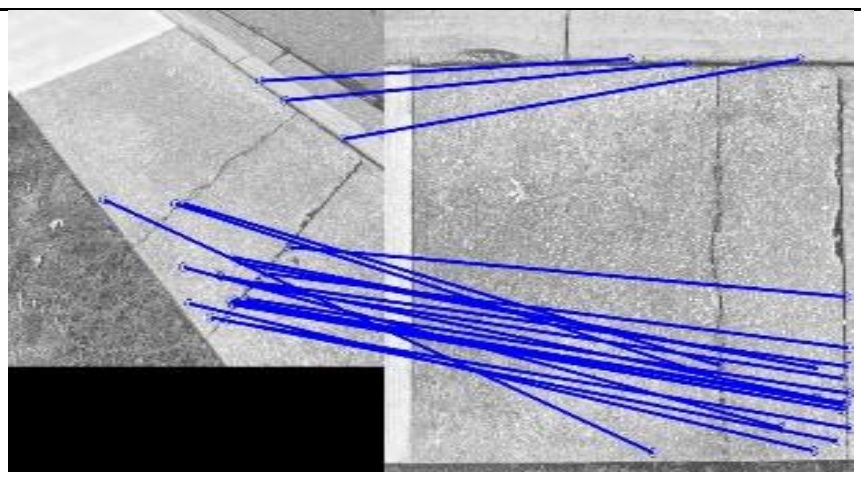 |
| SURF | 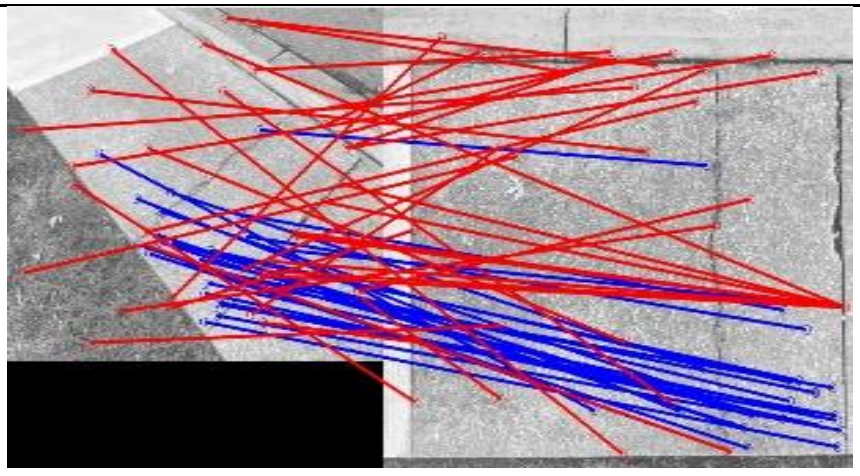 | 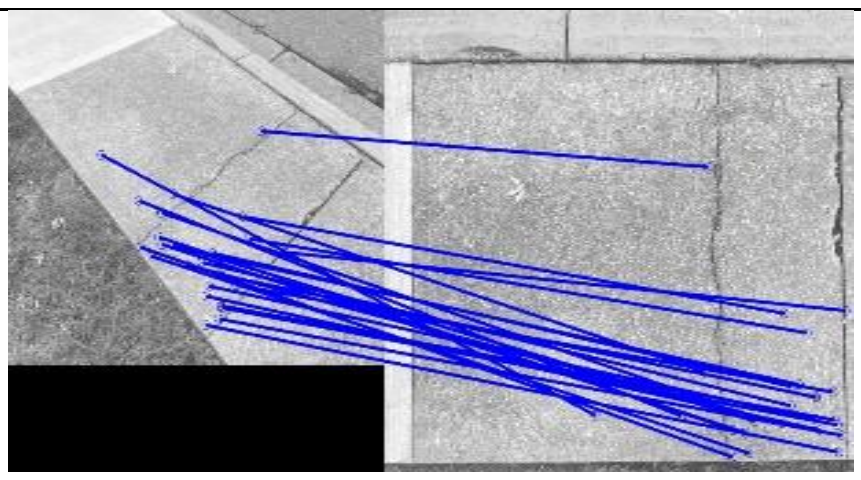 |
| ORB | 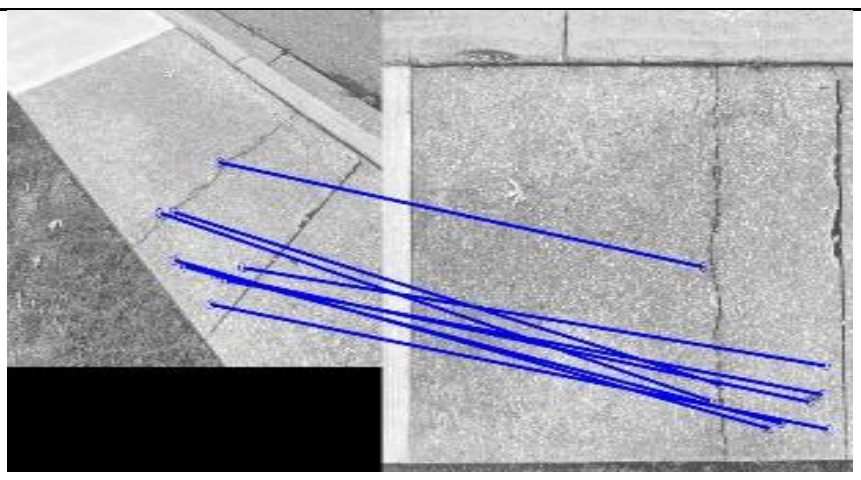 | 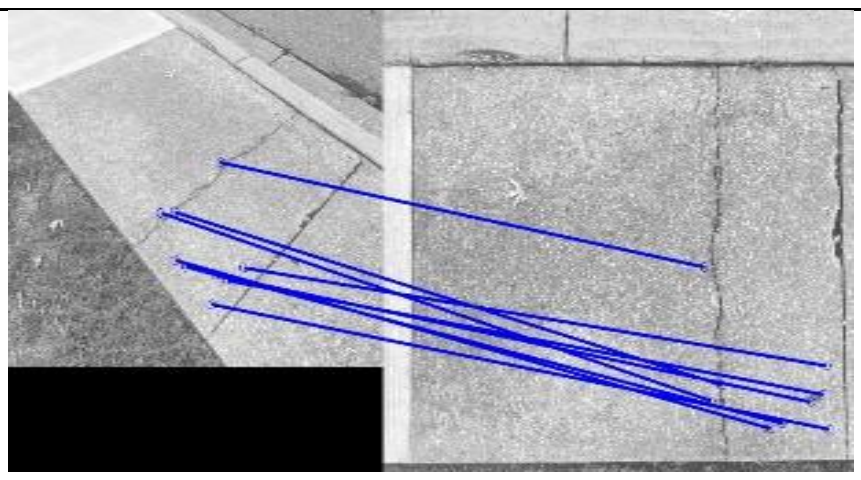 |
| BRISK | 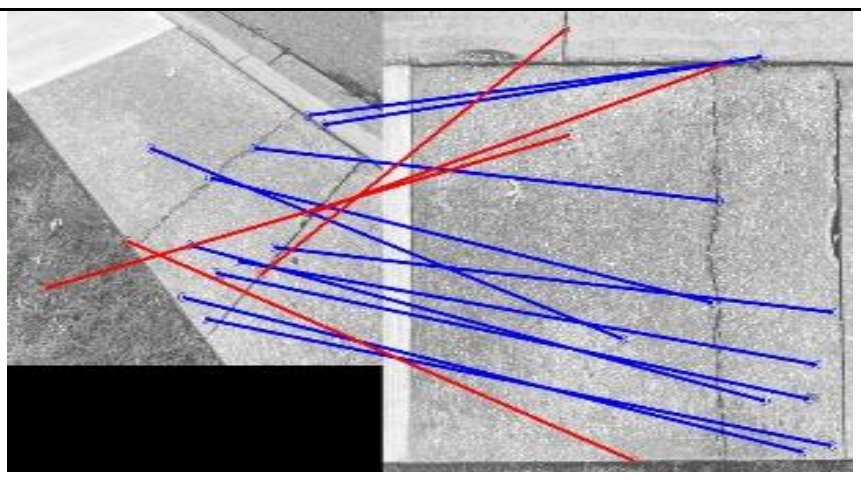 | 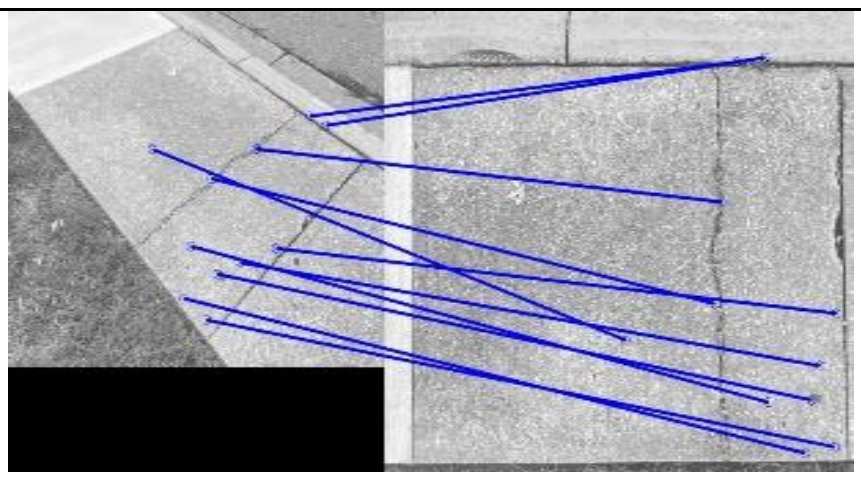 |
| KAZE | 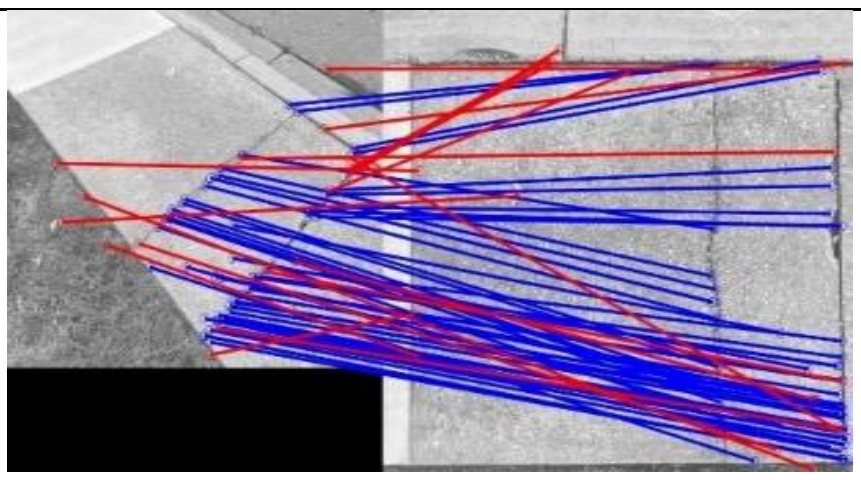 | 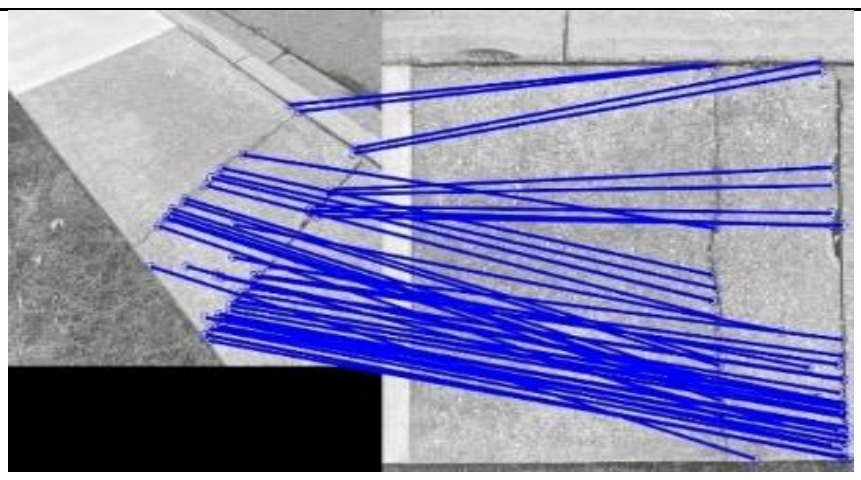 |

(b) Crack Overlap with Ground Anchors

| Method | Corrected Image 2 (Blue) | Baseline Image 1 (Red) | Overlap Accuracy |
|---|---|---|---|
| SIFT | 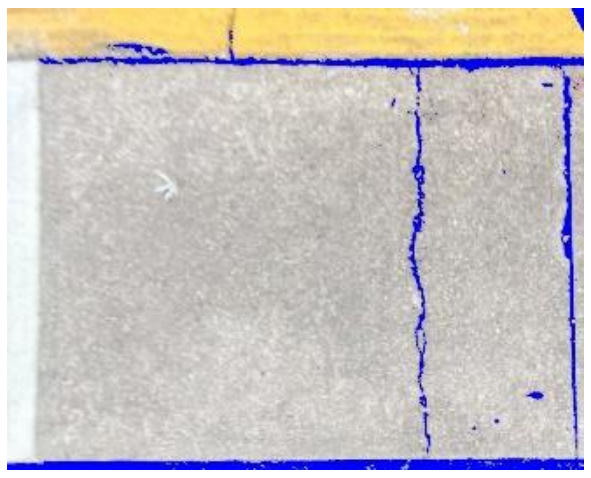 | 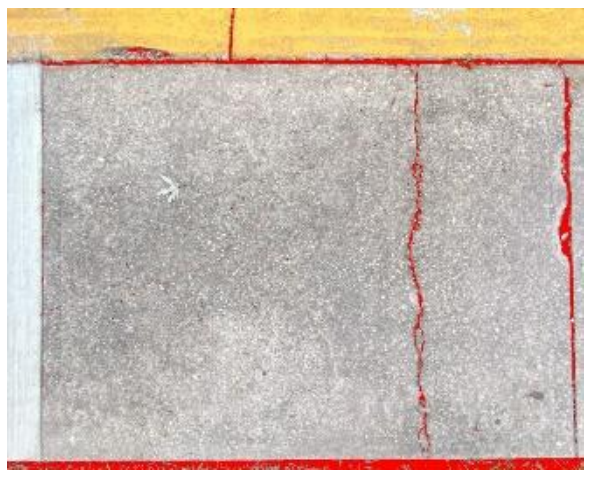 | 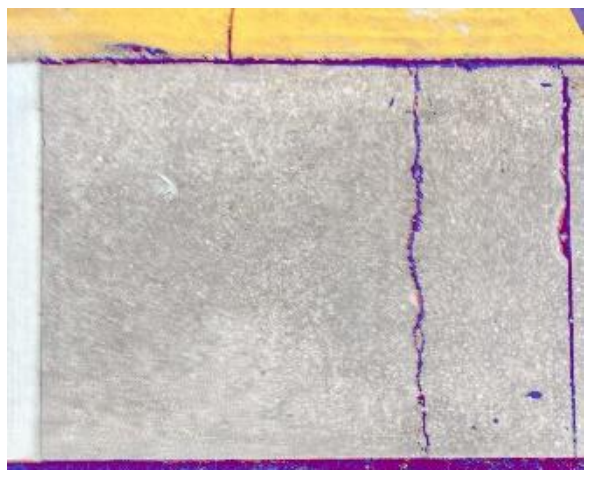 |
| SURF | 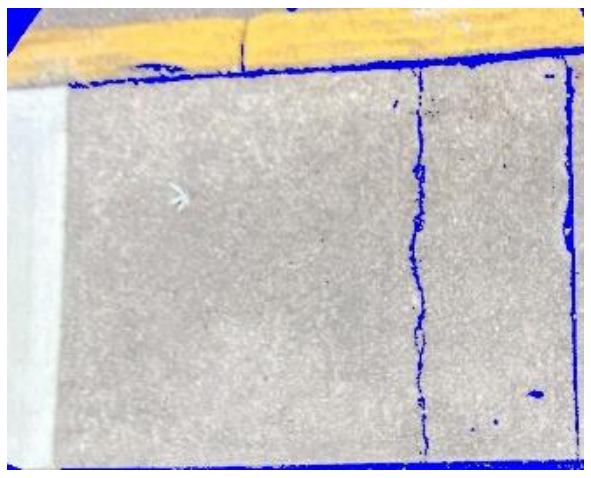 | 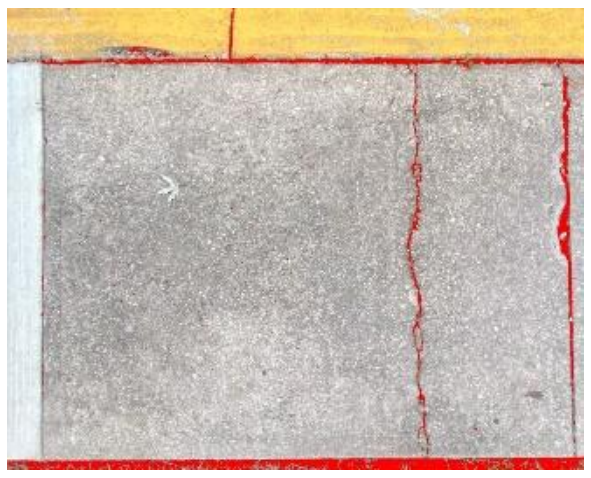 | 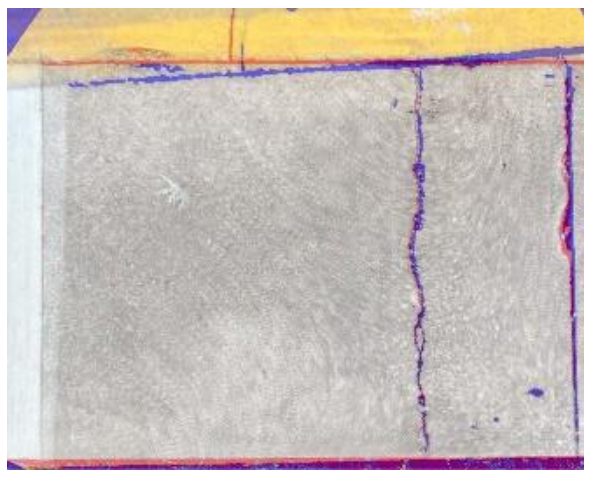 |
| ORB | 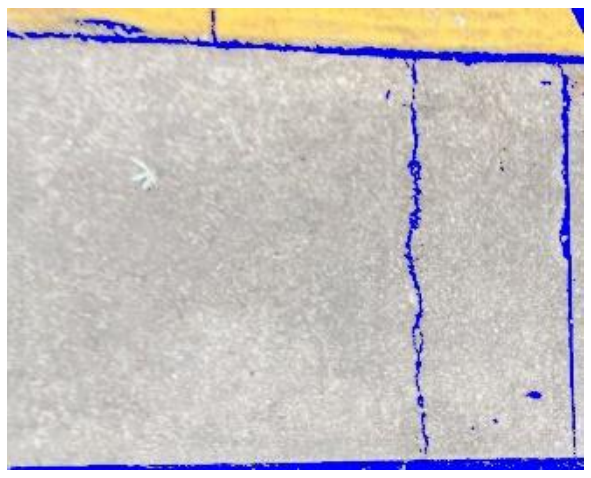 | 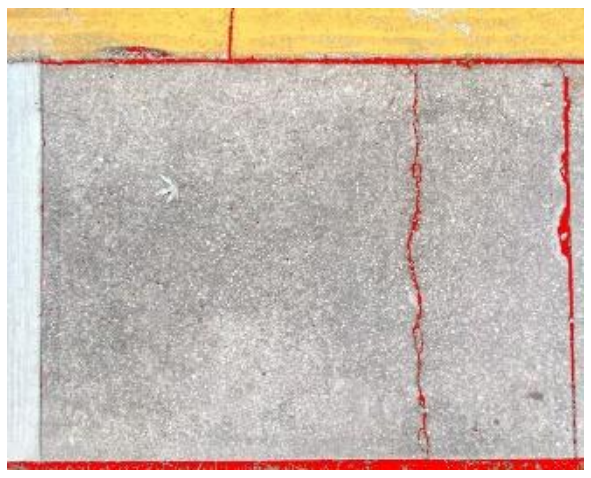 | 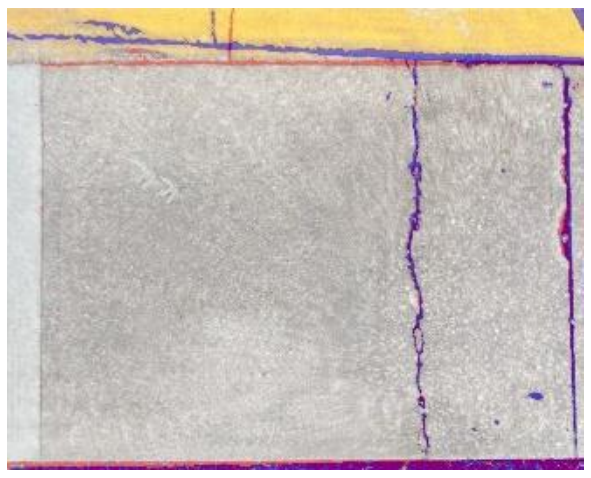 |
| BRISK | 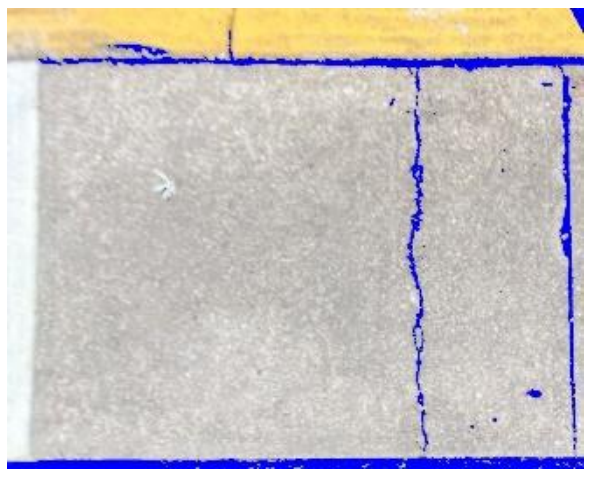 | 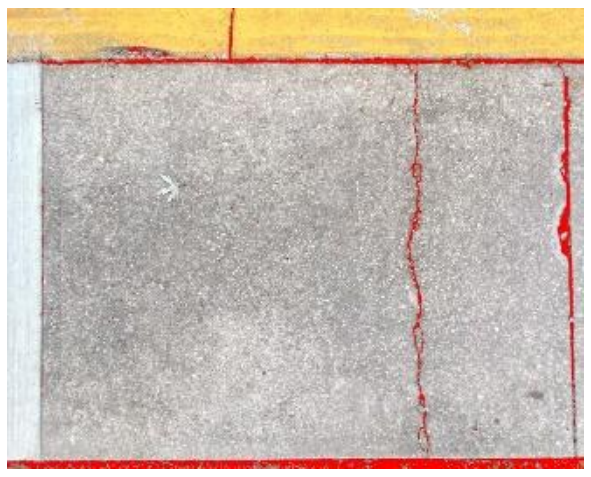 | 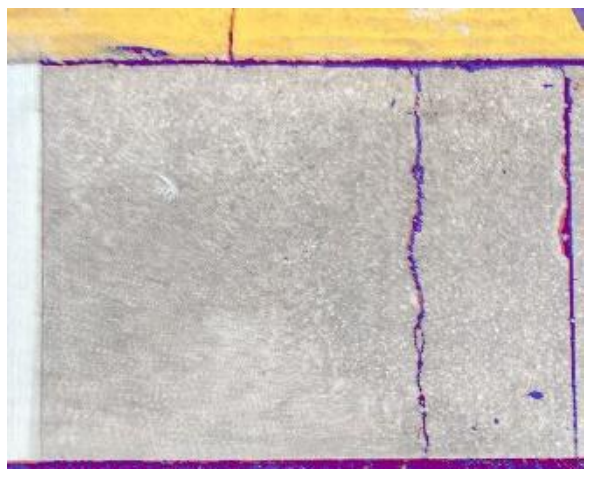 |
| KAZE | 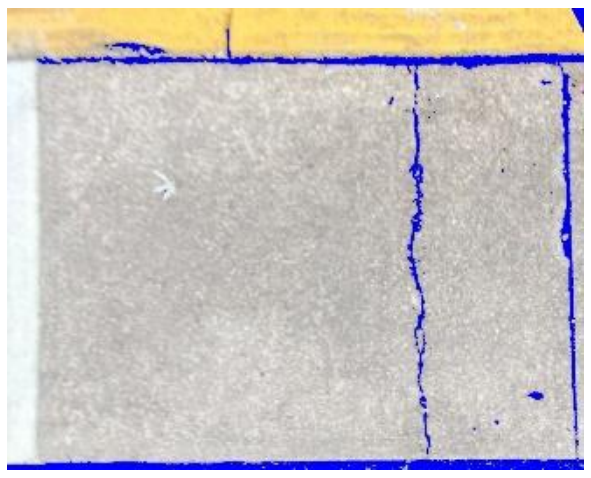 | 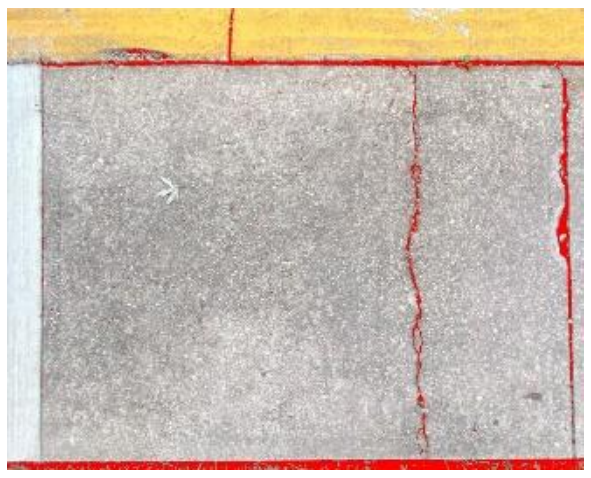 | 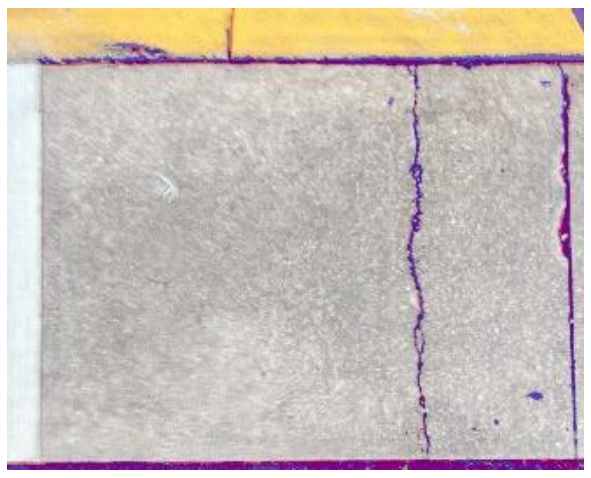 |

(c) Dimensional Accuracy vs. Baseline (pixels and % error)

| | Baseline | SIFT | SURF | ORB | BRISK | KAZE |
|---|---|---|---|---|---|---|
| Crack Area | 2422 | 2532 | 2410 | 2713 | 2463 | 2367 |
| Area Error (%) | – | 5 | 0.5 | 12 | 2 | 2 |
| Spine Length | 768 | 671 | 622 | 657 | 620 | 784 |
| Length Error (%) | – | 13 | 19 | 14 | 19 | 2 |
| Average Crack Width | 3.2 | 3.6 | 3.6 | 3.9 | 3.7 | 3.2 |
| Width Error (%) | – | 13 | 13 | 22 | 16 | 0 |

(d) Spatial Distribution of Inlier Matches

| Detector | Total Inliers | Crack-Aligned | Background | Structural Coverage | Reliability Summary |
|---|---|---|---|---|---|
| SIFT | 22 | 2 | 7 | Weak | Overconcentrated on concrete edges; low crack sensitivity |
| SURF | 23 | 3 | 18 | Very weak | Match density high but spatially misaligned; crack underrepresented |
| ORB | 9 | 3 | 6 | Moderate | Clean matches, but insufficient crack anchoring |
| BRISK | 18 | 2 | 7 | Weak | Border-biased matching with minor crack focus |
| KAZE | 59 | 13 | 6 | Strong | Focused on cracks and corner anchors; structurally meaningful |

*Note: "Background" refers to valid RANSAC inliers not located on the crack path, such as edge features or slab texture points. Corner anchors were not individually counted but contributed to total inlier counts.*

Despite moderate distortion and minimal interference, only KAZE-RANSAC achieved both geometric fidelity and complete structural alignment. Of its 59 inliers, 13 were crack-centered and several others anchored to corner features—yielding a well-constrained homography with ≤2% error across all metrics. The final overlay showed seamless alignment between crack edges and reference quadrants.

SIFT-RANSAC and BRISK-RANSAC also produced visually accurate results. Although each retained only two crack-centered inliers, their matches were well-distributed, enabling full crack overlap with minor deviations (≤5%). Their robustness stemmed from low background noise and adequate spatial anchoring.

SURF-RANSAC, despite 23 total matches, had most inliers (18) in noisy or irrelevant regions. The weak spatial distribution led to visible misalignment despite a deceptively low area error of 0.5%.

ORB-RANSAC produced only 9 matches, 3 on the crack. While valid, this sparse set failed to constrain the transformation, resulting in partial misalignment and incomplete crack recovery.

These findings highlight that in low-texture SHM environments, spatial relevance outweighs match quantity. KAZE's nonlinear scale space preserved subtle edge structures, enabling high-precision correction—outperforming classical detectors even under ideal conditions.

### 3.2: Perspective Correction for Cropped Crack Sections

This experiment evaluates alignment performance under compounded challenges involving both severe geometric distortion and partial crack visibility—conditions frequently encountered in real-world SHM applications such as constrained inspections near walls, joints, or sensor occlusions. The tested crack appeared on a concrete surface with bifurcated geometry and limited edge continuity, and was partially cropped out of frame in both reference and distorted views.

The reference image was captured with minimal distortion, while the second view introduced strong perspective skew and partial occlusion. Cracks were only partially visible and lacked strong geometric anchors. Visual clutter and the cropped field of view impaired consistent keypoint detection and alignment, making this an ideal stress test for evaluating detector robustness.

All five detectors—SIFT, SURF, ORB, BRISK, and KAZE—were tested in conjunction with RANSAC.

Table 3.4 summarizes the geometric deformation prior to correction, including a 24% drop in crack area and a 21% drop in spine length.

Table 3.4: Crack Dimensions Before Perspective Correction (All units in pixels)

| Image Description | Image 1 (baseline) | Image 2 |
|---|---|---|
| Image | 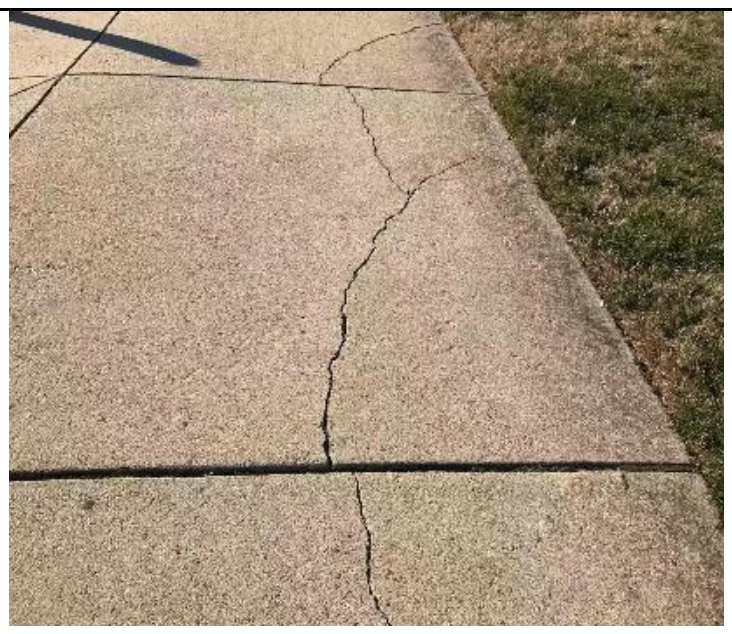 | 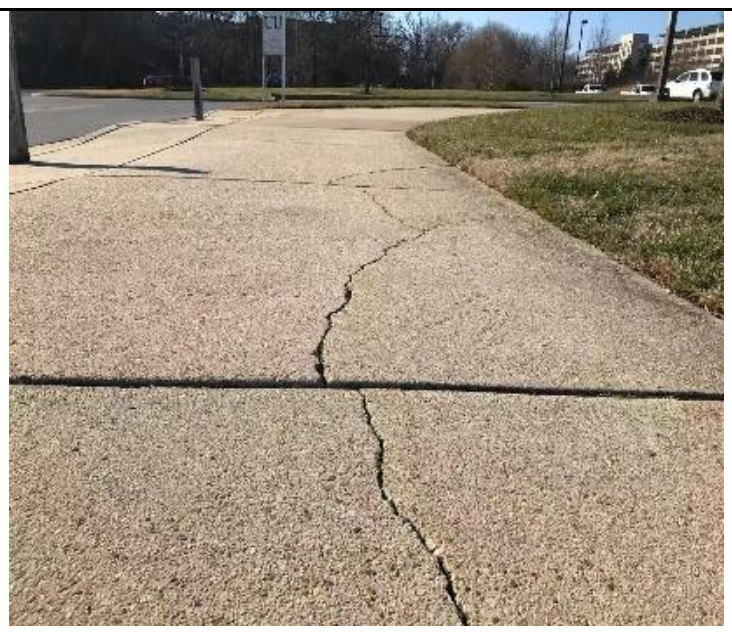 |
| Crack area | 1887 | 1425 |
| Spine length | 596 | 468 |
| Crack average width | 3.0 | 2.9 |

Table 3.5 presents post-correction results, including match statistics, geometric accuracy, and inlier distribution.

Table 3.5: Keypoint Matching and Calibration Results

(a) Matching Before/After RANSAC

| | Matches Before RANSAC | Matches After RANSAC |
|---|---|---|
| SIFT | 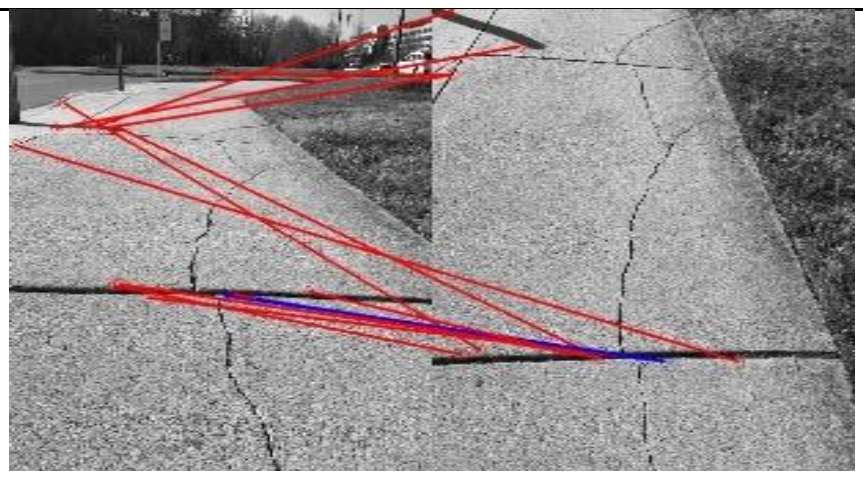 | 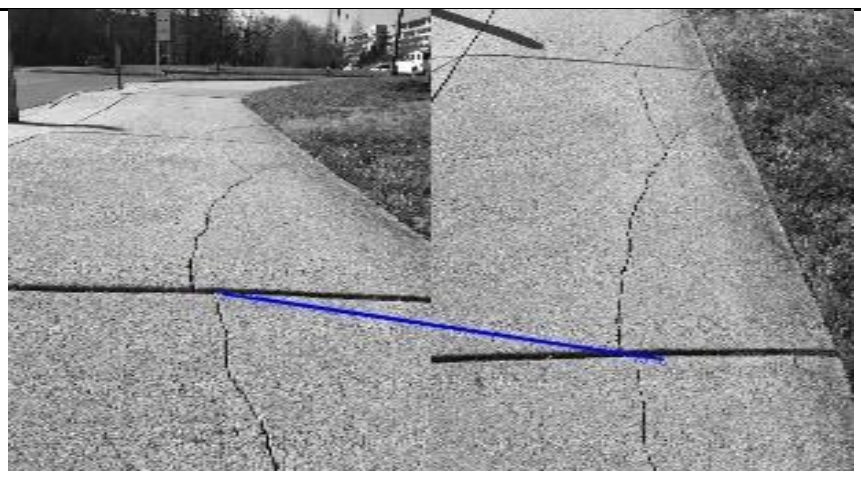 |
| SURF | 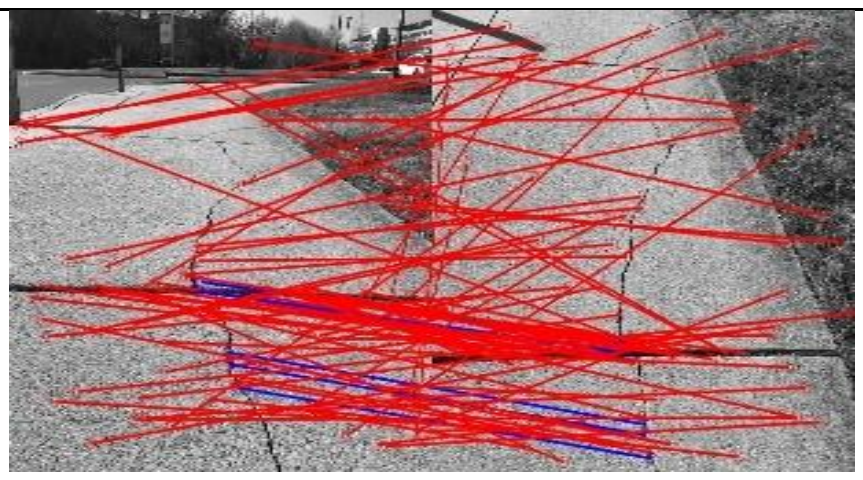 | 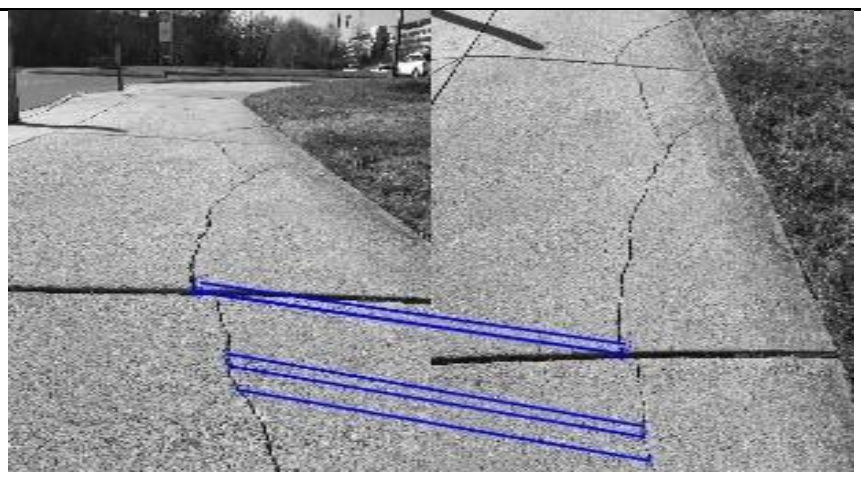 |
| ORB | N/A | N/A |
| BRISK | 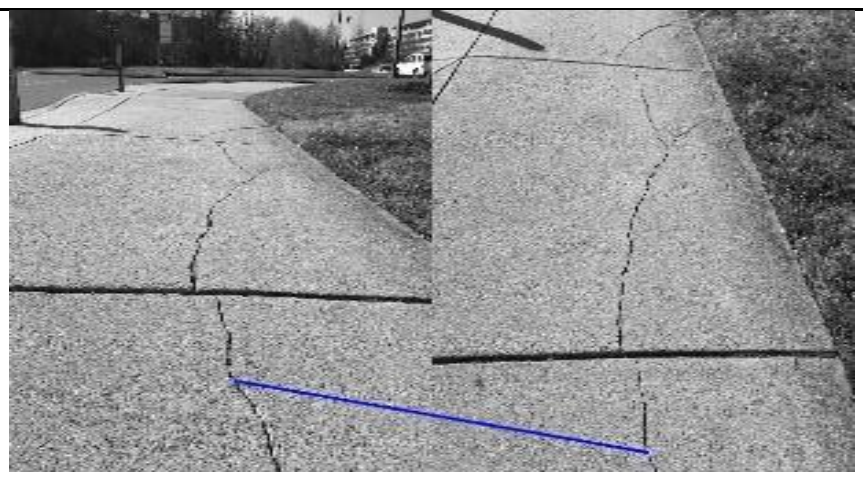 | 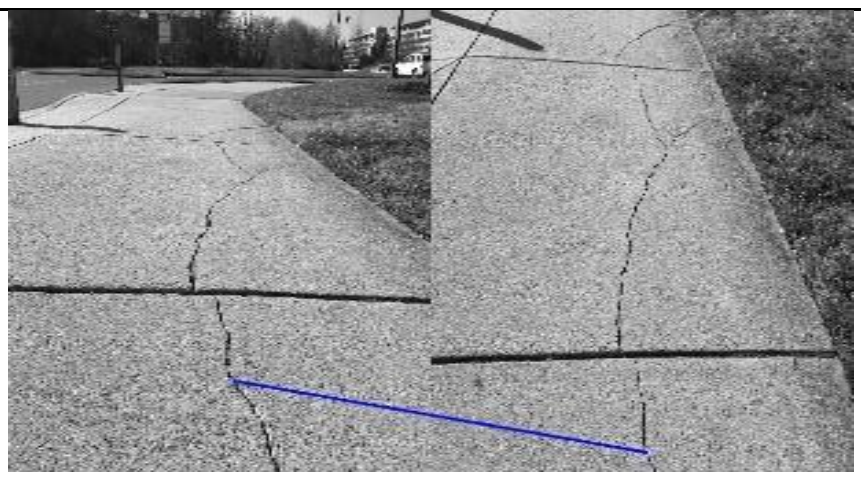 |
| KAZE | 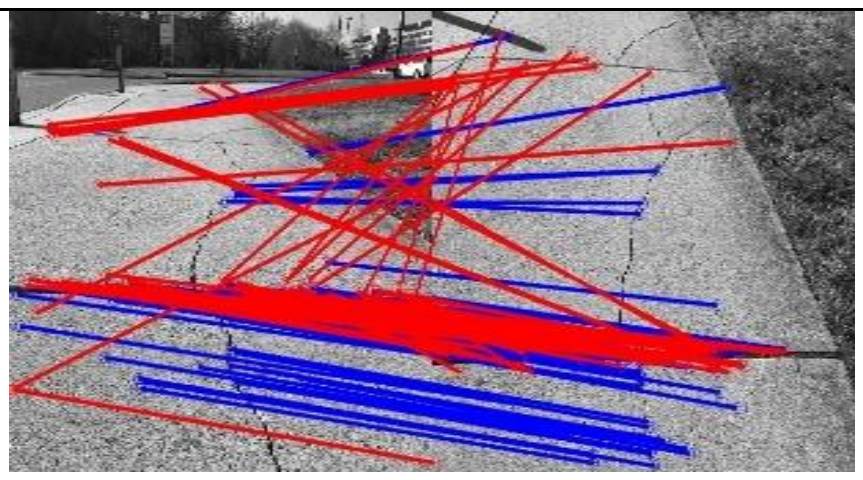 | 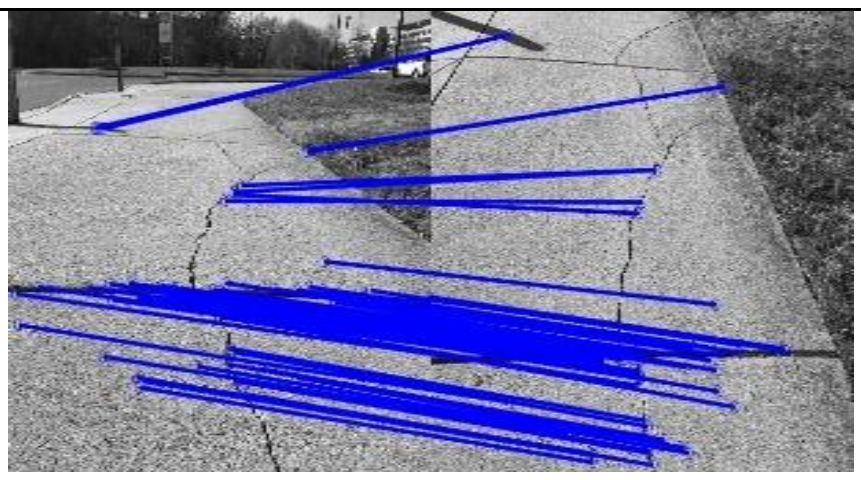 |

(b) Crack Overlap with Ground Anchors

| Method | Corrected Image 2 (Blue) | Baseline Image 1 (Red) | Overlap Accuracy |
|---|---|---|---|
| SIFT | N/A | N/A | Failed |
| SURF | 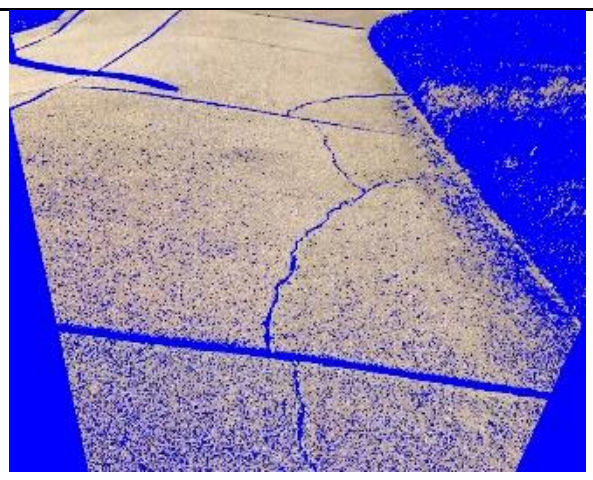 | 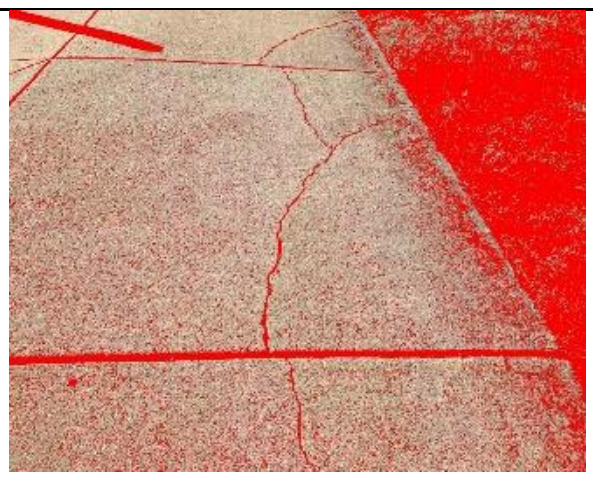 | 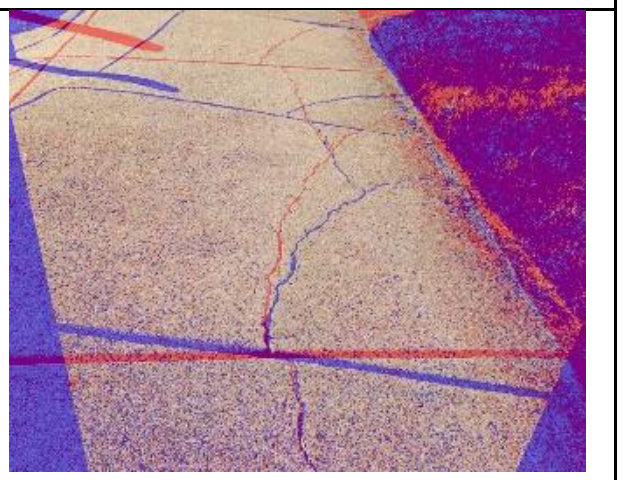 |
| ORB | N/A | N/A | Failed |
| BRISK | N/A | N/A | Failed |
| KAZE | 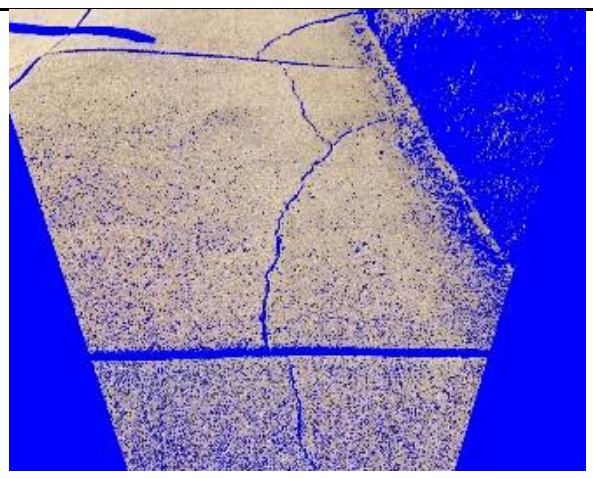 | 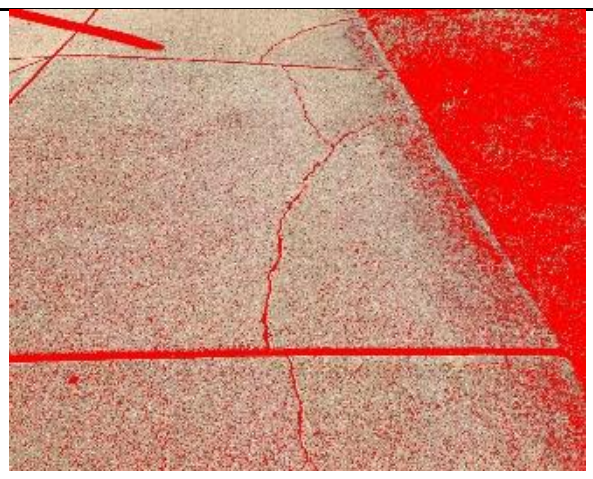 | 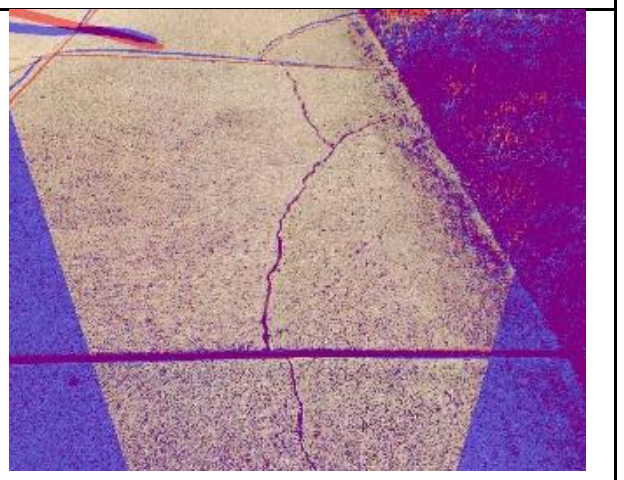 |

(c) Dimensional Accuracy vs. Baseline (pixels and % error)

| | Baseline | SIFT | SURF | ORB | BRISK | KAZE |
|---|---|---|---|---|---|---|
| Crack Area | 1887 | – | 1695 | – | – | 1815 |
| Area Error (%) | – | – | 10 | – | – | 4 |
| Spine Length | 596 | – | 439 | – | – | 509 |
| Length Error (%) | – | – | 26 | – | – | 15 |
| Average Crack Width | 3.0 | – | 3.4 | – | – | 3.2 |
| Width Error (%) | – | – | 13 | – | – | 7 |

(d) Spatial Distribution of Inlier Matches

| Detector | Total Inliers | Crack-Aligned | Background | Structural Coverage | Reliability Summary |
|---|---|---|---|---|---|
| SIFT | 1 | 0 | 0 | None | Too few matches; insufficient for crack anchoring |
| SURF | 6 | 5 | 1 | Weak | Mostly crack-aligned, but too sparse for reliable correction |
| ORB | 0 | 0 | 0 | None | No valid correspondences detected |
| BRISK | 1 | 1 | 0 | Very weak | Isolated match on crack; no support for global geometry |
| KAZE | 58 | 23 | 10 | Strong | Focused on cracks and corner anchors; structurally meaningful |

*Note: "Background" refers to valid RANSAC inliers not located on the crack path, such as edge features or slab texture points. Corner anchors were not individually counted but contributed to total inlier counts.*

Despite the compounded challenges of partial occlusion and severe geometric distortion, only KAZE-RANSAC achieved both geometric fidelity and structural coherence. It was the only method to consistently produce inliers along the crack and structural anchors, resulting in a well-constrained homography and near-perfect visual overlap with the baseline. KAZE also preserved spatial consistency across the cropped region, minimizing projection error even without full contextual visibility.

In contrast, SURF-RANSAC, though able to detect five crack-centered matches, failed to establish a stable homography. Its inliers were spatially clustered, lacking the distribution necessary for global correction, resulting in 26% spine length error and skewed geometry.

SIFT, ORB, and BRISK were unable to generate valid transformations. ORB detected no inliers; SIFT and BRISK each retained only one, neither of which provided sufficient spatial distribution. Although BRISK's match lay directly on the crack, it lacked geometric leverage to constrain the transformation.

These findings reinforce that in real-world SHM, spatial distribution and structural anchoring—not match count alone—are critical for alignment accuracy. KAZE's nonlinear scale space preserved edge continuity and resisted clutter, enabling robust perspective correction even under conditions that caused traditional detectors to fail.

### 3.3: Perspective Correction on Highly Textured Brick Surfaces

This experiment evaluates alignment accuracy in the presence of high-frequency textured backgrounds—a common challenge in masonry façade inspections where fine cracks coexist with dominant mortar patterns. The tested crack, embedded in a vertical brick wall, was surrounded by repetitive joints and sharp edges that generate dense but semantically irrelevant keypoints.

Two views were acquired: one front-facing as baseline and the other from a strong oblique angle to simulate constrained inspections. The distance to the wall surface was approximately 0.5 meters.

Perspective distortion caused substantial geometric inflation in the oblique image: crack area increased by 176%, spine length by 26%, and width by 38%, highlighting the disruptive effect of structured textures on alignment.

Table 3.6: Crack Dimensions Before Perspective Correction (All units in pixels)

| Image description | Image 1 (baseline) | Image 2 |
|---|---|---|
| Image | 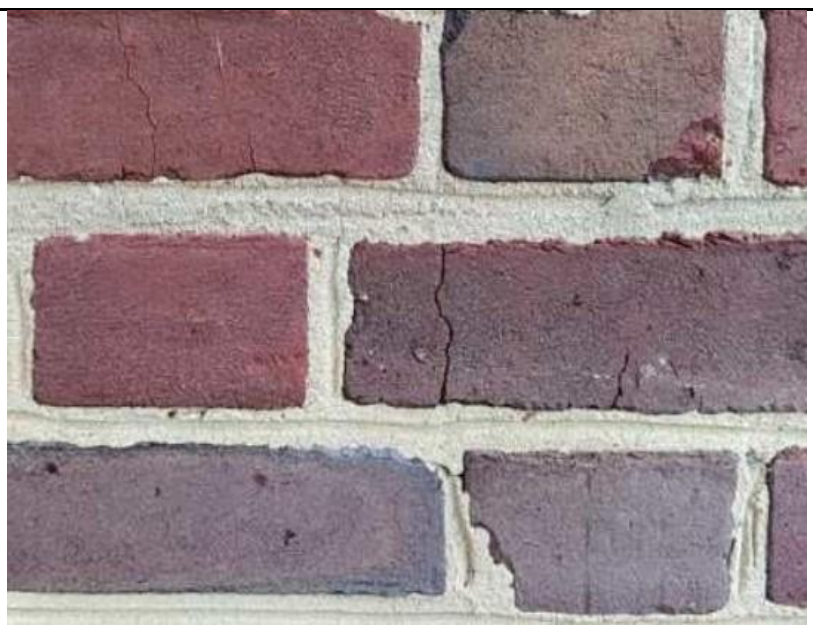 | 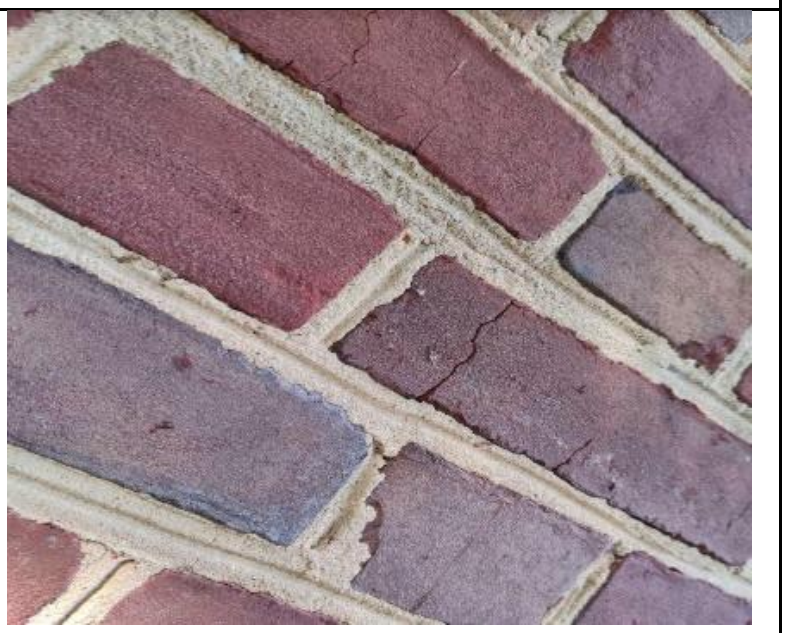 |
| Crack area | 278 | 767 |
| Spine length | 103 | 130 |
| Crack average width | 2.6 | 3.6 |

Table 3.7 presents post-correction results, including matching performance, alignment accuracy, and inlier distribution.

Table 3.7: Keypoint Matching and Calibration Results

(a) Matching Before/After RANSAC

| | Matches Before RANSAC | Matches After RANSAC |
|---|---|---|
| SIFT | 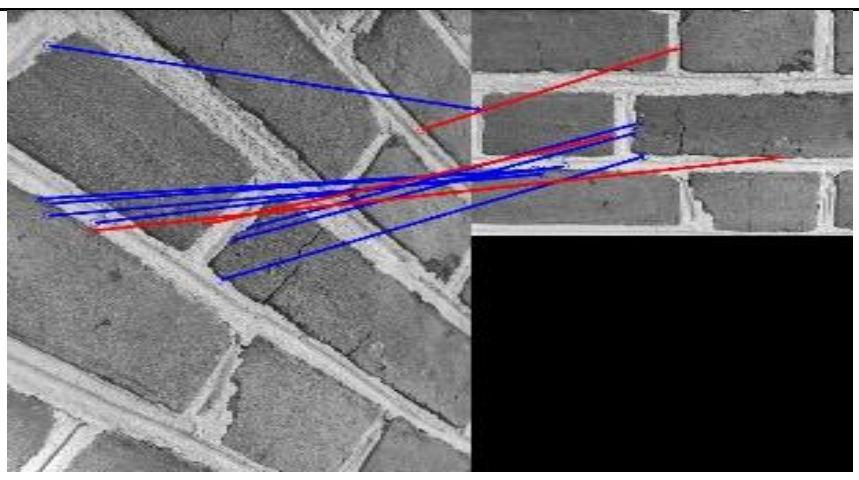 | 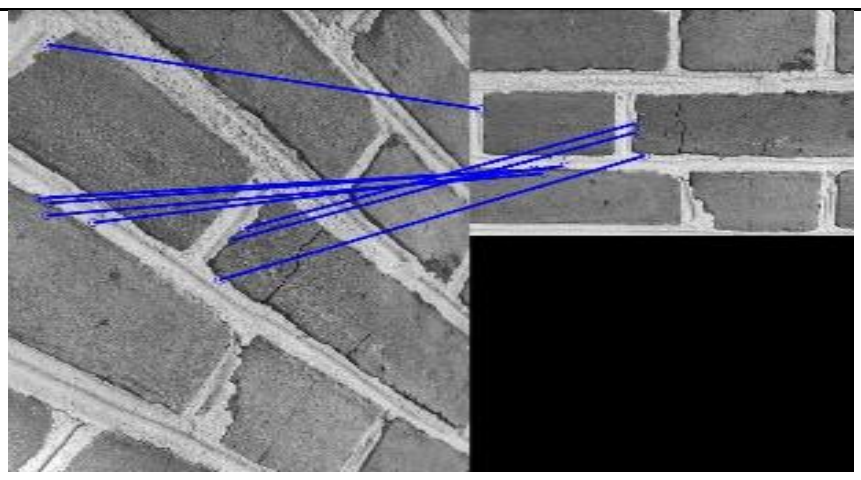 |
| SURF | 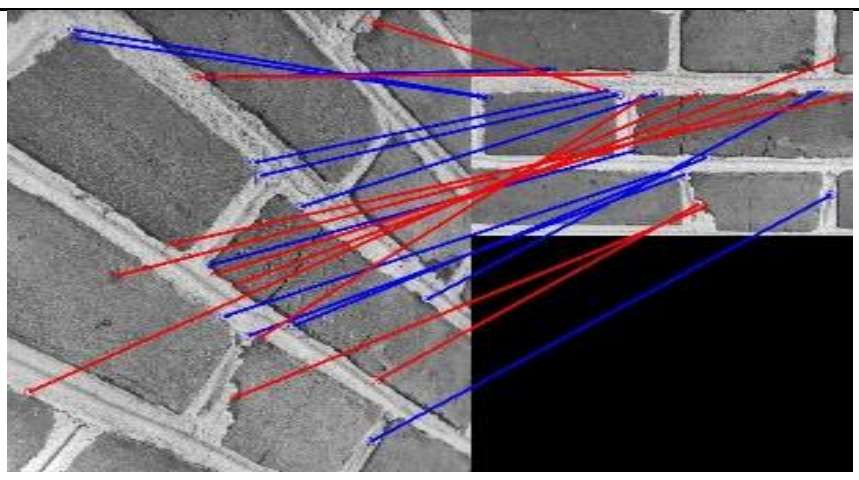 | 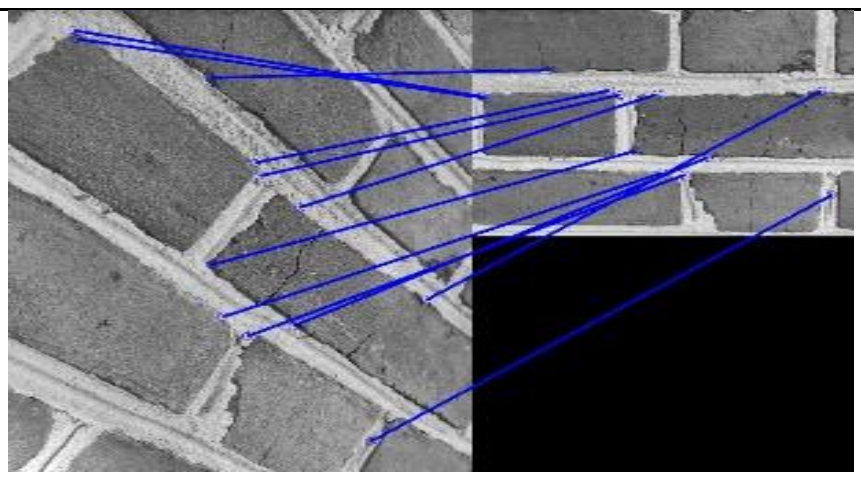 |
| ORB | 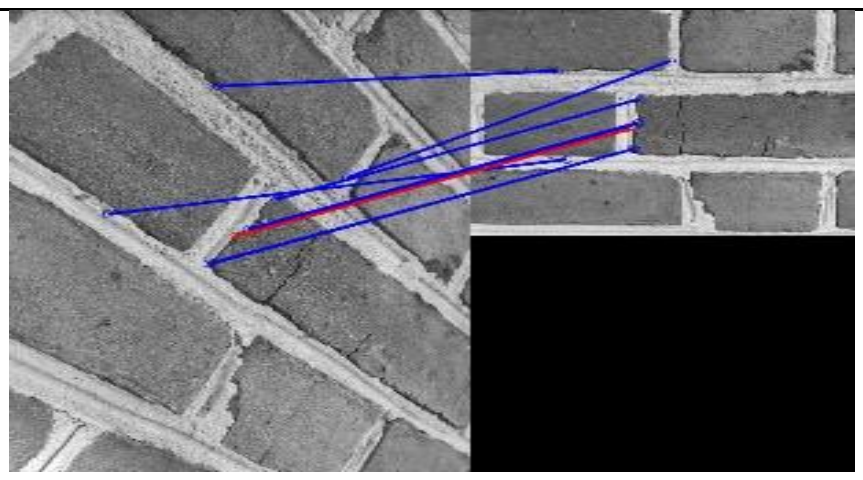 | 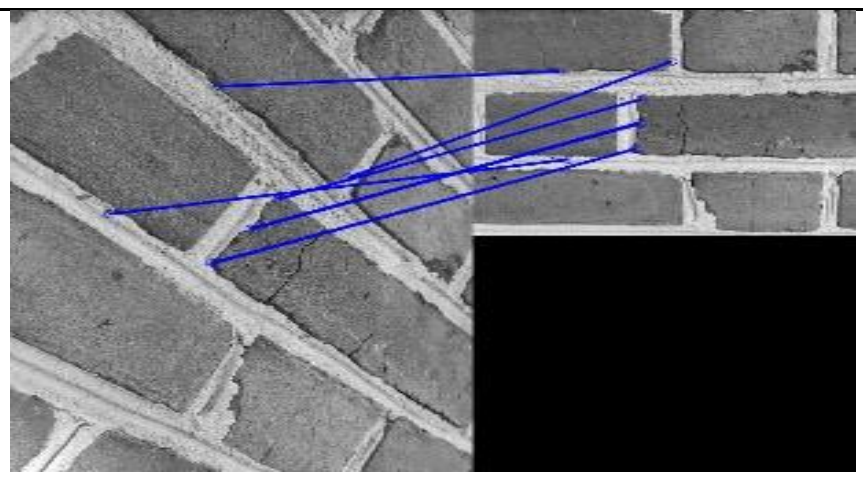 |
| BRISK | 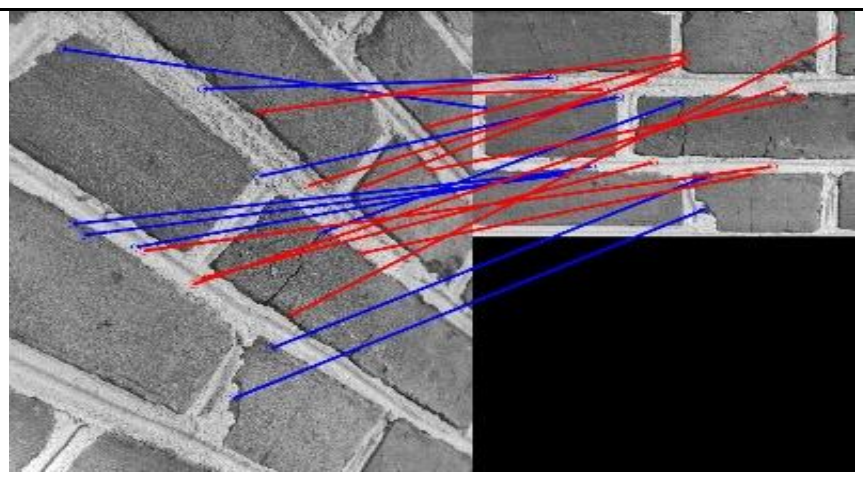 | 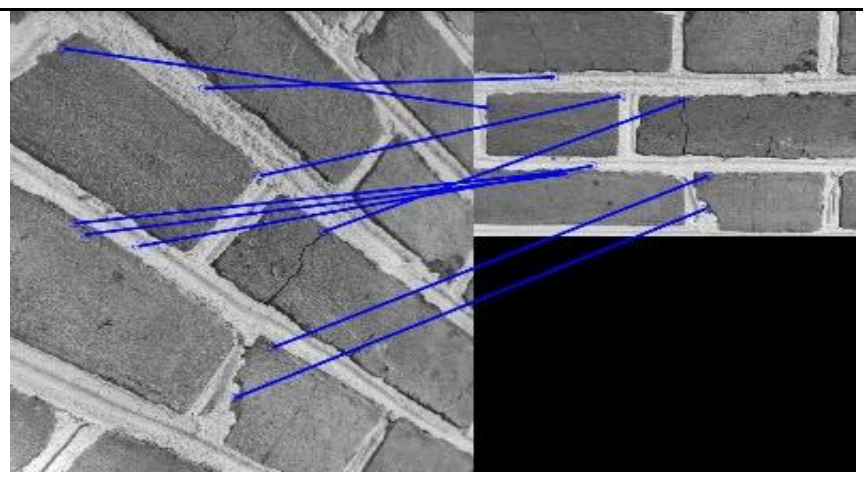 |
| KAZE | 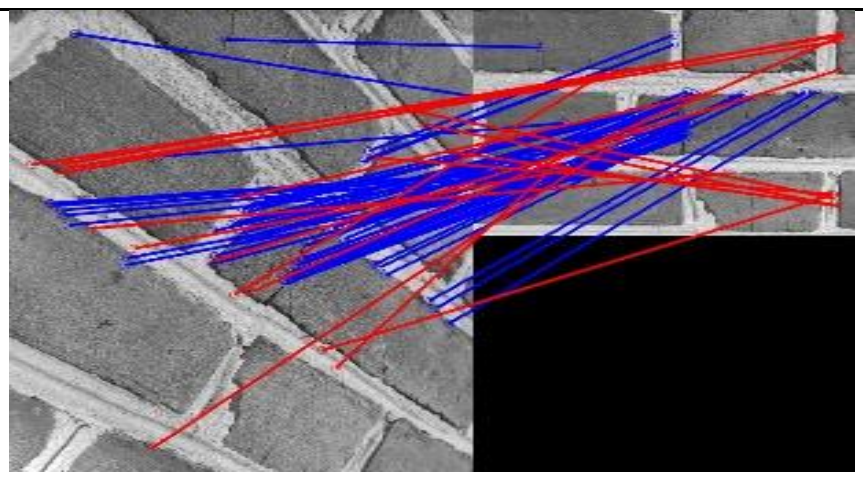 | 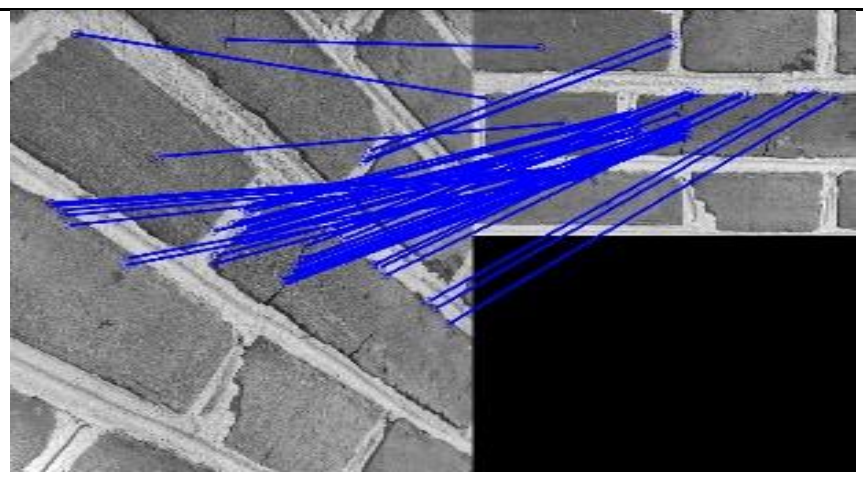 |

(b) Crack Overlap with Ground Anchors

| Method | Corrected Image 2 (Blue) | Baseline Image 1 (Red) | Overlap Accuracy |
|---|---|---|---|
| SIFT | 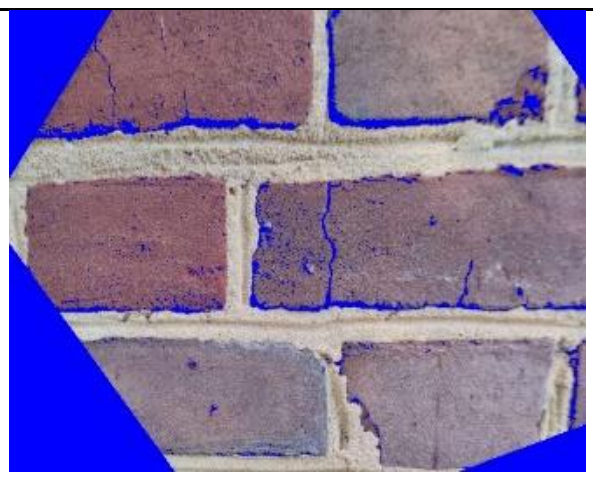 | 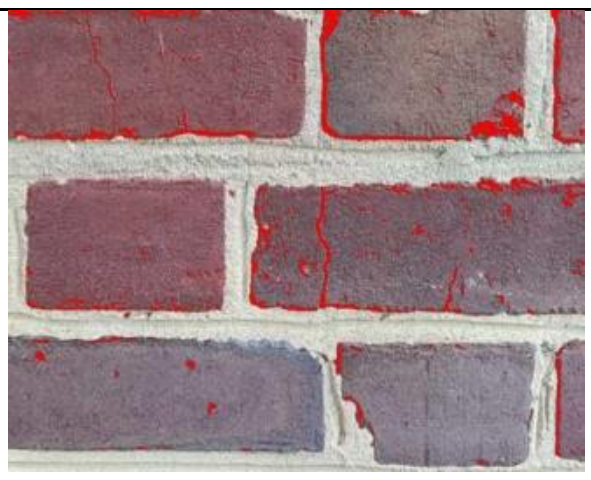 | 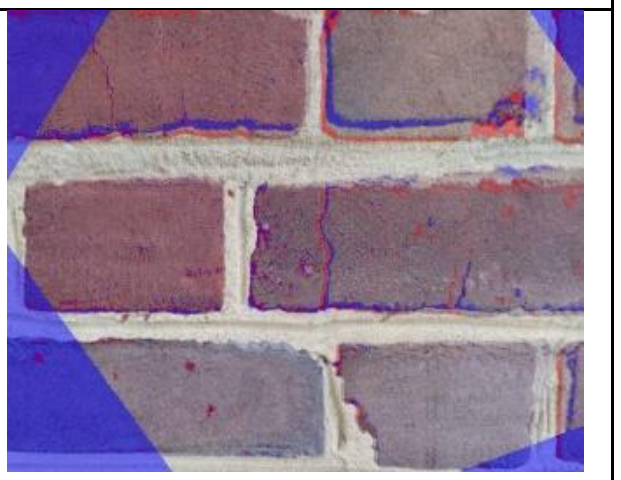 |
| SURF | 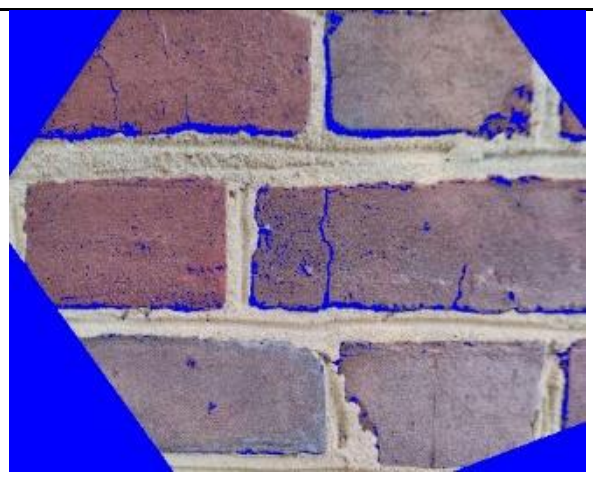 | 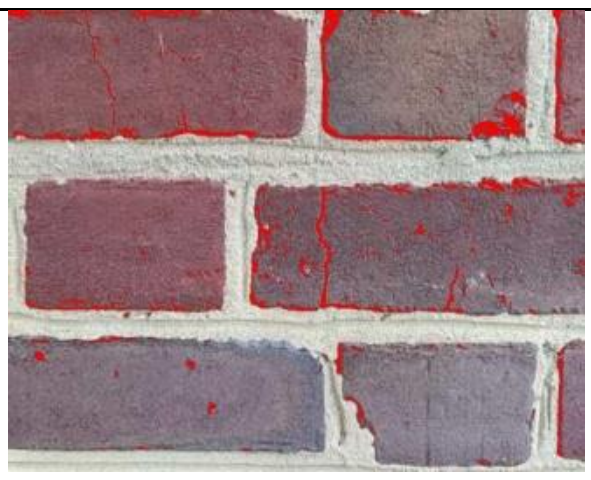 | 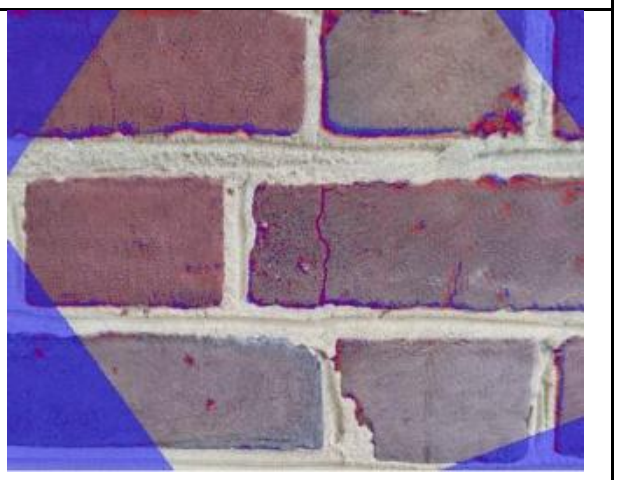 |
| ORB | 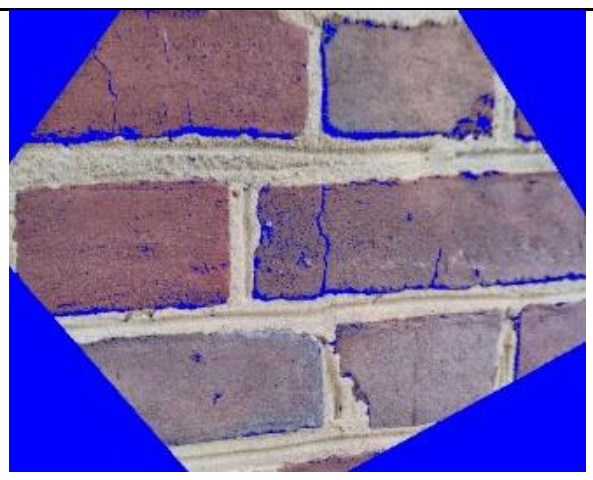 | 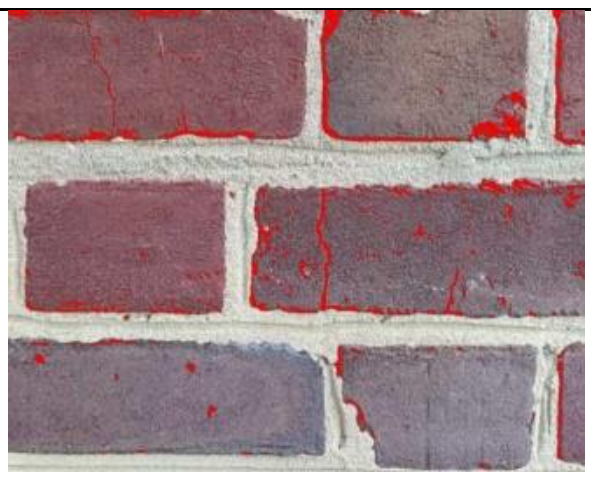 | 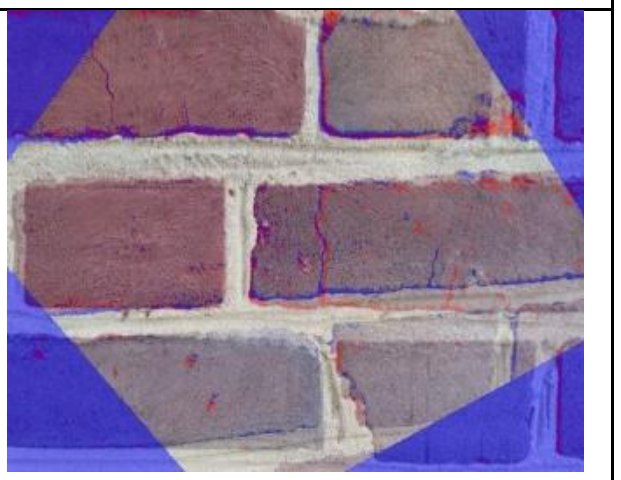 |
| BRISK | 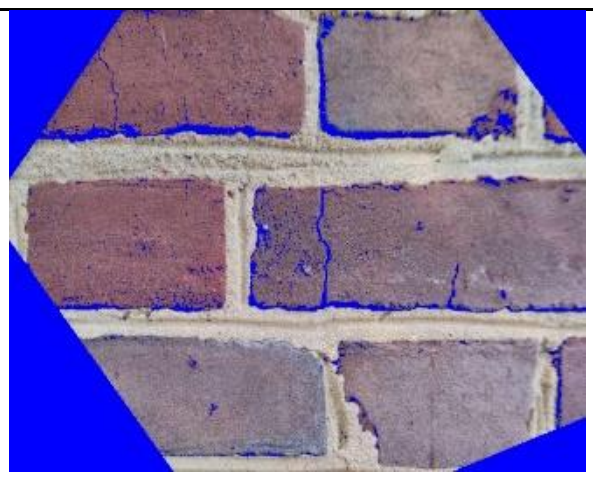 | 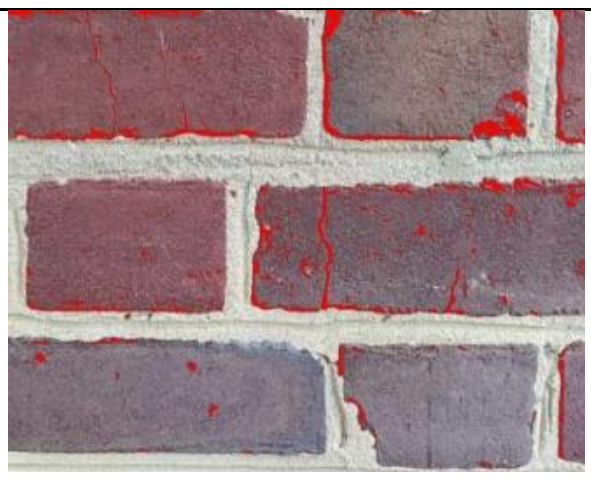 | 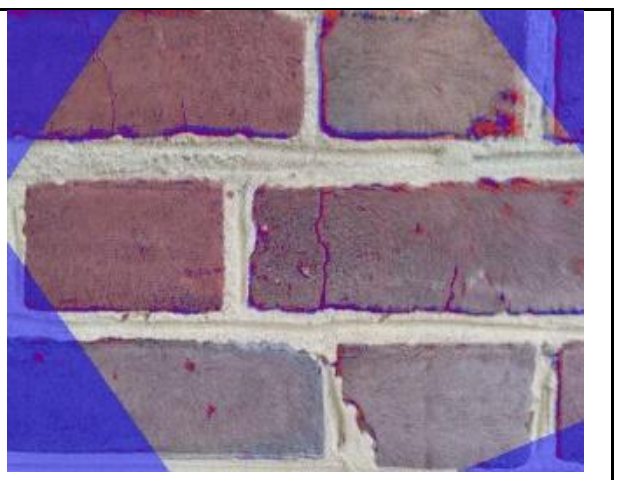 |
| KAZE | 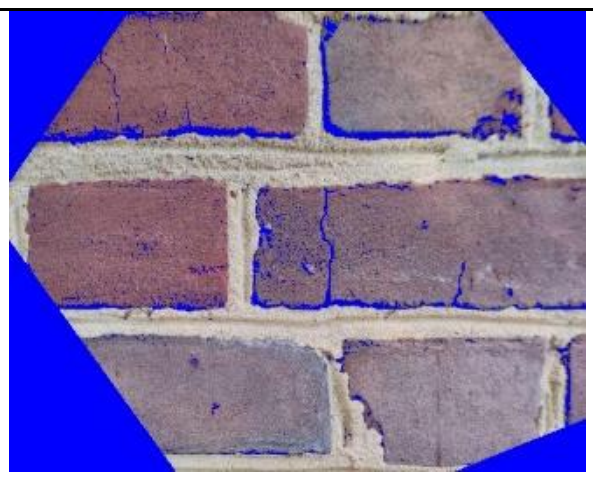 | 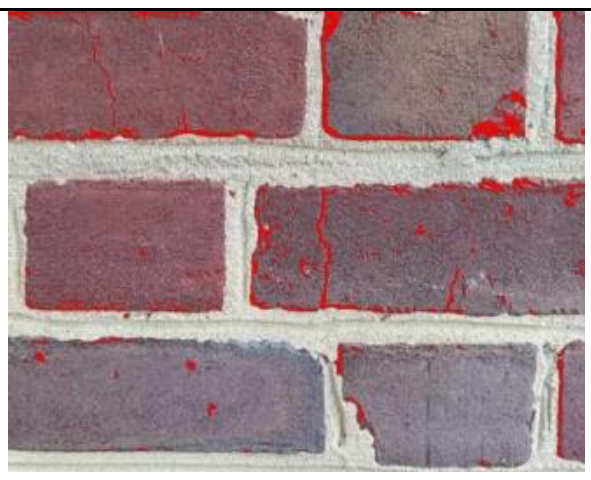 | 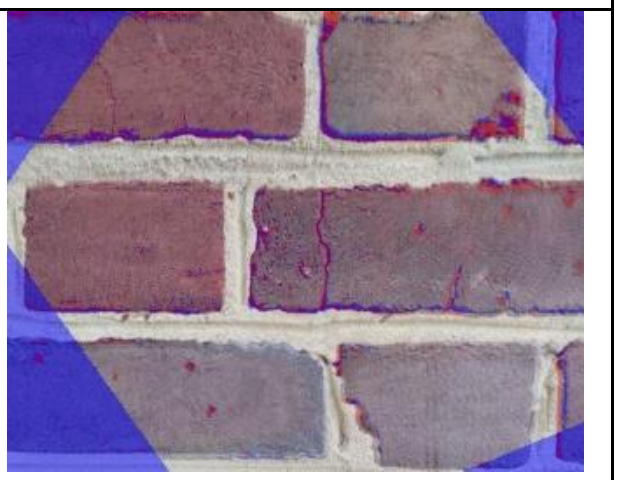 |

(c) Dimensional Accuracy vs. Baseline (pixels and % error)

| | Baseline | SIFT | SURF | ORB | BRISK | KAZE |
|---|---|---|---|---|---|---|
| Crack Area | 278 | 291 | 288 | 264 | 303 | 286 |
| Area Error (%) | – | 5 | 4 | 5 | 9 | 3 |
| Spine Length | 103 | 110 | 110 | 101 | 107 | 109 |
| Length Error (%) | – | 7 | 7 | 2 | 4 | 6 |
| Average Crack Width | 2.6 | 2.5 | 2.5 | 2.5 | 2.6 | 2.5 |
| Width Error (%) | – | 0.4 | 0.4 | 0.4 | 0 | 0.4 |

(d) Spatial Distribution of Inlier Matches

| Detector | Total Inliers | Crack-Aligned | Background | Structural Coverage | Reliability Summary |
|---|---|---|---|---|---|
| SIFT | 8 | 0 | 1 | Weak | Concentrated near bricks; no crack-specific focus |
| SURF | 12 | 0 | 4 | Weak | Background dominant; low semantic relevance |
| ORB | 6 | 0 | 1 | Very weak | Minimal spatial structure; fragile alignment |
| BRISK | 9 | 1 | 2 | Moderate | Limited crack sensitivity; partial correction |
| KAZE | 53 | 10 | 3 | Strong | Edge-preserving; crack and anchor focused |

*Note: "Background" refers to valid RANSAC inliers not located on the crack path, such as edge features or slab texture points. Corner anchors were not individually counted but contributed to total inlier counts.*

All five pipelines produced valid homographies, but crack alignment accuracy varied substantially.

KAZE-RANSAC achieved the most reliable results. Among 53 inliers, 10 were directly aligned with the crack, with others located at corner features. The resulting transformation preserved full crack continuity with only 3% area and 0.4% width error, highlighting resilience to structured background interference.

SURF-RANSAC and BRISK-RANSAC also achieved low numerical errors (≤9%), but their matches concentrated on repetitive textures rather than crack edges, resulting in brittle or misleading transformations.

SIFT-RANSAC, though producing visually acceptable results, misaligned the crack due to mortar-focused keypoints. ORB-RANSAC performed weakest, with insufficient match structure to recover full alignment.

These results indicate that while textured backgrounds may increase keypoint density, only KAZE consistently anchors semantically meaningful features. Its nonlinear diffusion process preserves subtle crack cues—making it particularly effective for masonry diagnostics under visual interference.

### 3.4: Perspective Correction for Cracks Under Moving Shadow Conditions

This experiment evaluates alignment robustness under dynamic shadow interference—a common challenge in field-deployed SHM inspections. Moving shadows from personnel or equipment introduce luminance gradients that obscure crack edges and generate spurious keypoints.

Two images were acquired at oblique and frontal angles, with one cast under uniform daylight and the other containing a manually introduced moving shadow. This scenario simulates real-world inspection variability caused by nonuniform lighting.

Before correction, shadow distortion caused crack area to increase by 36% and spine length by 14%, while average width remained unchanged—highlighting the impact of lighting artifacts on geometric accuracy.

Table 3.8: Crack Dimensions Before Perspective Correction (All units in pixels)

| Image description | Image 1 (baseline) | Image 2 |
|---|---|---|
| Image | 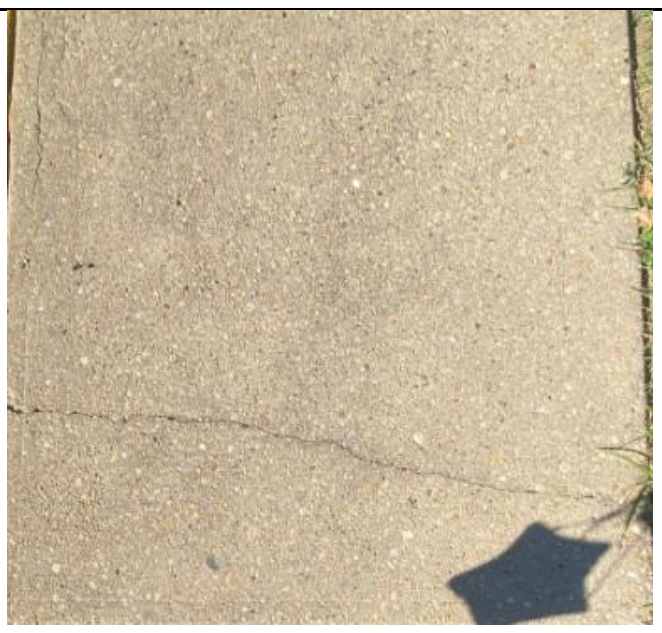 | 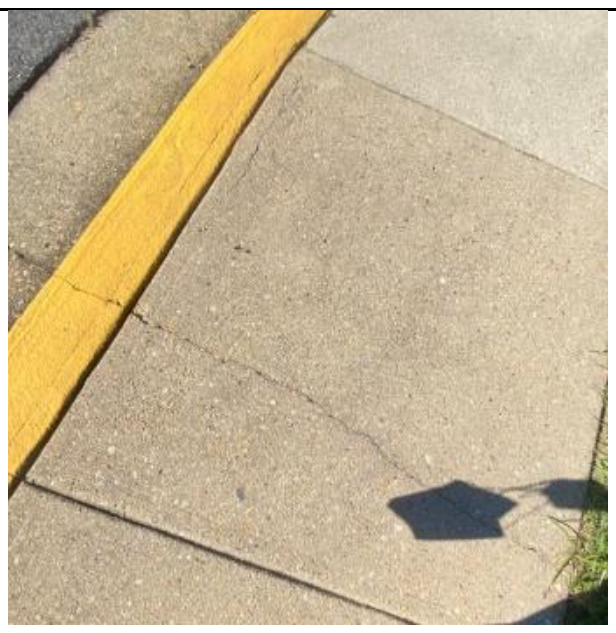 |
| Crack area (pixels) | 571 | 774 |
| Spine length (pixels) | 392 | 445 |
| Crack average width | 2.1 | 2.1 |

Post-correction results are summarized in Table 3.9, including match statistics, alignment quality, and inlier analysis.

Table 3.9: Keypoint Matching and Calibration Results

(a) Matching Before/After RANSAC

| | Matches Before RANSAC | Matches After RANSAC |
|---|---|---|
| SIFT | 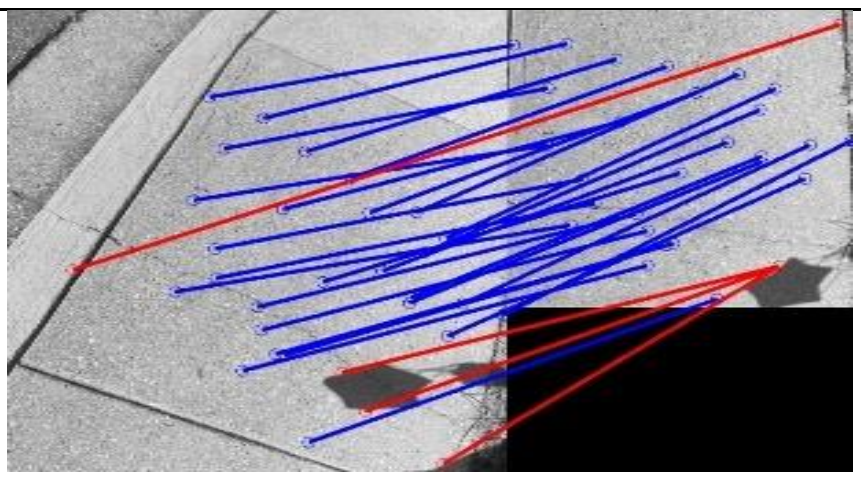 | 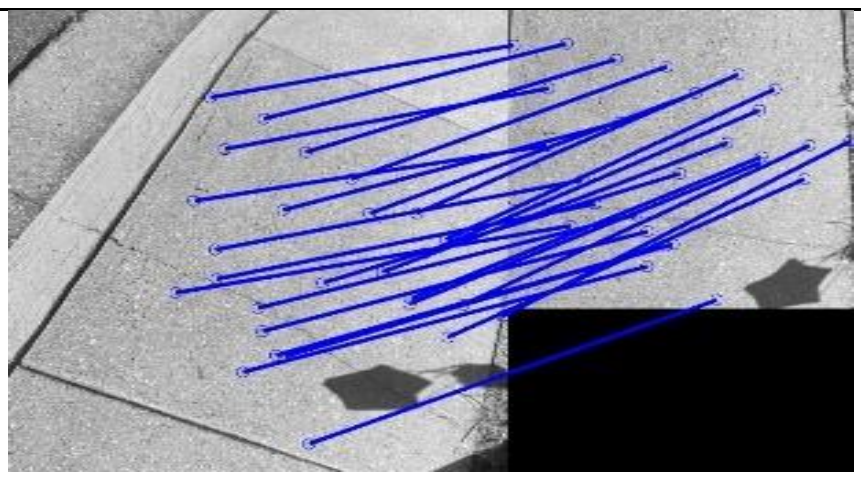 |
| SURF | 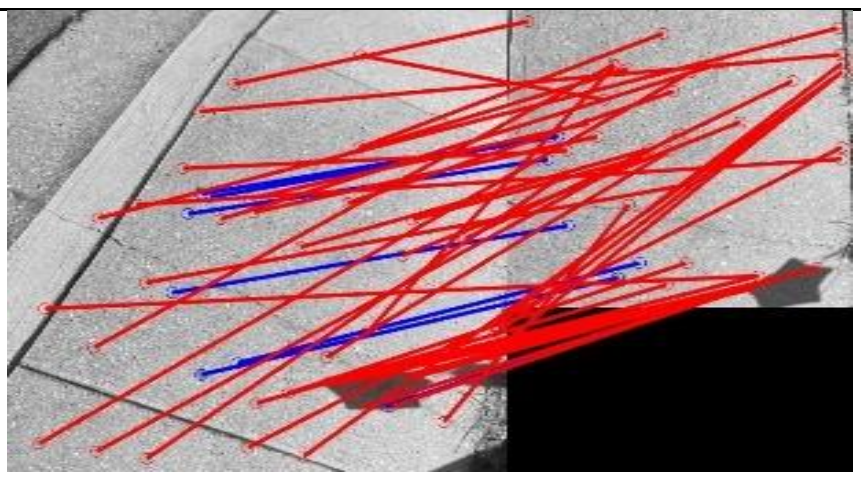 | 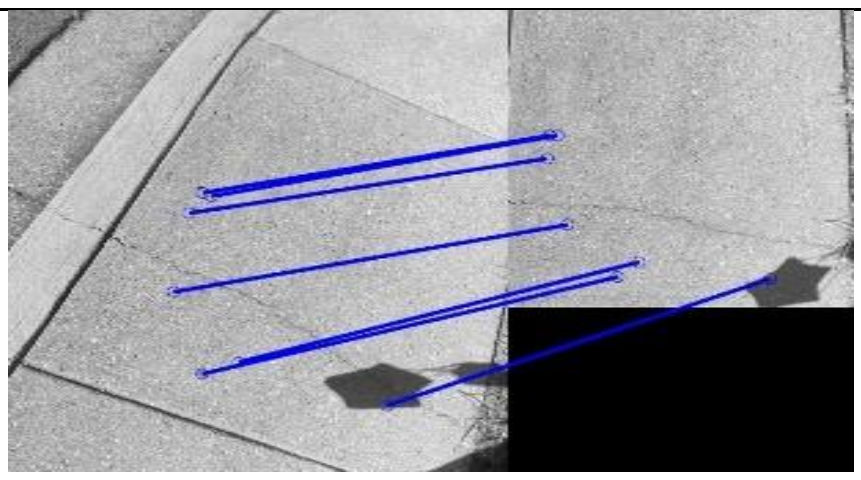 |
| ORB | 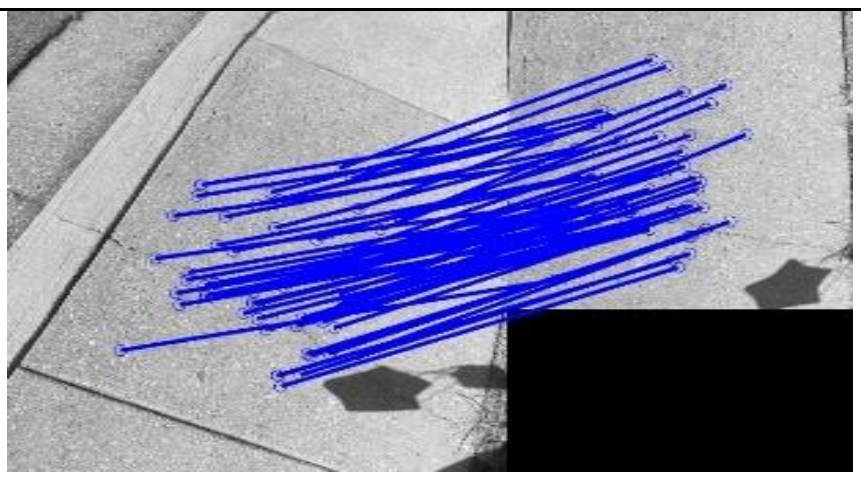 | 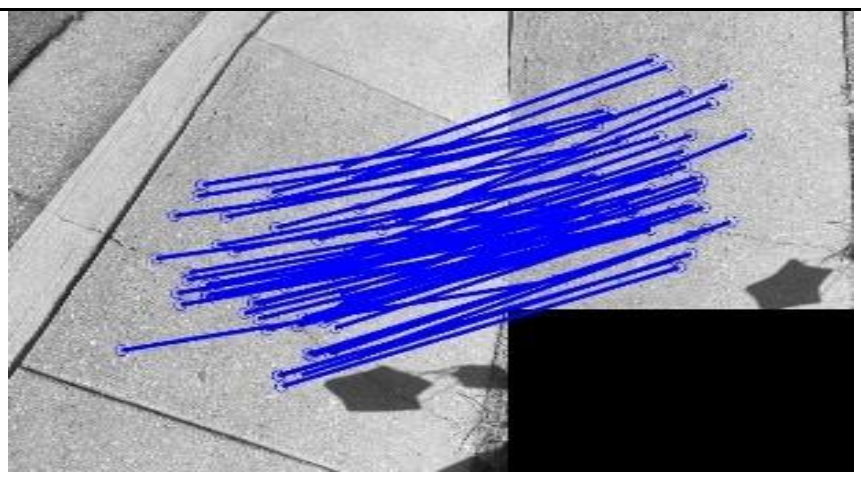 |
| BRISK | 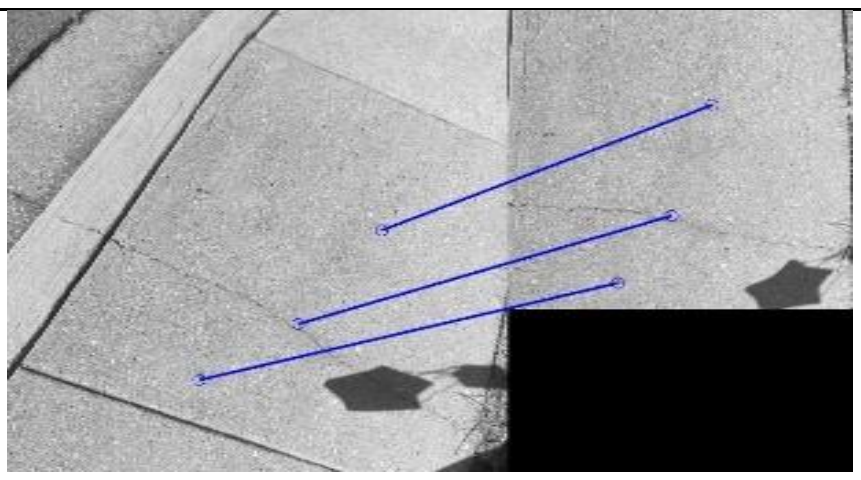 | 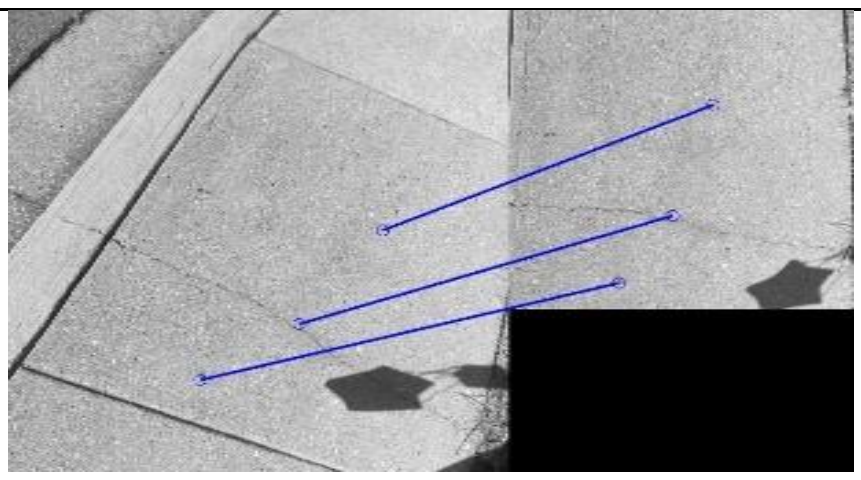 |
| KAZE | 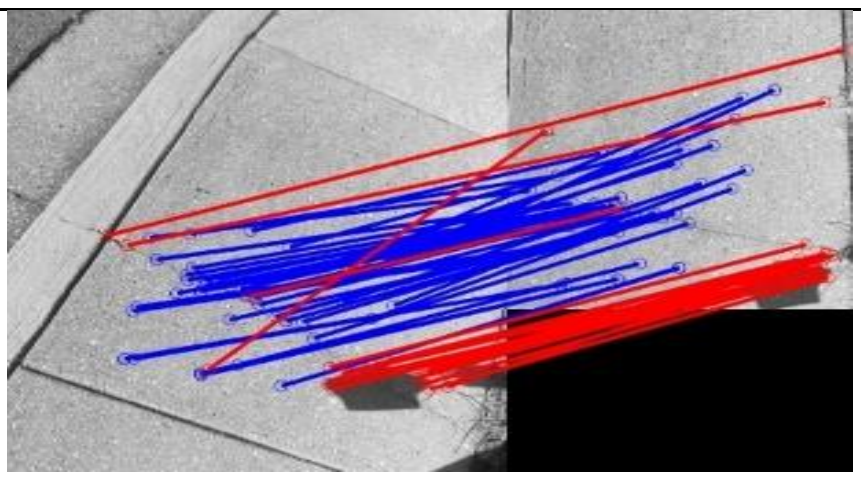 | 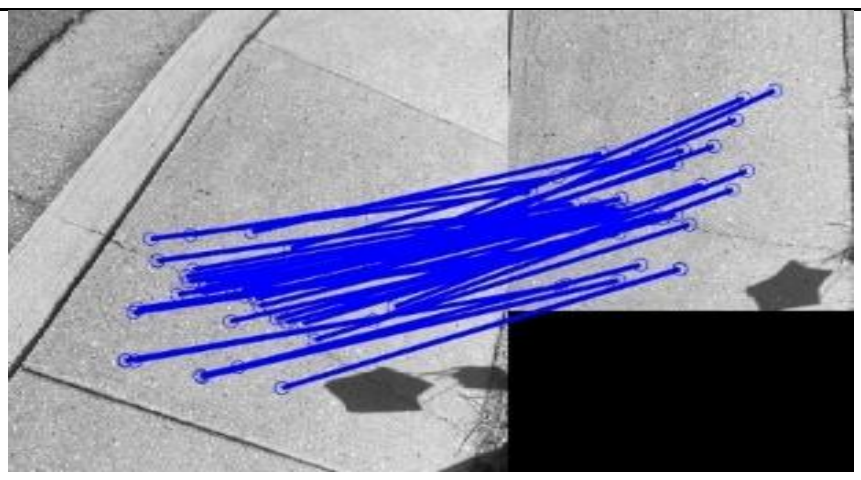 |

(b) Crack Overlap with Ground Anchors

| Method | Corrected Image 2 (Blue) | Baseline Image 1 (Red) | Overlap Accuracy |
|---|---|---|---|
| SIFT | 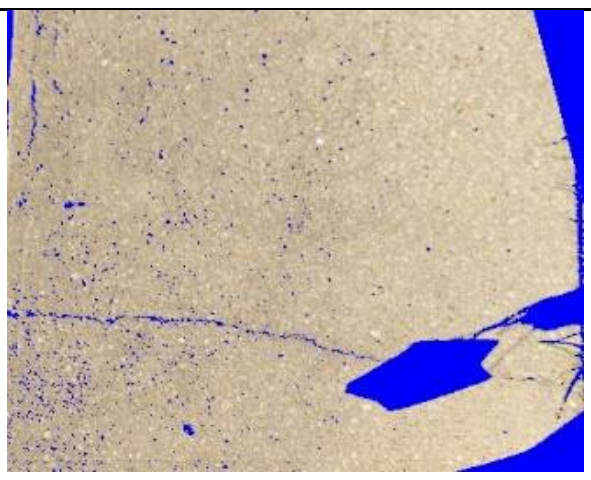 | 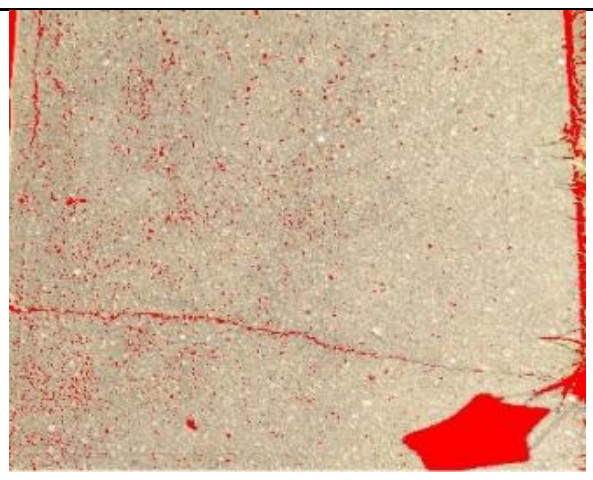 | 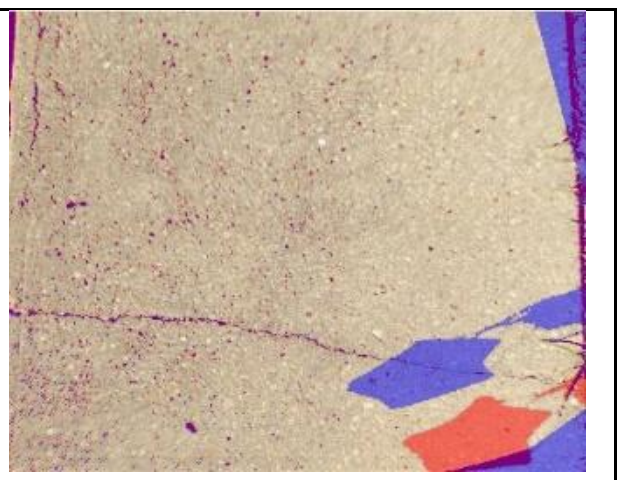 |
| SURF | 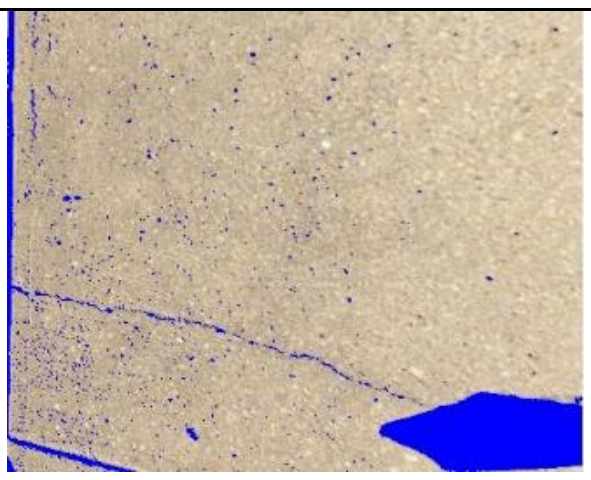 | 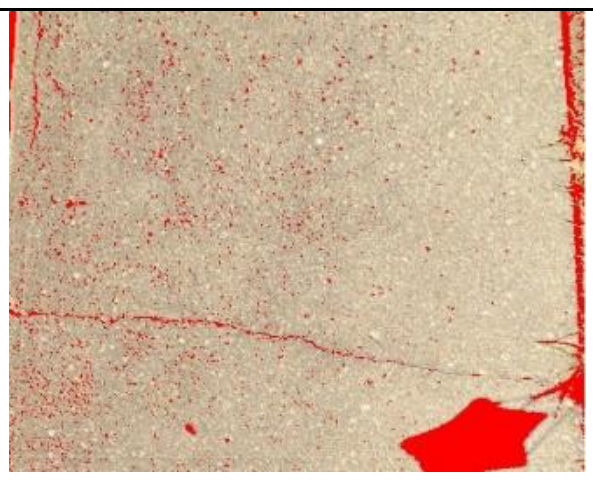 | 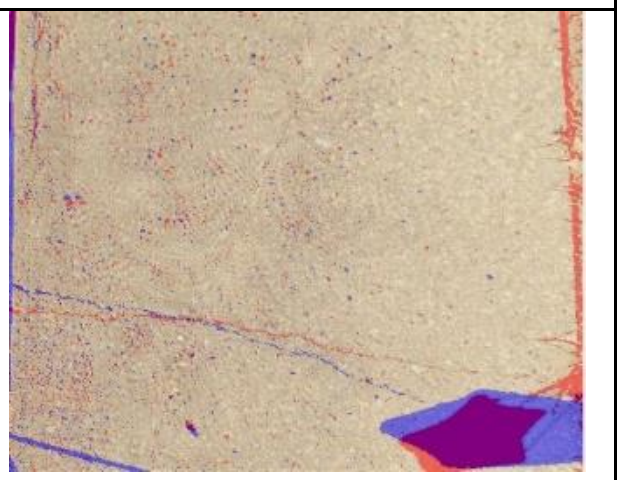 |
| ORB | 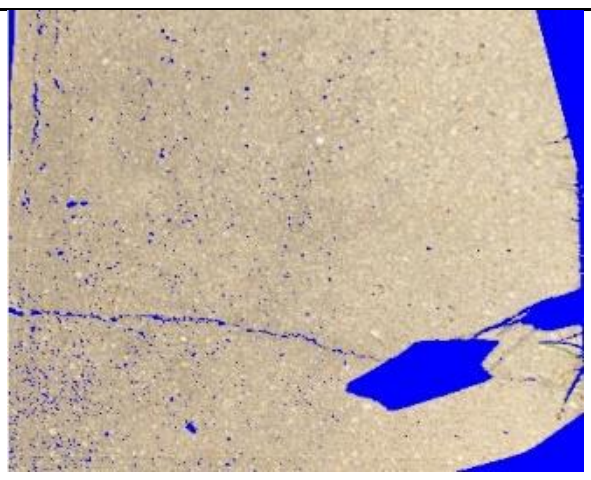 | 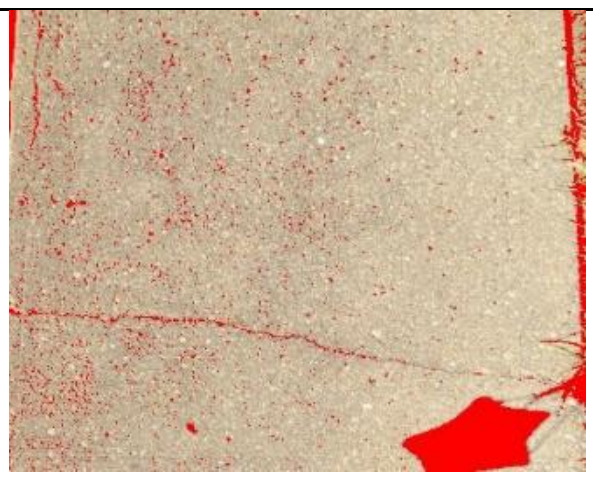 | 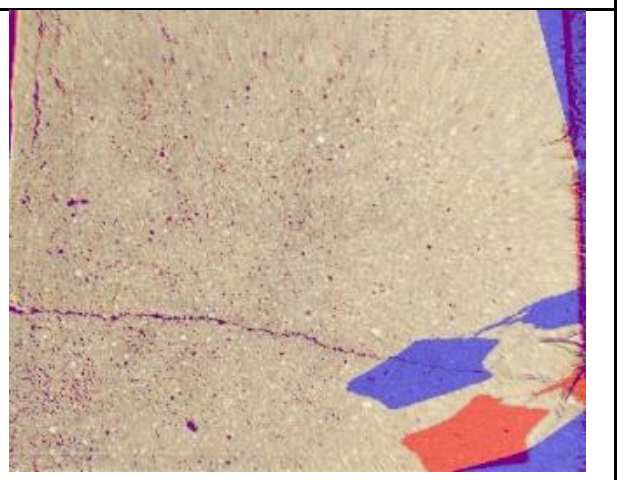 |
| BRISK | N/A | N/A | Failed |
| KAZE | 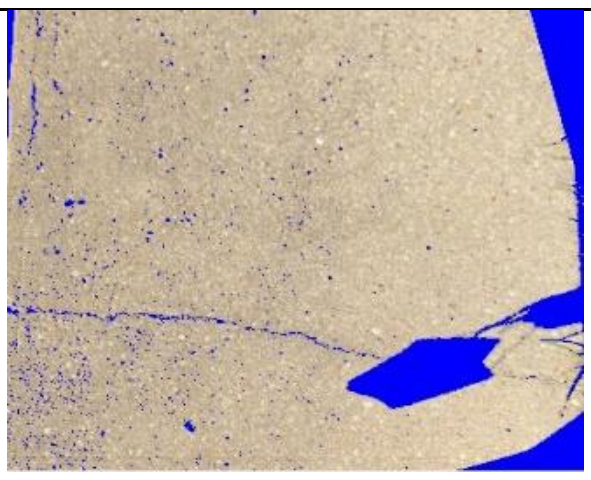 | 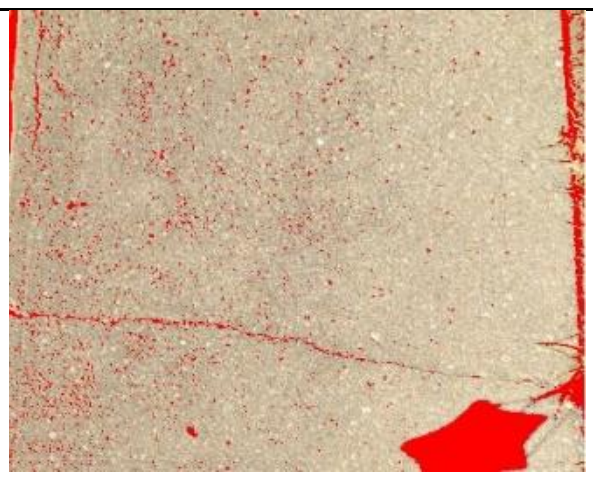 | 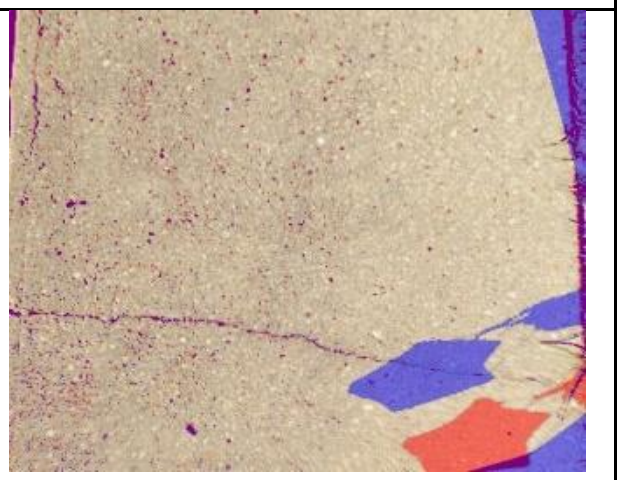 |

(c) Dimensional Accuracy vs. Baseline (pixels and % error)

| | Baseline | SIFT | SURF | ORB | BRISK | KAZE |
|---|---|---|---|---|---|---|
| Crack Area | 571 | 550 | 726 | 603 | – | 580 |
| Area Error (%) | – | 4 | 27 | 6 | – | 2 |
| Spine Length | 392 | 369 | 358 | 396 | – | 385 |
| Length Error (%) | – | 6 | 9 | 1 | – | 2 |
| Average Crack Width | 2.1 | 2.0 | 2.3 | 2.1 | – | 2.1 |
| Width Error (%) | – | 5 | 10 | 0 | – | 0 |

(d) Spatial Distribution of Inlier Matches

| Detector | Total Inliers | Crack-Aligned | Background | Structural Coverage | Reliability Summary |
|---|---|---|---|---|---|
| SIFT | 31 | 2 | 27 | Moderate | Despite high noise, geometrically valid alignment achieved |
| SURF | 7 | 0 | 6 | Weak | Shadow-dominated matches; no crack inliers |
| ORB | 59 | 11 | 48 | Strong | Crack-centered inliers dominate; stable under noise |
| BRISK | 3 | 1 | 2 | Very Weak | Single crack inlier; insufficient for alignment |
| KAZE | 254 | 38 | 213 | Strong | High crack match density; robust calibration under shadow |

*Note: "Background" refers to valid RANSAC inliers not located on the crack path, such as edge features or slab texture points. Corner anchors were not individually counted but contributed to total inlier counts.*

Under dynamic shadow interference, KAZE-RANSAC demonstrated the highest robustness and alignment fidelity among all tested methods. It retained 254 inliers, of which 38 were crack-centered and several others anchored along stable corner features. While the remaining background inliers were distributed along slab textures and surface edges, they did not interfere with homography estimation due to their geometric consistency. Instead, these points complemented the crack-aligned inliers by enhancing spatial coverage, resulting in a well-constrained transformation. The alignment achieved only 2% area error and 0% width error, preserving both geometric accuracy and visual coherence under lighting fluctuations.

ORB-RANSAC also yielded strong results, with 11 crack-aligned inliers out of 59. The spatial distribution of these matches followed the crack structure closely, enabling a stable transformation with low error across all metrics.

SIFT-RANSAC, although affected by shadow-induced gradient noise, produced a viable homography. Of its 31 inliers, only 2 were crack-centered, yet their locations were spatially aligned with the baseline geometry. The resulting transformation achieved 4% area error and preserved average crack width, though minor misalignment was observed due to high background noise ratio.

In contrast, SURF-RANSAC failed to localize any crack-specific inliers. All 7 retained matches were located on shadow edges or irrelevant background structures, producing a poorly constrained transformation with 27% area error and 9% spine length error.

BRISK-RANSAC did not return a valid homography, as only 3 inliers were detected—insufficient for reliable estimation.

These results emphasize that under challenging shadow conditions, the critical determinant of alignment quality is not the absolute number of matches, but the number and spatial distribution of crack-centered and structurally meaningful inliers. While KAZE produced more background matches due to its edge-preserving nature, it simultaneously maintained the highest count of crack-aligned and anchor-supported keypoints, which proved sufficient for accurate calibration. Its nonlinear diffusion filtering suppressed luminance artifacts while preserving subtle crack discontinuities—making it the only method that consistently maintained alignment accuracy under dynamic, real-world lighting variations.

### 3.5 Discussion: Overcoming Gaussian-Based Limitations in SHM Image Alignment

This study reveals that Gaussian-based feature detection—long dominant in image alignment—faces critical limitations in real-world SHM scenarios. Classical detectors such as SIFT, SURF, ORB, and BRISK, though efficient in general vision tasks, often falter under the complex visual and geometric conditions typical of SHM inspections.

1. Gaussian-Based Limitations in SHM

All four traditional methods rely on Gaussian smoothing or intensity-based corner detection, which present two key issues:

- Keypoint suppression: Narrow or low-contrast cracks are blurred, weakening or eliminating essential features.
- False match amplification: Repetitive backgrounds (e.g., bricks, shadows) are easily overemphasized, leading to misleading correspondences and spatial drift.

These effects were particularly evident in the cropped (3.2), masonry (3.3), and shadowed (3.4) scenarios, where most classical detectors failed to recover accurate alignment—even when yielding high inlier counts.

2. KAZE's Nonlinear Advantage

In contrast, KAZE constructs a nonlinear scale space via anisotropic diffusion, preserving high-frequency crack edges while suppressing irrelevant gradients. This enables robust detection of crack-centered features under occlusion, noise, and low contrast.

Across all experiments, KAZE consistently produced spatially meaningful inliers along cracks and anchor regions—proving more valuable than raw match count. For example, in 3.4, although KAZE generated over 250 inliers, it was the 38 crack-aligned ones that ensured geometry-consistent registration.

All experiments employed a uniform RANSAC sampling of 10 matched keypoints, empirically determined as the minimum for stable homography estimation across scenes. Higher counts often introduced noisy background features; lower counts led to instability. This fixed strategy balanced geometric reliability and computational efficiency.

3. Experimental Insights

Each test case exposed specific limitations of classical detectors and demonstrated the robustness of the proposed framework:

- 3.1 (Ideal perspective): Even with minimal interference, classical methods misaligned due to sparse anchors.
- 3.2 (Cropped cracks): Most methods failed entirely; only KAZE-RANSAC achieved valid alignment.
- 3.3 (Masonry wall): Classical methods favored repeated textures; KAZE retained crack geometry.

- 3.4 (Moving shadows): KAZE preserved alignment despite lighting changes, while others degraded or failed.

These results emphasize that the semantic relevance and spatial distribution of features are more critical than quantity—especially in SHM contexts where cracks may be degraded, occluded, or distorted.

4. Practical Considerations and Future Directions

While KAZE-RANSAC showed high robustness, several limitations remain. The framework currently assumes local planarity and was tested on images from a single device (iPhone 11, 12MP), representing a typical UAV/mobile platform. Broader validation is needed. Priorities for future work include:

- Cross-sensor and cross-platform validation across camera types and resolutions
- Adaptive keypoint sampling based on scene characteristics
- Extension to non-planar surfaces via stereo, SfM, or LiDAR
- Integration with deep features (e.g., SuperPoint, D2-Net) for hybrid pipelines

While CNN-based models show potential in semantic segmentation, they often require extensive training and high computational cost, and may lack geometric transparency in uncontrolled settings. By contrast, KAZE-RANSAC is interpretable, unsupervised, and computationally lightweight—making it attractive for field deployment. This work provides a practical alternative for scenarios where training data is limited and deployment transparency is critical.

5. Contribution in Context

This work presents what we believe to be one of the first detailed assessments of anisotropic scale-space filtering applied to SHM crack alignment under diverse field conditions. While KAZE has been explored in general vision contexts, its robustness for infrastructure imagery—particularly under distortion, occlusion, and shadow—has not been systematically benchmarked.

Unlike conventional vision datasets, SHM imagery often suffers from platform constraints (e.g., UAV capture), sparse textures, and a lack of labeled data. In this context, the KAZE-RANSAC framework offers a transparent, training-free alternative, tailored to SHM deployment needs—enabling reliable crack alignment without calibration or retraining.

Although the core algorithms (KAZE, RANSAC) are well known, this study contributes a domain-specific integration and evaluation framework, highlighting their practical value for infrastructure monitoring under field conditions.

### 3.6 Limitations

Despite its strong robustness under varied SHM conditions, the KAZE-RANSAC framework still presents practical limitations that may affect broader deployment and generalization.

1. Device Specificity and Sensor Variability

All experiments were conducted using a single imaging device (iPhone 11, 12MP, 1/2.55″ sensor), representative of handheld and UAV-based SHM workflows. However, imaging characteristics—such as lens distortion, dynamic range, and noise profiles—vary across devices. Cross-sensor validation is necessary to assess robustness across diverse platforms, including industrial cameras and lower-quality field equipment.

2. Homography Assumption and Non-Planar Surfaces

The method assumes local planarity between image pairs. In practice, SHM targets often include curved piers, warped panels, or corrugated facades. These violate planar assumptions, leading to residual misalignments. Extending the framework with stereo vision, multi-view geometry, or LiDAR-assisted depth modeling may enable correction on non-planar scenes.

3. Crack Occlusion and Low Visual Salience

While KAZE is more resilient than traditional methods, it still depends on edge gradients. Under conditions of occlusion, surface contamination, or diffuse clutter, cracks may not yield detectable features. Preprocessing techniques—such as contrast normalization, shadow removal, or thermal-RGB fusion—could enhance performance in degraded environments.

4. Image Quality Sensitivity and Lighting Extremes

All images were captured under stable daylight. In low-light, motion blur, glare, or overexposure conditions, feature stability may degrade. Real-world UAV applications may require additional quality control modules for exposure correction, blur suppression, or confidence-based match rejection.

5. Short, Hairline, or Textureless Cracks

On smooth surfaces or prefabricated concrete, very short or hairline cracks may produce insufficient gradients for detection. In these cases, dense descriptors, sequential optical flow, or multi-temporal SFM pipelines may be needed to reconstruct fine-grained damage evolution.

6. Fixed Keypoint Sampling Strategy

A fixed set of 10 matched keypoints was used for all RANSAC iterations. This number was empirically chosen as the minimum ensuring geometric stability without overfitting to background noise. However, this setting may not generalize across scenes of varying texture or resolution. Adaptive keypoint selection—based on saliency, confidence, or complexity—warrants future exploration, especially for dynamic or large-scale SHM deployments.

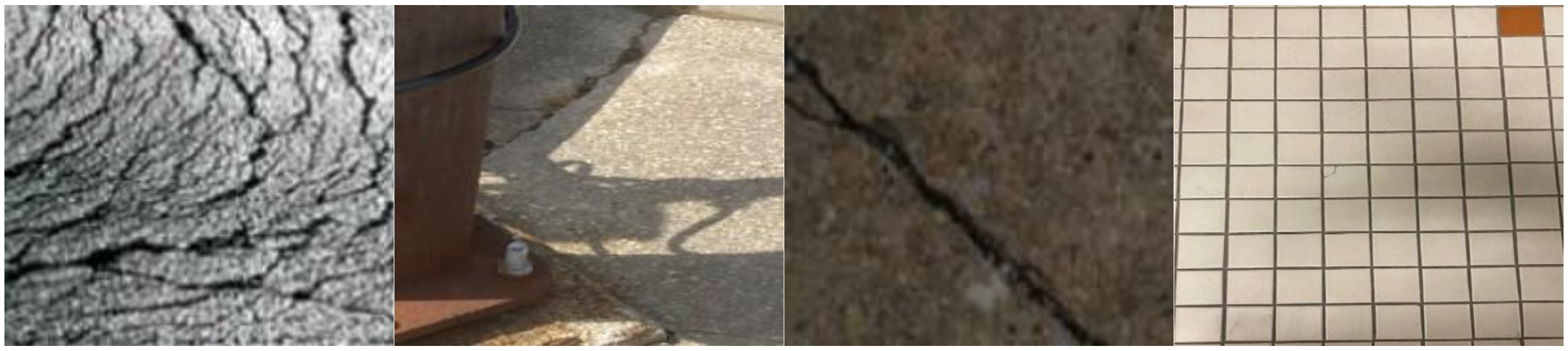

(a) Non-planar surfaces (b) Occlusion or low visual salience (c) Poor image quality (d) Short or textureless cracks

Fig. 3.1: Failure scenarios for KAZE-RANSAC

While these challenges exist, the proposed method remains fully unsupervised, interpretable, and training-free—traits that favor transparent and efficient deployment in field environments. To enhance future applicability, the following directions are prioritized:

- Cross-sensor and cross-platform performance validation
- Depth-informed geometric correction using stereo, LiDAR, or SfM
- Scene-aware or saliency-driven keypoint adaptation
- Hybrid descriptors combining KAZE with learned features (e.g., SuperPoint, D2-Net)
- On-device QA modules for UAV-based inspections (e.g., match dispersion, real-time feedback)

## 4. Conclusion

This study demonstrates that traditional Gaussian-based feature detectors—such as SIFT and SURF—struggle to achieve accurate crack alignment in field-based SHM imagery due to their suppression of fine edges and tendency to emphasize repetitive background textures. These limitations become critical under real-world conditions including shadow interference, occlusion, and low contrast. While binary descriptors like ORB and BRISK offer computational efficiency, they often suffer from sparse or semantically misaligned matches in challenging SHM scenes.

In contrast, the proposed KAZE-RANSAC framework—built on nonlinear anisotropic diffusion and empirically optimized fixed-keypoint sampling—consistently preserves crack geometry and enhances robustness. Across four diverse field scenarios, it achieved under 5% area error and alignment precision within 15% of baseline crack lengths, even under severe cropping and visual noise.

Key advantages of this framework include:

1. Overcoming the Gaussian-based bottleneck by preserving thin cracks in shadowed or low-texture environments;
2. Fully interpretable, training-free deployment, suitable for UAVs or mobile SHM platforms;
3. Superior spatial anchoring and semantic alignment compared to both classical and binary feature pipelines.

While limitations remain—such as performance under non-planar geometry and minimal-contrast cracks—this lightweight, physics-informed approach provides a transparent alternative to black-box CNN models. Future work will prioritize hybrid descriptors, stereo-informed geometry correction, and adaptive keypoint sampling to extend applicability in complex SHM applications.

**Declarations**

Funding

This research did not receive any specific grant from funding agencies in the public, commercial, or not-for-profit sectors.

Declaration of Generative AI and AI-Assisted Technologies in the Writing Process

During the preparation of this article, the author utilized ChatGPT to assist with language refinement and editing. Following the use of this tool, the author reviewed and edited the content independently and takes full responsibility for the final version of the work.

Data Availability

The data supporting the findings of this study are available from the corresponding author upon reasonable request.

Submission Declaration

This article is an original work that has not been published previously and is not under consideration for publication elsewhere.

Acknowledgements

The author would like to thank Professor Peter Chang for his valuable discussions and constructive feedback throughout this work.